\pdfoutput=1
\documentclass[10pt]{article}
\usepackage{symbiont}
\hypersetup{pdftitle={From Autonomous Agents to Cognitive Symbionts: Toward AI-Augmented Human Intelligence},
            pdfauthor={Hao Fei}}
\graphicspath{{figures/}}

\begin{document}
\thispagestyle{empty}
\newcommand{\coverimage}{figures/cover-v5.png}
\tikzfading[name=coverfade,top color=transparent!0,middle color=transparent!15,bottom color=transparent!100]
\begin{tikzpicture}[remember picture,overlay]
  \begin{scope}
    \clip (current page.south west) rectangle (current page.north east);
    \node[anchor=center,inner sep=0pt] at (current page.center)
      {\includegraphics[width=\paperwidth]{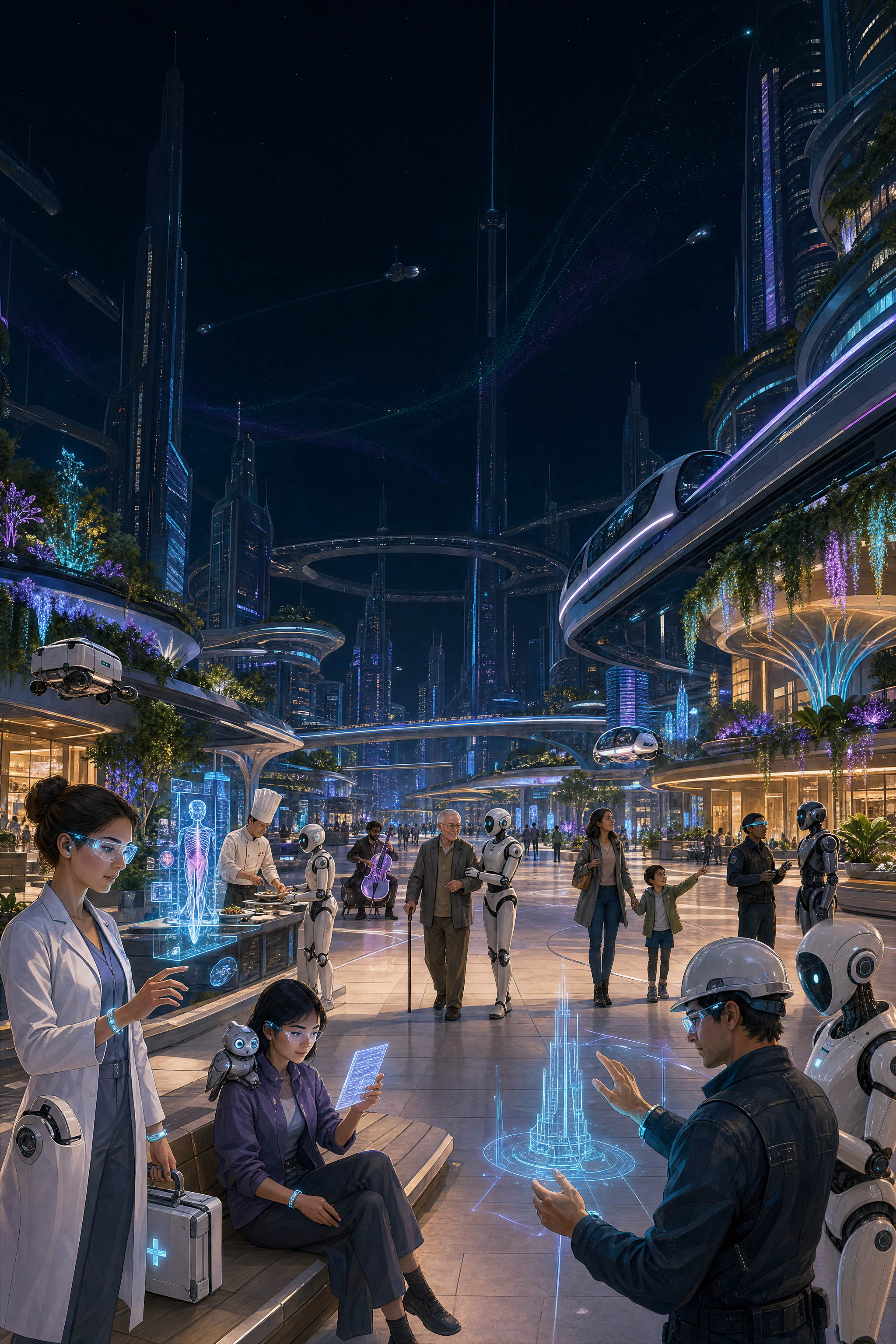}};
  \end{scope}
  \fill[black,path fading=coverfade,fill opacity=0.68]
    ([yshift=0.3cm]current page.north west) rectangle ([yshift=-22.5cm]current page.north east);
  \node[anchor=north,text width=0.86\paperwidth,align=center,text=white,inner sep=0pt] at ([yshift=-1.35cm]current page.north) {%
    {\fontsize{23}{27}\selectfont\bfseries\mbox{From Autonomous Agents to Cognitive Symbionts}\par}
    \vspace{0.28cm}
    {\fontsize{15.5}{18}\selectfont Toward AI-Augmented Human Intelligence\par}
    \vspace{0.62cm}
    {\fontsize{12}{14}\selectfont\bfseries Hao Fei\par}
    \vspace{0.14cm}
    {\fontsize{9.5}{12}\selectfont\color{white!88!symCyan}University of Oxford\quad\textbullet\quad haofei7419@gmail.com\par}
    \vspace{0.34cm}
    {\fontsize{9.5}{12}\selectfont
     \tikz[baseline=(u.base)]\node[draw=white!72,rounded corners=7pt,inner xsep=8pt,inner ysep=3.2pt,line width=0.5pt] (u) {\color{white!88!symCyan}\faGlobe\ \href{https://cognitive-symbiont.github.io}{\color{white!88!symCyan}cognitive-symbiont.github.io}};\par}
    \vspace{0.5cm}
    {\fontsize{11}{13}\selectfont\bfseries Abstract\par}
    \vspace{0.18cm}
    \begin{minipage}{0.97\linewidth}
      \fontsize{10.2}{12.6}\selectfont\color{white!94!symSlate}\justifying\noindent
      \let\keyterm\coverkeyterm
      Over the past decade, progress in artificial intelligence has been measured by how much a machine can do on its own.
Large foundation models now write code, draft documents, search, and run multistep tasks, and the autonomous personal agents run in the background and act on a person's behalf.
Yet people still forget what they once knew, miss what they are not looking at, fail to find the right words, and face hard choices without the benefit of their own history.
The machine that could help usually stands outside the situation: it does not know who the person is, where they are, what they are thinking, or how to interact and communicate with them in the moment.
As AI moves into the physical world, attention has settled on one kind of intelligent individual, the embodied robot; we argue that the era of Physical AI holds a second kind of equal importance, the person augmented by AI, and with it a second half of the endgame of AI.
Alongside artificial general intelligence, which asks how intelligent a machine can be, stands \keyterm{AI-augmented human intelligence (AHI)}, which asks how much intelligence a machine can enable in a particular person.
Realizing it calls for a new kind of system, which we call a \keyterm{cognitive symbiont}: a lifelong, embedded, real-time, private, low-friction, and proactive companion that senses what the person senses through external devices, keeps their context and knowledge across years, models their situation and their mind, takes part in their thinking at the right moment, carries out everyday tasks with their context in hand, and keeps learning the person as both of them change.
We give the paradigm a conceptual core, the cognitive gap between what a person could bring to a situation and what they can hold in working memory, and organize its design space as a symbiosis loop of sensing, keeping, understanding, living alongside, acting for and thinking with the person, and growing, enclosed by a personal runtime and by trust.
Because a symbiont changes what a person can do in front of others, it also raises risks of a new kind, from covert assistance in examinations and conversations to pressure on everyone to adopt it, and we treat their governance as part of the paradigm.

    \end{minipage}\par
  };
\end{tikzpicture}
\null

\clearpage
\pagenumbering{arabic}\setcounter{page}{2}
\setcounter{tocdepth}{2}
{\small\tableofcontents}
\clearpage

\markright{Map of the Perspective Paper}
\noindent\begin{minipage}{\linewidth}
\centering
\resizebox{!}{0.8\textheight}{\begin{tikzpicture}[x=1cm,y=1cm,font=\sffamily,
  pbox/.style={draw=#1,fill=#1!8,rounded corners=4pt,line width=0.7pt,text width=3.4cm,inner sep=5pt,align=left,font=\sffamily\scriptsize,minimum height=2.5cm},
  rbox/.style={draw=#1,fill=#1!10,rounded corners=4pt,line width=0.8pt,text width=3.4cm,inner sep=5pt,align=left,font=\sffamily\scriptsize,minimum height=2.15cm},
  flow/.style={-{Stealth[length=2mm]},line width=0.7pt,color=symSlate},
  loopa/.style={-{Stealth[length=2.6mm]},line width=1.3pt,color=symInk!75},
  plab/.style={font=\sffamily\footnotesize\bfseries,anchor=west,text=symInk},
  psub/.style={font=\sffamily\scriptsize,anchor=west,text=symSlate}]
\newcommand{\bt}[3]{{\footnotesize\bfseries\textcolor{#3}{#1}\;\;#2}\\[1.5pt]}
\newcommand{\ba}[1]{\\[1.5pt]{\color{symSlate}\fontsize{6}{7}\selectfont #1}}
\node[plab] at (-7.95,1.85) {Part I\quad The Paradigm};
\node[psub] at (-3.85,1.85) {Sections 1 to 4: the case for a second half of the endgame, and the concepts that define it};
\node[pbox=symSlate] (b1) at (-5.9,0) {\bt{1}{Introduction}{symSlate} A second form of intelligent individual for the era of Physical AI; AHI beside AGI; the bottleneck has moved to coupling; risk and legitimacy; thesis.\ba{Fig.~\ref{fig:twohalves}, \ref{fig:bottleneck}}};
\node[pbox=symSlate] (b2) at (-1.97,0) {\bt{2}{Where we stand}{symSlate} Four roles in human--AI relations; two forms of intelligent individuals; eighty years of the same question; why now.\ba{Fig.~\ref{fig:twoforms}\;\textbullet\;Tab.~\ref{tab:roles}, \ref{tab:lineage}}};
\node[pbox=symSlate] (b3) at (1.97,0) {\bt{3}{Technical foundations}{symSlate} Five lines of technology (foundation models, agents, memory, world models, wearables): what each already gives a symbiont, and what it lacks.\ba{Fig.~\ref{fig:stack}\;\textbullet\;Tab.~\ref{tab:foundations}}};
\node[pbox=symSlate] (b4) at (5.9,0) {\bt{4}{Conceptual foundations}{symSlate} Self-context, the cognitive gap, the symbiont, the augmented individual; six properties and the Jarvis test; the symbiosis loop.\ba{Eq.~\ref{eq:gap}, \ref{eq:aug}\;\textbullet\;Fig.~\ref{fig:loop}\;\textbullet\;Tab.~\ref{tab:properties}}};
\draw[flow] (b1) -- (b2); \draw[flow] (b2) -- (b3); \draw[flow] (b3) -- (b4);
\node[plab] at (-7.95,-3.55) {Part II\quad The Anatomy of a Symbiont};
\node[psub] at (-2.95,-3.55) {Sections 5 to 12: one loop around the person, with a runtime beneath it and trust around it};
\draw[draw=stTrust,line width=0.9pt,fill=stTrust!3,rounded corners=8pt] (-7.95,-4.25) rectangle (7.95,-14.05);
\node[fill=white,draw=stTrust,line width=0.7pt,rounded corners=3pt,inner sep=3pt,align=center,font=\sffamily\scriptsize] at (0,-4.25)
  {{\footnotesize\bfseries\textcolor{stTrust}{12}\;\;Trust}\\[1pt]the trust curve, the symbiont's code, the privacy of thought, bystanders, design principles\\[1pt]{\color{symSlate}\fontsize{6}{7}\selectfont Fig.~\ref{fig:trustcurve}\;\textbullet\;Tab.~\ref{tab:principles}}};
\node[rbox=stSense] (s5) at (-3.95,-6.1) {\bt{5}{Sensing and interacting}{stSense!80!black} Borrowed senses and a borrowed voice: the device constellation, channels out, private channels, always on.\ba{Fig.~\ref{fig:devices}\;\textbullet\;Tab.~\ref{tab:signals}, \ref{tab:channels}}};
\node[rbox=stKeep] (s6) at (3.95,-6.1) {\bt{6}{Context and knowledge}{stKeep!80!black} One record on three timescales; one entry point for every context; many views; forgetting, repair, and the future.\ba{Fig.~\ref{fig:timescales}, \ref{fig:views}\;\textbullet\;Tab.~\ref{tab:sources}, \ref{tab:views}}};
\node[rbox=stUnderstand] (s7) at (5.9,-9.0) {\bt{7}{Understanding a person}{stUnderstand!80!black} Two roads to the mind; seven layers of the personal model; a replaceable backbone; two world models.\ba{Eq.~\ref{eq:belief}\;\textbullet\;Fig.~\ref{fig:tworoads}, \ref{fig:backbones}\;\textbullet\;Tab.~\ref{tab:layers}}};
\node[rbox=stLive] (s8) at (3.95,-11.9) {\bt{8}{Living alongside}{stLive!80!black} Cognitive full-duplex; fast and slow loops; implicit intent; an intervention policy with silence as an output.\ba{Eq.~\ref{eq:policy}\;\textbullet\;Fig.~\ref{fig:duplex}\;\textbullet\;Tab.~\ref{tab:policy}}};
\node[rbox=stHelp] (s9) at (-3.95,-11.9) {\bt{9}{The ladder of assistance}{stHelp!80!black} Getting things done; thinking and feeling with you; living among others; caring; symbiont to symbiont.\ba{Fig.~\ref{fig:ladder}, \ref{fig:handshake}\;\textbullet\;Tab.~\ref{tab:sametask}, \ref{tab:catalogue}}};
\node[rbox=stGrow] (s10) at (-5.9,-9.0) {\bt{10}{Growing together}{stGrow!80!black} Seven places of learning; the personal data flywheel; learning the way a person learns; evolving the whole stack.\ba{Eq.~\ref{eq:update}\;\textbullet\;Fig.~\ref{fig:spectrum}, \ref{fig:coevolution}\;\textbullet\;Tab.~\ref{tab:learning}}};
\node[circle,draw=symInk,fill=white,line width=1pt,minimum size=2.1cm,align=center,font=\sffamily\scriptsize] (person) at (0,-9.0)
  {{\footnotesize\bfseries the person}\\[1pt]and the symbiont\\$h \oplus s$};
\draw[loopa] (s5.east) -- (s6.west);
\draw[loopa] (s6.south) to[out=-90,in=90] (s7.north);
\draw[loopa] (s7.south) to[out=-90,in=90] (s8.north);
\draw[loopa] (s8.west) -- (s9.east);
\draw[loopa] (s9.north) to[out=90,in=-90] (s10.south);
\draw[loopa] (s10.north) to[out=90,in=-90] (s5.south);
\node[draw=stRuntime,fill=stRuntime!8,rounded corners=4pt,line width=0.8pt,minimum width=13.6cm,minimum height=0.8cm,font=\sffamily\scriptsize] (rt) at (0,-13.45)
  {{\footnotesize\bfseries\textcolor{stRuntime}{11}\;\;The personal runtime}\quad devices, permissions, scheduling, continuity, the cost of always-on cognition\quad{\color{symSlate}\fontsize{6}{7}\selectfont Fig.~\ref{fig:runtime}}};
\node[plab] at (-7.15,-15.15) {Part III\quad Where This Leads};
\node[psub] at (-2.0,-15.15) {Sections 13 and 14: who gains, and what a society of augmented individuals looks like};
\node[pbox=symAccent,text width=7.25cm,minimum height=1.6cm] (b13) at (-3.93,-16.75) {\bt{13}{Whom it serves}{symAccent} Four criteria turn ``everyone can benefit'' into an argument: everyday life, knowledge work, learners, families, barriers, mental health, public roles; who should not come first.\ba{Fig.~\ref{fig:beneficiaries}\;\textbullet\;Tab.~\ref{tab:whogains}}};
\node[pbox=symAccent,text width=7.25cm,minimum height=1.6cm] (b14) at (3.93,-16.75) {\bt{14}{A world of augmented individuals}{symAccent} A street of augmented individuals; humans and robots, the two forms of intelligent individual in the era of Physical AI, share one cognitive core; reach into the physical world.\ba{returns to Fig.~\ref{fig:twoforms}}};
\draw[flow] (b13) -- (b14);
\node[plab] at (-7.15,-18.6) {Part IV\quad Risk, Governance, and Evidence};
\node[psub] at (-0.75,-18.6) {Sections 15 to 18: the risks, their governance, and the evidence that would test the paradigm};
\node[pbox=stTrust] (b15) at (-5.9,-20.5) {\bt{15}{Risks}{stTrust} Why augmentation is different; three directions of harm; public acceptance; examinations, conversations, institutions; the social fabric; contested uses.\ba{Fig.~\ref{fig:harms}\;\textbullet\;Tab.~\ref{tab:contested}}};
\node[pbox=stTrust] (b16) at (-1.97,-20.5) {\bt{16}{Governance}{stTrust} Private content, declarable status; settings set the rules; verifiable abstention; what credentials measure; access; who governs; open ethical questions.\ba{vignette: Exam Day}};
\node[pbox=stTrust] (b17) at (1.97,-20.5) {\bt{17}{Evaluation and agenda}{stTrust} The augmented individual as the unit of measurement; ten dimensions; twenty-three open problems; a capability roadmap; falsifiers.\ba{Eq.~\ref{eq:syn}--\ref{eq:pers}\;\textbullet\;Fig.~\ref{fig:roadmap}\;\textbullet\;Tab.~\ref{tab:evaluation}, \ref{tab:openproblems}}};
\node[pbox=stTrust] (b18) at (5.9,-20.5) {\bt{18}{Conclusion}{stTrust} The name, the map, and the standard of evidence this paper leaves behind.\ba{}};
\draw[flow] (b15) -- (b16); \draw[flow] (b16) -- (b17); \draw[flow] (b17) -- (b18);
\draw[flow,line width=0.9pt] (b4.south) -- node[right,font=\sffamily\scriptsize,text=symSlate,align=left,xshift=1pt] {the loop of Section 4\\is the map of Part II} (5.9,-4.25);
\draw[flow,line width=0.9pt] (-7.6,-14.05) -- node[right,font=\sffamily\scriptsize,text=symSlate,pos=0.3,xshift=1pt] {from one augmented individual to many} (-7.6,-15.92);
\draw[flow,line width=0.9pt] (-7.6,-17.58) -- node[right,font=\sffamily\scriptsize,text=symSlate,pos=0.3,xshift=1pt] {what could go wrong, and how to know} (-7.6,-19.22);
\end{tikzpicture}}

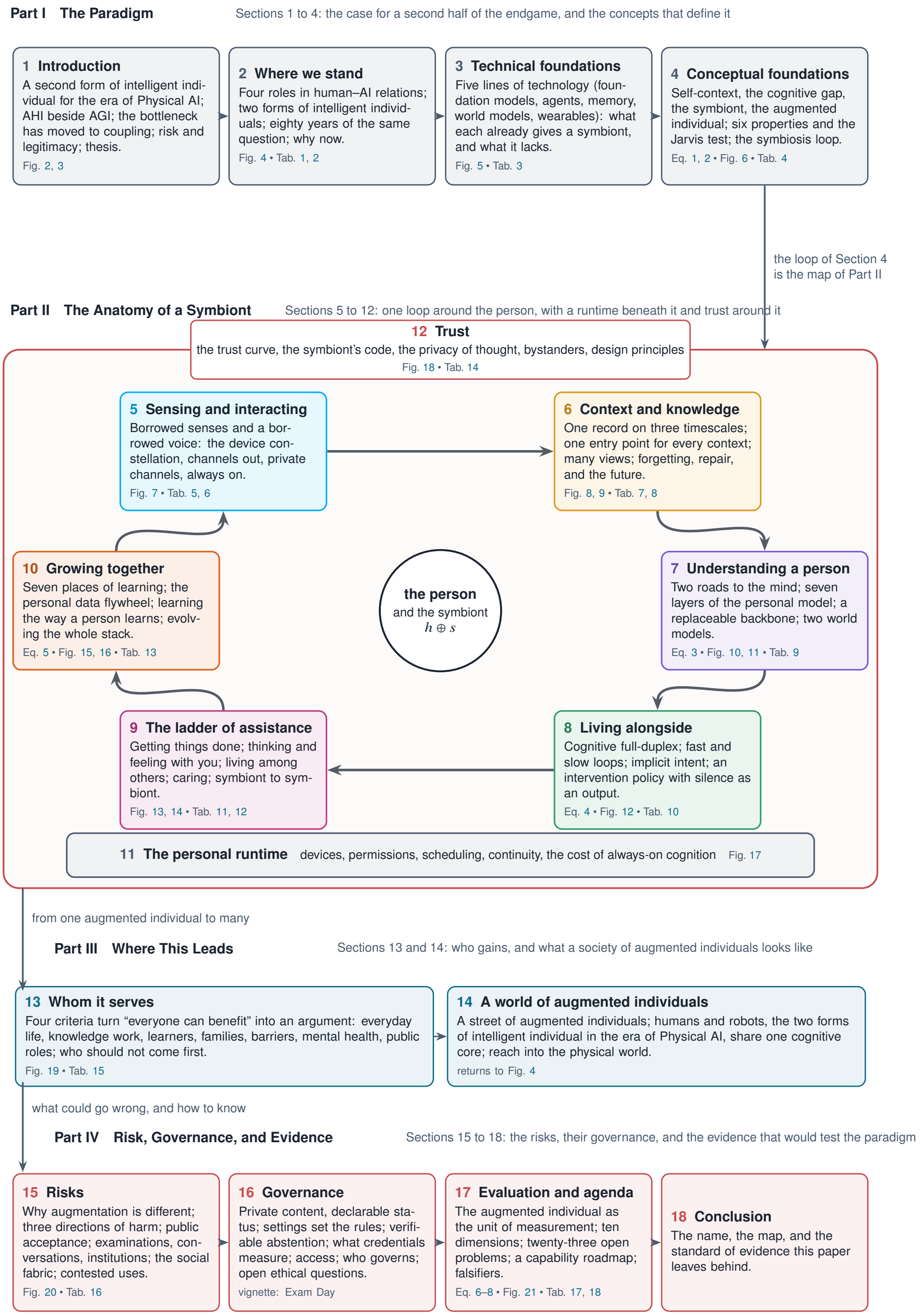
\captionof{figure}{A map of the perspective paper.
Part~I says why the era of Physical AI holds a second form of intelligent individual, the person augmented by AI, and what a cognitive symbiont is.
Part~II follows one loop, in which the output of each section is the input of the next, with a runtime beneath it and trust around it.
Part~III widens the view from one augmented individual to many.
Part~IV examines the risks, how augmentation can be governed, and how the whole can be measured.
Each box names the section's purpose and the figures, tables, and equations it contributes.}
\label{fig:map}
\end{minipage}

\clearpage

\paperpart{The Paradigm}\label{part:paradigm}
\begin{partlead}
\noindent Part~I makes the case for a second half of the endgame of AI and gives it a conceptual form.
Section~\ref{sec:intro} places the augmented person beside the embodied robot as a form of intelligent individual in the era of Physical AI, argues that the bottleneck of personal AI has moved from the capability of the model to the coupling between a model and a particular person, and says why the risks of augmentation belong to the proposal.
Section~\ref{sec:wherewestand} places the proposal in a coordinate system of four roles the AI has played in its relationship with a person and two forms an intelligent individual can take, and locates it in seventy years of work on augmenting the human intellect.
Section~\ref{sec:foundations} reads the five lines of technology a symbiont is built on, stating for each what it already provides and what it still lacks; the gaps form the research space of Part~II.
Section~\ref{sec:concepts} defines the objects of the paper, introduces the cognitive gap as its invariant, lists the six properties a symbiont must have, and draws the symbiosis loop that organizes everything that follows.
\end{partlead}

\section{Introduction}\label{sec:intro}

\begin{epigraphbox}
\small\itshape ``\ldots human brains and computing machines will be coupled together very tightly.''\hfill\upshape -- J.~C.~R.~Licklider, \textit{Man-Computer Symbiosis}, 1960~\citep{licklider1960symbiosis}
\end{epigraphbox}

For the past decade, progress in artificial intelligence has been scored on a single axis: how much a machine can accomplish on its own.
Measured on that axis, the results are substantial.
AI models now can write and review code, draft and summarize documents, search, plan, and carry out multistep tasks on a computer~\citep{openai2023gpt4,yang2024sweagent,anthropic2024computeruse}.
Productivity at the desk, the part of work that takes place in front of a screen, has been transformed and in large part solved.
An ordinary day, however, still contains the same losses it always did, and one person's day is enough to show them.
In the morning, working alone, a researcher has an idea that slips away before it is written down.
In the afternoon, in a meeting, a colleague asks about a result she published three years ago, and she cannot recall the number.
In the evening, facing a decision, she cannot remember how she reasoned the last time she faced one like it.
Two further losses belong to the same family: a visitor in an unfamiliar country who does not know how to decline an invitation politely, and someone who has just failed at something, who needs one sentence from a person who knows their story and receives a paragraph of generic encouragement instead.
The first three are failures of access, in which something the person already possesses fails to reach the moment in which it is needed, and the last two are gaps of knowledge and of support that the person cannot close alone at that moment; none of the five is a failure of intelligence.
In each case an AI that holds the relevant knowledge exists, and in each case it is absent from the situation: it does not know who this person is, where they are, what they are trying to do, or what they are feeling, and it has no natural way to interact and communicate with them in the moment.
Sections~\ref{sec:thinking}, \ref{sec:beyondmeeting}, and~\ref{sec:future} return to the three moments of this day.

Most readers already carry a picture of the system that would close these gaps.
In the \textit{Iron Man} films, Jarvis is always present, has known its owner for years, acts while he is speaking and moving, and never asks him to stop and explain.\footnote{J.A.R.V.I.S.\ appears in \textit{Iron Man} (2008) through \textit{Avengers: Age of Ultron} (2015).
We use it only as a shared reference point, never as evidence.}
The reference is useful because it allows the subject of this perspective paper to be stated in one sentence: we are concerned with how every person could have such a system, and with how far beyond Jarvis such a system should go.
The vignette \emph{Beyond Jarvis} in Section~\ref{sec:properties} lists six directions, of which three matter most here, and each is developed later in the paper.
Jarvis sees through a laboratory and a suit that its owner built, whereas a \sym for everyone must see through a handful of affordable external devices, and it gains channels Jarvis never had, such as physiological signals, first-person video, and the owner's own account, in their own words, of what they are thinking.
Jarvis mostly executes and alerts, and rarely models its owner's goals, emotions, or blind spots, whereas the center of gravity of a \sym is exactly that layer of mind.
And Jarvis gives almost no help with other people or other cultures, which is where a \sym is most distinctive.
Section~\ref{sec:properties} makes this comparison precise.

\subsection{The Other Half of the Endgame}\label{sec:otherhalf}
AI is leaving the screen for the physical world, and with it comes the question of what kinds of intelligent individuals that world will contain.
The answer that has drawn most of the field's attention is the embodied robot, an AI with a body, senses, and goals of its own~\citep{geminirobotics2025,nvidia2025cosmos}.
We argue that the era of Physical AI will contain a second kind of individual of equal importance: a person whose own senses, memory, and judgment are joined to an AI that lives on their body (Section~\ref{sec:twoforms}).
This paper is a perspective on that second kind of individual.
Foundation models, world models, and methods of self-improvement enter it as enabling technologies (Section~\ref{sec:foundations}) and as replaceable components (Section~\ref{sec:backbone}), while the proposal itself concerns the form in which an augmented person exists and is helped.

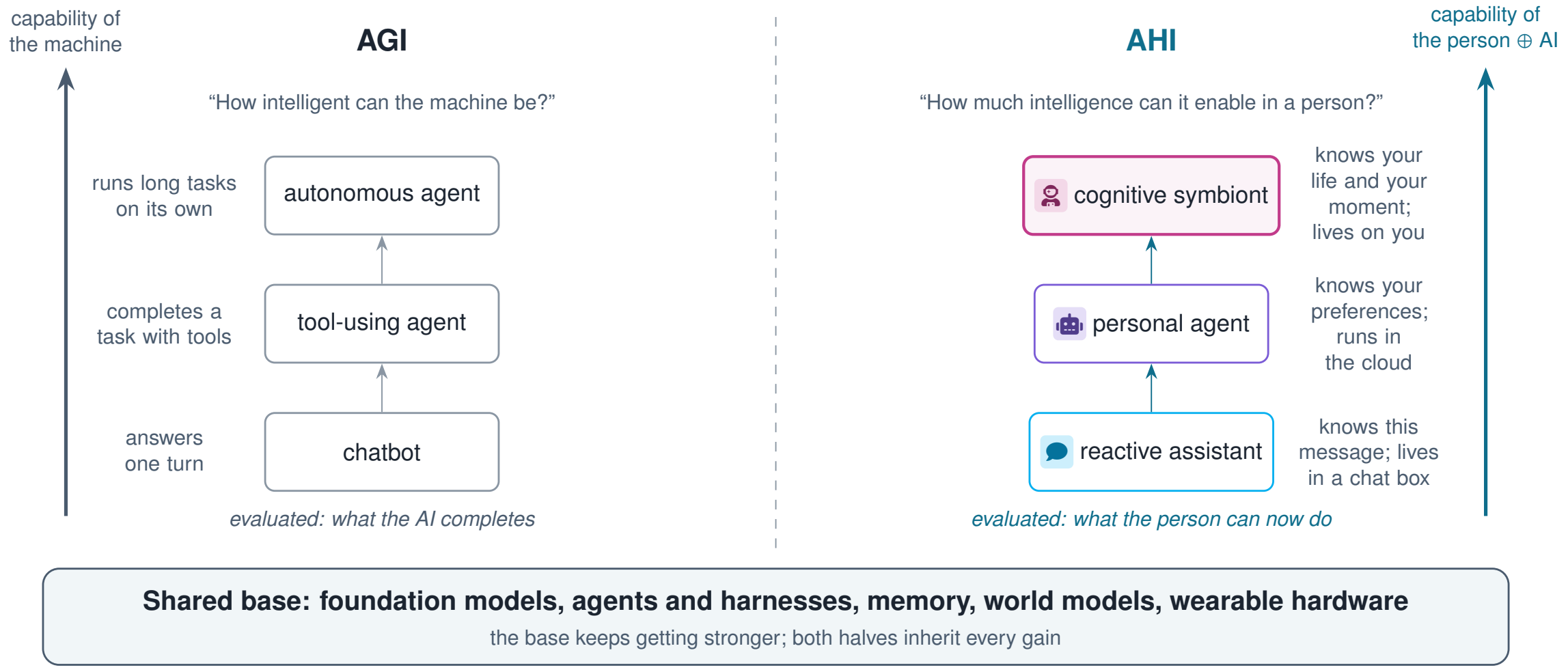
\begin{figure}[t]
  \centering
  \resizebox{0.96\linewidth}{!}{\begin{tikzpicture}[x=1cm,y=1cm,font=\small\sffamily,
  axisline/.style={-{Stealth[length=2.5mm]},line width=1.1pt},
  stage/.style={draw,rounded corners=3pt,minimum width=2.55cm,minimum height=0.85cm,align=center,font=\footnotesize\sffamily,fill=white,line width=0.6pt},
  lab/.style={font=\scriptsize\sffamily\color{symSlate},align=center}]
  \node[draw=symSlate,fill=symFaint,rounded corners=5pt,minimum width=16.0cm,minimum height=1.05cm,line width=0.6pt] (base) at (0,0) {};
  \node[font=\small\sffamily\bfseries,text=symInk] at (0,0.15) {Shared base: foundation models, agents and harnesses, memory, world models, wearable hardware};
  \node[lab] at (0,-0.25) {the base keeps getting stronger; both halves inherit every gain};
  \draw[axisline,color=symSlate] (-7.75,1.1) -- (-7.75,6.0);
  \node[lab,anchor=south] at (-7.75,6.05) {capability of\\the machine};
  \node[stage,draw=symGray] (c1) at (-4.3,1.8) {chatbot};
  \node[stage,draw=symGray] (c2) at (-4.3,3.2) {tool-using agent};
  \node[stage,draw=symGray] (c3) at (-4.3,4.6) {autonomous agent};
  \draw[-{Stealth},symGray] (c1) -- (c2); \draw[-{Stealth},symGray] (c2) -- (c3);
  \node[font=\normalsize\sffamily\bfseries,text=symInk] at (-4.3,6.3) {AGI};
  \node[lab] at (-4.3,5.6) {``How intelligent can the machine be?''};
  \node[lab,text=symSlate,fill=white,inner sep=2pt] at (-4.3,1.05) {\textit{evaluated: what the AI completes}};
  \draw[dashed,symGray] (0,0.75) -- (0,6.6);
  \draw[axisline,color=symAccent] (7.75,1.1) -- (7.75,6.0);
  \node[lab,anchor=south,text=symAccent] at (7.75,6.05) {capability of\\the person $\oplus$ AI};
  \node[stage,draw=symCyan] (r1) at (4.1,1.8) {\badge{symCyan}{\faComment}\ reactive assistant};
  \node[stage,draw=symViolet] (r2) at (4.1,3.2) {\badge{symViolet}{\faRobot}\ personal agent};
  \node[stage,draw=symMagenta,fill=symMagenta!6,line width=0.9pt] (r3) at (4.1,4.6) {\badge{symMagenta}{\faUserAstronaut}\ cognitive symbiont};
  \draw[-{Stealth},symAccent] (r1) -- (r2); \draw[-{Stealth},symAccent] (r2) -- (r3);
  \node[font=\normalsize\sffamily\bfseries,text=symAccent] at (4.1,6.3) {AHI};
  \node[lab] at (4.1,5.6) {``How much intelligence can it enable in a person?''};
  \node[lab,text=symAccent,fill=white,inner sep=2pt] at (4.1,1.05) {\textit{evaluated: what the person can now do}};
  \node[lab,anchor=west,text width=1.6cm] at (5.55,1.8) {knows this message; lives in a chat box};
  \node[lab,anchor=west,text width=1.6cm] at (5.55,3.2) {knows your preferences; runs in the cloud};
  \node[lab,anchor=west,text width=1.6cm] at (5.55,4.6) {knows your life and your moment; lives on you};
  \node[lab,anchor=east,text width=1.6cm] at (-5.75,1.8) {answers one turn};
  \node[lab,anchor=east,text width=1.6cm] at (-5.75,3.2) {completes a task with tools};
  \node[lab,anchor=east,text width=1.6cm] at (-5.75,4.6) {runs long tasks on its own};
\end{tikzpicture}}
  \caption{The two halves of the endgame.
  AGI runs from chatbots to autonomous agents and is scored on what the AI completes.
  AHI runs from reactive assistants to cognitive symbionts and is scored on what the person can now do.
  Both share one base of models, agents, memory, and hardware, and both inherit every improvement to it.
  A system sits on one half or the other according to what it is evaluated on, so the same chat product is a chatbot when its answers are scored and a reactive assistant when what its user can then do is scored; today the first two stages of the AHI half are still scored on what the AI produces (\tabref{tab:roles}).
  Arrows mark growth within a half, and the two halves are parallel, not successive stages.}
  \label{fig:twohalves}
\end{figure}

The second kind of individual also changes what progress in AI is measured against, and in this sense the endgame of AI has two halves.
Artificial general intelligence (AGI) asks \emph{how intelligent the machine can be}.
The other half, which we call \keyterm{AI-augmented human intelligence (AHI)}, asks \emph{how much intelligence a machine can enable in a particular person}.
The two are built on the same base: the same foundation model can act as an autonomous agent that completes tasks, or as a \keyterm{cognitive symbiont} that augments one human being.
What differs is the object of evaluation, the mode of existence, and the relationship with the person (\figref{fig:twohalves}).
This paper is entirely about the second half, and it makes two commitments to the first.
A symbiont delegates errands to standard agents (Section~\ref{sec:tier1}), and it improves with every gain in the base model (Section~\ref{sec:backbone}).

\subsection{The Bottleneck Has Moved}\label{sec:bottleneck}
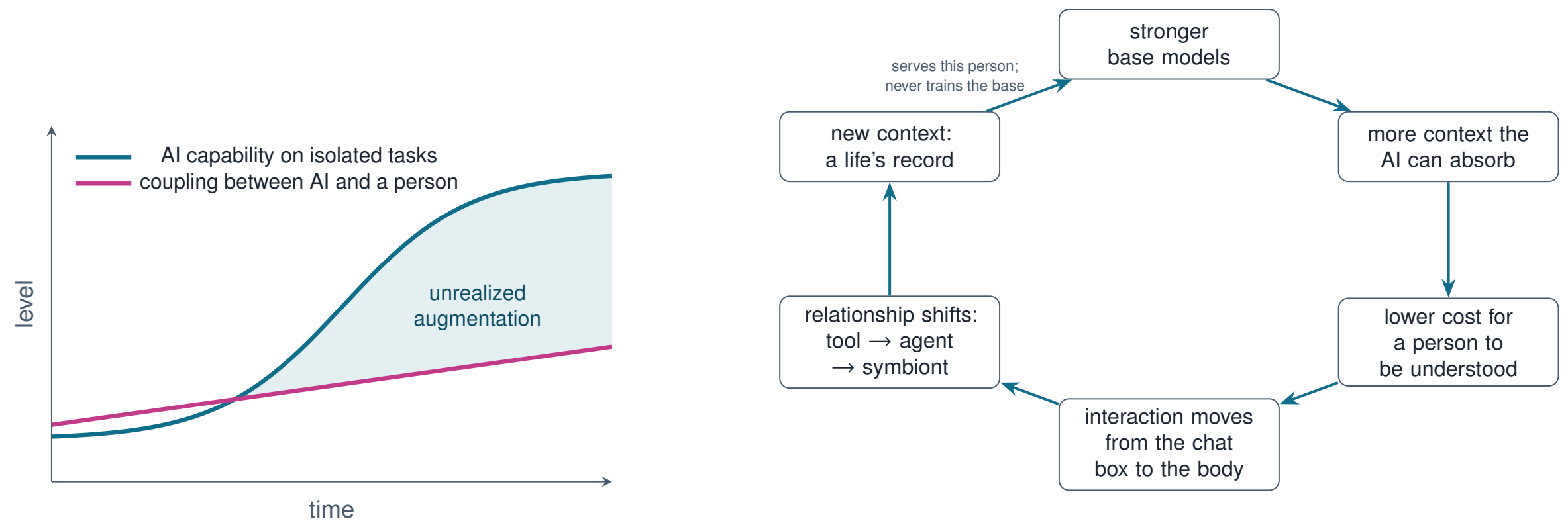
\begin{figure}[t]
  \centering
  \resizebox{0.98\linewidth}{!}{\begin{tikzpicture}[font=\small\sffamily]
  \begin{axis}[name=ax,width=7.6cm,height=5.4cm,axis lines=left,xmin=0,xmax=10,ymin=0,ymax=10,
     xlabel={time},ylabel={level},xtick=\empty,ytick=\empty,xlabel near ticks,ylabel near ticks,
     every axis label/.append style={font=\footnotesize\sffamily\color{symSlate}},
     legend style={at={(0.02,0.98)},anchor=north west,draw=none,fill=none,font=\scriptsize\sffamily,row sep=-2pt},
     axis line style={symSlate}]
     \addplot[name path=A,symAccent,line width=1.3pt,domain=0:10,samples=60] {1.2+7.5/(1+exp(-(x-5.2)*0.9))};
     \addlegendentry{AI capability on isolated tasks}
     \addplot[name path=B,symMagenta,line width=1.3pt,domain=0:10,samples=40] {1.6+0.22*x};
     \addlegendentry{coupling between AI and a person}
     \addplot[symAccent!20,opacity=0.55] fill between[of=A and B,soft clip={domain=3.2:10}];
     \node[font=\scriptsize\sffamily,text=symAccent!70!black,align=center] at (axis cs:7.6,4.9) {unrealized\\augmentation};
  \end{axis}
  \node[anchor=north,font=\footnotesize\sffamily\bfseries] at ([yshift=-0.35cm]ax.south) {(a) The bottleneck has moved};
  \begin{scope}[shift={(12.0cm,0.3cm)},node distance=0.4cm,
      stepbox/.style={draw=symSlate,rounded corners=3pt,fill=white,align=center,font=\scriptsize\sffamily,text width=2.15cm,inner sep=3pt,minimum height=0.75cm},
      arr/.style={-{Stealth[length=2mm]},symAccent,line width=0.8pt}]
    \node[stepbox] (s1) at (0,4.4) {stronger base models};
    \node[stepbox] (s2) at (3.0,3.3) {more context the AI can absorb};
    \node[stepbox] (s3) at (3.0,1.2) {lower cost for a person to be understood};
    \node[stepbox] (s4) at (0,0.1) {interaction moves from the chat box to the body};
    \node[stepbox] (s5) at (-3.0,1.2) {relationship shifts: tool $\to$ agent $\to$ symbiont};
    \node[stepbox] (s6) at (-3.0,3.3) {new context: a life's record};
    \draw[arr] (s1) -- (s2); \draw[arr] (s2) -- (s3); \draw[arr] (s3) -- (s4);
    \draw[arr] (s4) -- (s5); \draw[arr] (s5) -- (s6); \draw[arr] (s6) -- node[above left,font=\tiny\sffamily,text=symSlate,align=center,inner sep=1pt] {serves this person;\\never trains the base} (s1);
    \node[anchor=north,font=\footnotesize\sffamily\bfseries] at (0,-0.75) {(b) Why the relationship shifts};
  \end{scope}
\end{tikzpicture}}
  \caption{(a) AI capability on isolated tasks keeps rising while coupling with a particular person stays nearly flat; the shaded region is augmentation that is never realized.
  The curves are a conceptual sketch, not measurements.
  (b) The mechanism of the shift: a causal loop in which stronger base models lower the cost of being understood, move interaction onto the body, and generate a new kind of context that makes the models more useful to each person in turn, without being used to train them.}
  \label{fig:bottleneck}
\end{figure}

A person's cognitive bandwidth is fixed.
Working memory holds a few items, attention goes one place at a time, and a day has twenty-four hours.
The capability of AI on isolated tasks keeps rising, but the \emph{coupling} between an AI and a given person has barely changed: how much the AI knows about the person, how fast it can step in, how much it interrupts when it does, and how far the person trusts it.
The result is a widening gap between what a model could contribute and what a person actually receives (\figref{fig:bottleneck}a).
\claim{The next step for personal AI is tighter coupling, and tighter coupling is a different problem from a larger model.}
The claim requires qualification, since coupling is itself built on strong base models; Section~\ref{sec:foundations} describes which technologies first made coupling possible, and the paper opposes only the assumption that a larger base is sufficient.
Three observable symptoms mark insufficient coupling: every conversation starts by re-explaining the background, help arrives too late or too loudly, and people entrust nothing important to the system.
They point, respectively, to Sections~\ref{sec:context}, \ref{sec:living}, and~\ref{sec:trust}.
\figref{fig:bottleneck}b shows the mechanism by which the relationship nonetheless shifts: stronger models absorb more context, which lowers the cost of being understood, which moves interaction off the screen and onto the body, which produces a new kind of context, the record of a life, which makes those models more useful to that person in turn.

\subsection{Risk and Legitimacy}\label{sec:risklegit}
A system that makes a person more capable in front of other people raises questions that a system which only completes tasks does not.
When a student can recall a textbook during an examination, or a negotiator is handed a line that reads the other side's hesitation, the gain of one person is a cost to another, and the other party usually does not know.
The same privacy that keeps the exchange between a person and their symbiont from intruding on others also hides the help from an examiner, an interviewer, or a friend.
Products built on this capability have already been sold for cheating, as Cluely was in 2025~\citep{techcrunch2025cluely}, and chatbot providers have switched off features for the duration of China's college entrance examination~\citep{bloomberg2025gaokao}.
Public attitudes toward cognitive enhancement are cautious, and the main objection is fairness~\citep{pew2022enhancement,faber2016unacceptable}.
We treat these concerns as part of the proposal.
A symbiont differs from earlier cognitive enhancements in a respect that matters for governance: it is external, removable, and inspectable, so it can follow the rules of the settings a person enters and show that it did.
Part~IV develops the risks and the governance that follows from this property.

\subsection{Thesis and Contribution}\label{sec:thesis}
Our thesis is the following.
Realizing the second half of the endgame requires an AI that exists in a new way: a cognitive symbiont that senses what the person senses, keeps their life, models their situation and their mind, takes part in their thinking at the right moment, carries out their errands with their context in hand, and keeps learning them while both of them change.
The paper develops this thesis in four parts.
\figref{fig:map} lays them out.
Part~I (Sections~\ref{sec:wherewestand} to~\ref{sec:concepts}) locates the paradigm historically and technically and gives it a conceptual core.
Part~II (Sections~\ref{sec:sensing} to~\ref{sec:trust}) follows one loop from sensing to trust, in which the output of each section is the input of the next.
Part~III (Sections~\ref{sec:whom} and~\ref{sec:world}) asks who gains and what a society of augmented individuals looks like.
Part~IV (Sections~\ref{sec:risks} to~\ref{sec:conclusion}) examines what could go wrong, how augmentation can be governed, and how the claims of the paper can be measured.

The contribution of the paper is the paradigm itself, taken as a whole.
We name a second kind of intelligent individual for the era of Physical AI, the person augmented by AI, and a second half of the endgame of AI, AI-augmented human intelligence, which differs from the first by its object of evaluation, its mode of existence, and its relationship with the person.
We propose the cognitive symbiont as the form of system that realizes it, define it through the cognitive gap and six necessary properties, and organize its design space as a symbiosis loop whose stages are each given a mechanism, a set of research questions, and a way of being evaluated.
We treat the risks of augmentation, to the person and through the person to others, as part of the paradigm, and propose that a symbiont follow the rules of the settings its owner enters and be able to show that it did.
We change the unit of evaluation from the AI to the augmented individual and state the open problems of the direction as a research program.
We propose no new algorithm; the claim of the paper is to the framing, to the organization of the design space, and to the standard of evidence it sets.
Our hope is that the concept will orient a new paradigm of AI research, one in which the measure of progress is how much more capable each particular person has become.

\section{Where We Stand}\label{sec:wherewestand}
This section sets up the coordinate system used throughout the paper.
The vertical axis is the relationship: four roles the AI has played with a person, in the order in which each became technically possible.
The horizontal axis is existence: two forms an intelligent individual can take.
We then place the paper in eighty years of work on the same question and state what has changed.

\subsection{Four Roles in Human--AI Relations}\label{sec:roles}
\tabref{tab:roles} divides the relationship between a person and an AI into four roles.
Reading down the table, three properties change together.
What the AI knows widens from the current message to a whole life.
Where the AI sits moves from a chat window to a computer, to the cloud, and finally onto the body.
And the value it offers shifts from giving the person information, to doing tasks for them, to doing those tasks their way, to making the person themselves more capable.
The first three roles share an evaluation target, the AI, and the fourth changes it to the person.

\begin{table}[t]
\centering\footnotesize
\caption{Four roles in human--AI relations, ordered by the relationship with the person. Down the table the AI knows more (a message, a task, your preferences, your life), moves closer (a chat box, a computer, the cloud, your body), and its value shifts from giving you information to making you stronger. A symbiont contains the first three roles; it does not replace them.}
\label{tab:roles}
\setlength{\tabcolsep}{4pt}\renewcommand{\arraystretch}{1.3}
\begin{tabularx}{\linewidth}{@{}L{2.45cm}L{2.15cm}L{2.55cm}YL{2.0cm}YC{0.95cm}@{}}
\toprule
\tabhead \colhead{Role} & \colhead{Relation} & \colhead{Representative systems} & \colhead{What the AI knows} & \colhead{Where it is} & \colhead{Value to the person} & \colhead{Scored on} \\
\midrule
\lb{tbGray}{R}\ \rowhead{Reactive assistant} & query $\to$ response & ChatGPT, Claude, Gemini, and other chat products & this message & \ico{\faCommentDots}\ in a chat box & information on demand & \ib{tbTeal}{\faRobot} \\
\rowcolor{tabStripe}
\lb{tbTeal}{A}\ \rowhead{Autonomous agent} & goal $\to$ execution & Claude Code, Codex, computer-use agents & this task & \ico{\faDesktop}\ inside a computer & tasks completed on your behalf & \ib{tbTeal}{\faRobot} \\
\lb{tbViolet}{P}\ \rowhead{Personal agent} & context $\to$ personalized service & Muse, Dots, Grok Bot and other 2026 personal agents & your preferences, accounts, and long-term goals & \ico{\faCloud}\ in the cloud, in the background & tasks done your way; first steps out of the computer & \ib{tbTeal}{\faRobot} \\
\rowcolor{tabStripe}
\lb{tbPink}{S}\ \rowhead{Cognitive symbiont} & continuous shared context $\to$ joint cognition & none yet (fictional reference: Jarvis) & your life, your situation, what you are thinking now & \ico{\faUser}\ on you, all day & your own cognition and ability, augmented; present in the physical world & \ib{tbPink}{\faUser} \\
\bottomrule
\end{tabularx}
\tablegend{\textbullet\ Scored on: \ib{tbTeal}{\faRobot}\ what the AI produced\quad \ib{tbPink}{\faUser}\ how much stronger the person became.\quad \textbullet\ Role letters \lb{tbGray}{R} \lb{tbTeal}{A} \lb{tbViolet}{P} \lb{tbPink}{S} are reused in \tabref{tab:sametask}.}
\end{table}

The third role deserves an accurate description, since it is the closest existing relative of the fourth.
The personal agents released in 2026, such as Meta's Muse~\citep{meta2026muse}, xAI's Grok Bot~\citep{gigazine2026grokbot}, and OpenAI's Dots~\citep{openai2026dots}, already track long-term goals, run continuously in the background, and learn preferences.\footnote{Capabilities as described at launch; we do not evaluate these products here.}
It would be inaccurate to describe them as systems that merely complete tasks.
The difference from the fourth role is one of position and purpose: a personal agent acts for the person from the cloud, while a symbiont thinks with the person at their side.
\claim{The fourth role includes the first three.}
A symbiont has to be a competent autonomous agent and a competent personal agent before it can be anything more (Section~\ref{sec:tier1}), and the four roles can coexist inside one system.
We therefore read the table as four solutions to one problem under four technical conditions, and we avoid reading it as an inevitable sequence of eras.

\subsection{Two Forms of Intelligent Individuals}\label{sec:twoforms}
\begin{figure}[t]
  \centering
  \includegraphics[width=0.92\linewidth]{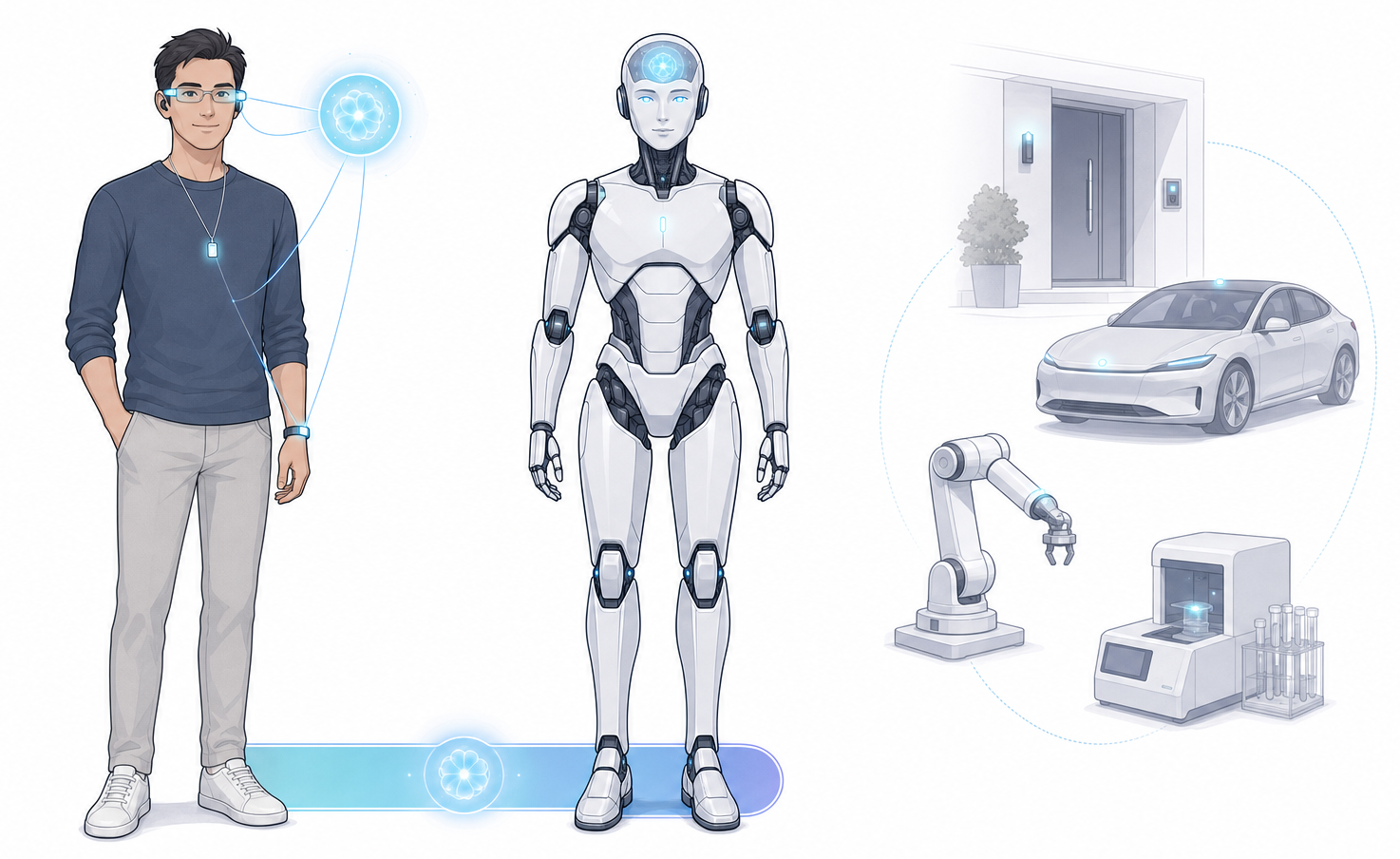}
  \caption{Two forms of intelligent individuals.
  Left: an AI-augmented individual, a person with a symbiont that senses and speaks through wearable devices.
  Middle: an AI as individual, an embodied robot.
  The two share one cognitive core, indicated by the common bar.
  Right, in lighter tone: devices in the environment that either form can reach into (Section~\ref{sec:reach}).
  The paper's scope is the left form.
  ``AI as individual'' is a functional description and carries no claim about consciousness or legal personhood.}
  \label{fig:twoforms}
\end{figure}

On the horizontal axis, we foresee, there are two ways for an intelligent individual to exist in the future world of Physical AI (\figref{fig:twoforms}).
An \keyterm{AI-augmented individual} is a real person who gains a symbiont through wearables and other interfaces.
The person remains the subject: the goals are theirs, the body is theirs, and the AI borrows their senses and their hands.
An \keyterm{AI as individual} is an embodied robot with its own body, perception, actions, and goals.
The two forms matter equally for the future we describe, and this paper is concerned only with the first.
Our central claim about the pair is that they share a \keyterm{cognitive core}: long-term memory, a model of the situation and of other minds, timely intervention, and learning from experience.
They differ in three places only: whose goals, whose body, and whose responsibility.
The cognitive capabilities developed in Sections~\ref{sec:context} to~\ref{sec:growing} therefore transfer to embodied robots, and this is the source of the paradigm's generality; Section~\ref{sec:twoformsone} returns to it.
The third column of \figref{fig:twoforms}, AI-driven devices in the environment, appears in the paper only as the symbiont's extended reach into the physical world (Section~\ref{sec:reach}).

\subsection{Eighty Years of the Same Question}\label{sec:lineage}
Using machines to augment the human intellect is an old question, and \tabref{tab:lineage} places the paper inside that lineage.
\citet{licklider1960symbiosis} imagined people and computers thinking together in real time; \citet{engelbart1962augmenting} made the unit of augmentation the person together with their tools and methods; \citet{clark1998extended} argued that cognition can extend into external tools; \citet{weiser1991computer} and the first wearable computing researchers~\citep{starner1997augmented,mann1997wearable} put computing into daily life and onto the body; and lifelogging, from \citeauthor{bush1945think}'s Memex~\citeyearpar{bush1945think} through MyLifeBits~\citep{gemmell2006mylifebits} and SenseCam~\citep{hodges2006sensecam}, recorded a life and showed that the record could support memory.
More recent neighbors are egocentric AI~\citep{grauman2022ego4d,yang2025egolife}, mixed-initiative interaction~\citep{horvitz1999mixed,amershi2019guidelines}, just-in-time adaptive interventions in mobile health~\citep{nahumshani2018jitai}, the thought-partner program of \citet{collins2024machines}, and the lifecycle view of context engineering~\citep{hua2025context}.
For each line the table records what has been settled and what we ask next, and each of those questions is taken up in a later section.
Our claim with respect to these lines is a narrow one: none of them has taken, as its target, the augmentation of a particular person across a life.

\begin{table}[t]
\centering\footnotesize
\caption{Lineage and nearest neighbors. Each line of work has settled something this paper relies on, and the fourth column states the question we take up from it.}
\label{tab:lineage}
\setlength{\tabcolsep}{4pt}\renewcommand{\arraystretch}{1.22}
\begin{tabularx}{\linewidth}{@{}L{2.75cm}L{2.75cm}YYC{1.1cm}@{}}
\toprule
\tabhead \colhead{Line of work} & \colhead{Representatives} & \colhead{What it has already settled} & \colhead{What this paper asks next} & \colhead{Taken up in} \\
\midrule
\chipx{tbGray}{1945}\ Lifelogging & \citet{bush1945think}; \citet{gemmell2006mylifebits}; \citet{hodges2006sensecam} & recording a life, many views, memory support & how a record turns into understanding of a person and timely help & \sch{sec:context} \\
\rowcolor{tabStripe} \chipx{tbGray}{1960}\ Human--computer symbiosis & \citet{licklider1960symbiosis} & people and machines complementing each other in real time, thinking together & what symbiosis looks like once machines understand language and images & \sch{sec:living} \\
\chipx{tbGray}{1962}\ Augmenting intellect & \citet{engelbart1962augmenting} & the unit of augmentation is the person plus tools plus methods & how the research agenda changes once that unit is the object of evaluation & \sch{sec:agenda} \\
\rowcolor{tabStripe} \chipx{tbGray}{1991}\ Ubiquitous and wearable computing & \citet{weiser1991computer}; \citet{starner1997augmented,mann1997wearable} & computing woven into daily life; the first-person viewpoint of wearables & what unnoticed computing can do once a model understands content & \sch{sec:sensing} \\
\chipx{tbGray}{1998}\ Extended mind & \citet{clark1998extended} & cognition can extend into tools outside the body & where the boundary lies for a tool that learns and takes initiative & \sch{sec:concepts} \\
\rowcolor{tabStripe} \chipx{tbGray}{1999}\ Mixed-initiative interaction & \citet{horvitz1999mixed}; \citet{amershi2019guidelines} & principles for when to act and when to wait & how these principles are rewritten for an all-day companion & \sch{sec:living} \\
\chipx{tbGray}{2018}\ Just-in-time adaptive interventions & \citet{nahumshani2018jitai} & the right intervention, at the right strength, at a receptive moment, for health behavior & extending the framework from health to cognition and social life, driven by a model of the person & \sch{sec:ladder} \\
\rowcolor{tabStripe} \chipx{tbGray}{2022}\ Egocentric AI & \citet{grauman2022ego4d}; \citet{engel2023project}; \citet{yang2025egolife} & understanding activity from first-person video; answering questions about the past & from seeing what you see to knowing what you are thinking and need & \sch{sec:understanding} \\
\chipx{tbGray}{2024}\ Thought partners & \citet{collins2024machines} & AI should model the person and the world and think with them & putting joint thinking into real life, wearable hardware, and a lifelong timescale & \sch{sec:thinking} \\
\rowcolor{tabStripe} \chipx{tbGray}{2025}\ Context engineering & \citet{hua2025context} & a lifecycle of collecting, managing, and using context; four stages of evolution & what the lifecycle needs when the context is a person's life & \sch{sec:context} \\
\chipx{tbGray}{2026}\ Personal agent products & \citet{meta2026muse}; \citet{openai2026dots} & long-running, personalized, acting on your behalf & the step from doing things \emph{for} you to thinking \emph{with} you & \sch{sec:tier1} \\
\bottomrule
\end{tabularx}
\tablegend{\textbullet\ The year chip marks the earliest representative work of each line.\quad \textbullet\ The last column names the section that takes up the question.}
\end{table}

\subsection{Why Now}\label{sec:whynow}
Five conditions turn an eighty-year-old question into a tractable research problem, and each is developed in the next section.
Foundation models understand language, images, and sound and can reason over them, so that a symbiont has, for the first time, a brain (Section~\ref{sec:fm}).
Agent technology lets a model plan, use tools, operate a computer, and run for a long time, so that it has hands (Section~\ref{sec:agents}).
Long-term memory, retrieval, and context management let it keep an understanding of one person across months and years (Section~\ref{sec:memory}).
World models let it begin to predict what the environment and the person will do next (Section~\ref{sec:wm}).
And wearable devices have matured to the point where first-person audio and video can be captured all day and a reply can be delivered to the ear with low latency, so that the symbiont can have senses and be present (Section~\ref{sec:hardware}).
What has not changed matters equally: the social infrastructure of privacy and trust, the observability of inner states, the stability of continual learning, and the cost of computing around the clock.
These are the sources of Sections~\ref{sec:trust} and~\ref{sec:agenda}.

\section{Technical Foundations}\label{sec:foundations}
\begin{figure}[t]
  \centering
  \resizebox{0.98\linewidth}{!}{\begin{tikzpicture}[x=1cm,y=1cm,font=\small\sffamily,
  layer/.style={draw=#1,fill=#1!10,rounded corners=4pt,minimum width=9.6cm,minimum height=1.05cm,align=center,line width=0.7pt},
  have/.style={font=\scriptsize\sffamily,text=symSlate,text width=3.1cm,align=left,anchor=west},
  add/.style={font=\scriptsize\sffamily,text=symMagenta!80!black,text width=3.1cm,align=left,anchor=west}]
  \node[layer=symMagenta,fill=symMagenta!8,line width=1pt] (l5) at (0,5.2) {\textbf{Symbiont layer}\\[-1pt]{\scriptsize record of context and knowledge $\cdot$ personal model $\cdot$ intervention policy $\cdot$ trust}};
  \node[layer=stRuntime] (l3) at (0,2.6) {\textbf{Personal runtime}\\[-1pt]{\scriptsize identity, permissions, device abstraction, continuity, attention budget, one entry point}};
  \node[layer=stLive] (l4) at (0,3.9) {\textbf{Agentic harness}\\[-1pt]{\scriptsize tool loop, planning, memory, context management, delegation, sandboxing, skills}};
  \node[layer=stUnderstand] (l2) at (0,1.3) {\textbf{Foundation models}\\[-1pt]{\scriptsize language $\to$ omni-modal $\to$ reasoning $\to$ world models (one axis that keeps getting stronger)}};
  \node[layer=stSense] (l1) at (0,0) {\textbf{Wearable and embedded hardware}\\[-1pt]{\scriptsize glasses, pendant, earbuds, whisper mic, ring and band, phone, desktop, on-device compute}};
  \foreach \a/\b in {l1/l2,l2/l3,l3/l4,l4/l5}{\draw[-{Stealth[length=1.8mm]},symSlate,line width=0.6pt] (\a.north) -- (\b.south);}
  \node[font=\scriptsize\sffamily\bfseries,text=symSlate,anchor=south west] at (5.1,5.75) {today};
  \node[font=\scriptsize\sffamily\bfseries,text=symMagenta!80!black,anchor=south west] at (8.5,5.75) {this paper adds};
  \node[have] at (5.1,5.2) {none as a system; pieces in memory and personalization research};
  \node[add] at (8.5,5.2) {the whole layer (Sections 4--12)};
  \node[have] at (5.1,2.6) {single apps; cloud agents; early device SDKs};
  \node[add] at (8.5,2.6) {a runtime for a life, not an app (Section 11)};
  \node[have] at (5.1,3.9) {task-level harnesses, request-driven, API tools};
  \node[add] at (8.5,3.9) {event-driven, memory-native, silence as output (Section 11.3)};
  \node[have] at (5.1,1.3) {strong and improving; no persistence, no body, trained on the average person};
  \node[add] at (8.5,1.3) {an information-level interface so any backbone fits (Section 7.4)};
  \node[have] at (5.1,0) {first-generation glasses, pendants, bands, on-device models};
  \node[add] at (8.5,0) {designed as one constellation of senses and voice (Section 5)};
\end{tikzpicture}}
  \caption{The stack a symbiont is built on.
  From the bottom: wearable and embedded hardware, foundation models (one axis that runs from language to omni-modal to reasoning to world models), the personal runtime, the agentic harness, and the symbiont layer that this paper proposes.
  The side notes record what exists today and what the paper adds at each level.}
  \label{fig:stack}
\end{figure}
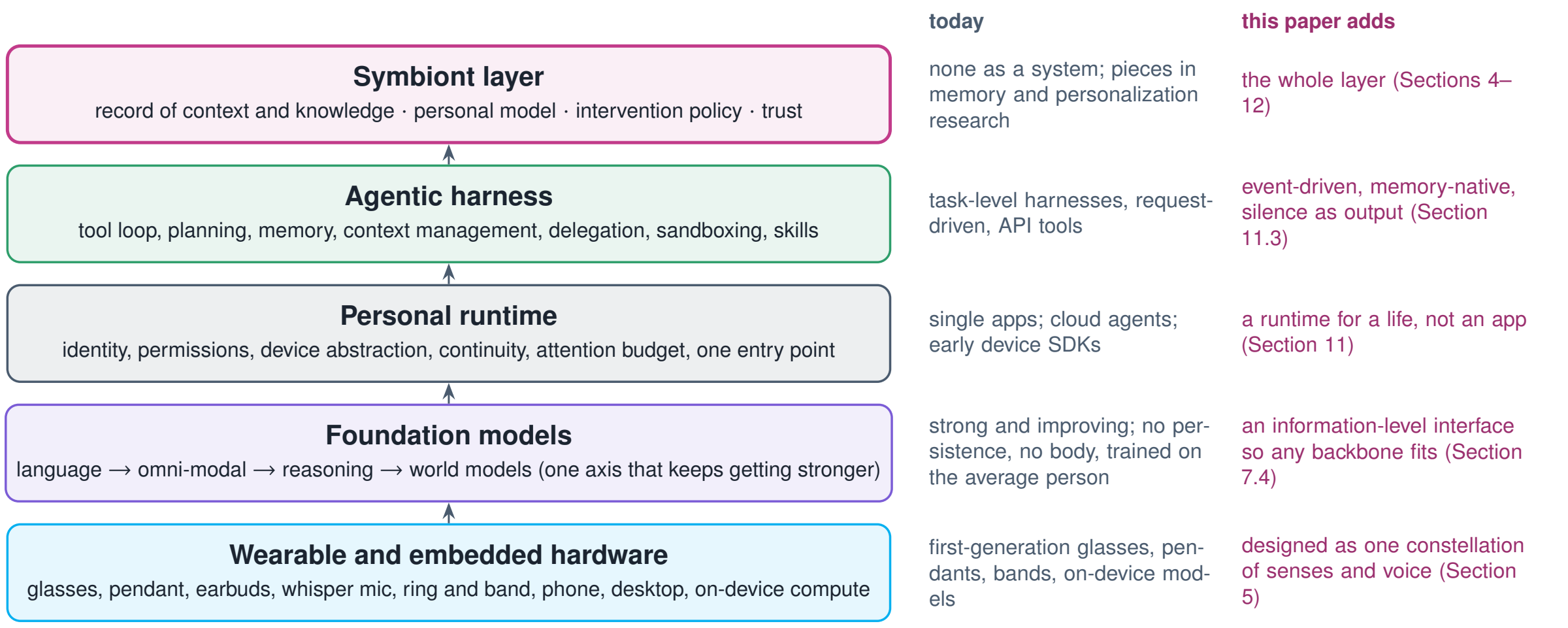

A symbiont is built on technologies that already exist and are improving quickly.
This section reads five of them for one purpose: to state, for each, what it already provides and what it still lacks.
The gaps, collected in \tabref{tab:foundations}, are the research space of the rest of the paper.
\figref{fig:stack} shows how the five lines stack up and where the layer this paper proposes sits.
We keep to a few representative works per line, report no benchmark numbers, and end every subsection with what the line means for a symbiont.

\subsection{Foundation Models}\label{sec:fm}
The brain of a symbiont comes from four steps in the development of foundation models.
Scale came first: the Transformer~\citep{vaswani2017attention}, in-context learning at the scale of GPT-3~\citep{brown2020language}, and the scaling laws that made the next steps predictable~\citep{kaplan2020scaling,hoffmann2022training}.
Alignment made the models usable by ordinary people in ordinary language~\citep{ouyang2022training}, and GPT-4~\citep{openai2023gpt4} together with openly released weights~\citep{touvron2023llama2} spread the capability.
Reasoning followed, from chain-of-thought prompting~\citep{wei2022chain} to models that think before they answer~\citep{openai2024o1systemcard,guo2025deepseekr1}; this is what makes the slow loop of Section~\ref{sec:loops} possible.
The fourth step is the one that matters most here.
Multimodal and then omni-modal models took images into the language model~\citep{radford2021learning,alayrac2022flamingo,liu2023visual}, then speech and video in real time~\citep{openai2024gpt4osystemcard,geminiteam2024gemini15}, then any-to-any input and output~\citep{wu2024nextgpt,xu2025qwen25omni}, and full-duplex speech models learned to listen and speak at the same time~\citep{defossez2024moshi}.
Context windows of a million tokens made the inclusion of an entire day in the context window expensive rather than impossible.
For a symbiont, omni-modality is a requirement: its input is a first-person stream of sight and sound and its output is a whisper or a line of text, and a model that only reads and writes text never reaches the scene.
What foundation models lack can be stated with equal precision.
They have no persistence, since each conversation starts from nothing; no sense of time; no body; and they are trained on the average person and know nothing about the one in front of them.
Sections~\ref{sec:context}, \ref{sec:living}, \ref{sec:sensing}, and~\ref{sec:understanding} answer these four gaps in turn.

\subsection{Agents and Harnesses}\label{sec:agents}
An \keyterm{agent} is a model system that repeatedly observes, thinks, and acts toward a goal.
A \keyterm{harness} is everything outside the model that turns it into an agent: the loop itself, tool interfaces, memory, context management, planning and decomposition, permissions and sandboxing, verification of results, and reusable procedural knowledge, often called skills.
The same model placed in different harnesses performs very differently, a fact that recurs across published results on coding agents, and it makes the harness a research object in its own right.
The line of work runs from interleaving reasoning with action~\citep{yao2023react}, learning to call tools~\citep{schick2023toolformer}, and reflecting on failure~\citep{shinn2023reflexion}, to agents that build a library of reusable skills~\citep{wang2024voyager}, which is an early precedent for a harness that improves itself (Section~\ref{sec:stack}).
Generative Agents gave agents a memory stream, reflection, and planning~\citep{park2023generative}, which remains the closest academic prototype of an agent with a life, and CoALA organized these components in the language of cognitive architectures~\citep{sumers2024coala}.
On the engineering side, Anthropic's guide to building agents argued for simple, composable patterns and for the agent--computer interface as something to be designed~\citep{schluntz2024building}; SWE-agent showed that an interface built for the agent changes its performance~\citep{yang2024sweagent}; computer-use agents learned to operate any desktop or web page~\citep{anthropic2024computeruse,openai2025operator}; the Model Context Protocol standardized tool access~\citep{anthropic2024mcp}; and coding agents such as Claude Code and Codex became the first autonomous agents in large-scale daily use~\citep{anthropic2025claudecode,openai2025codexcli}.
In 2026 Muse and Dots pushed agents into personal life, and the Muse Gadget SDK for embedded platforms is a sign that agent platforms are reaching into devices~\citep{meta2026muse,meta2026musegadgetsdk,openai2026dots}.
The benchmarks of the field all score task completion~\citep{zhou2024webarena,xie2024osworld,jimenez2024swebench,yao2025taubench,mialon2024gaia}.

For a symbiont this line provides hands.
Booking a trip, buying something, handling a desktop, writing code: today's agents can already do these, and a symbiont delegates such work through its harness (Section~\ref{sec:tier1}).
Beyond delegation, a symbiont \emph{is} a new kind of harness, one organized around a life instead of a task (Section~\ref{sec:harness}).
Today's harnesses are task-level: they have a beginning and an end, they are driven by requests, their tools are APIs, their memory is scoped to the task, and their score is completion.
They have no body, no first-person senses, no model of the person, and no output that means ``do nothing.''
Sections~\ref{sec:living}, \ref{sec:ladder}, and~\ref{sec:runtime} take up these gaps.

\subsection{Memory, Context, and Personalization}\label{sec:memory}
Three kinds of work let a model remember a person.
Memory systems grew from retrieval-augmented generation~\citep{lewis2020retrieval} to MemGPT, which brought the tiered storage of an operating system into the model~\citep{packer2023memgpt}; to the memory stream and reflection of Generative Agents~\citep{park2023generative} and MemoryBank~\citep{zhong2024memorybank}; to A-MEM, which lets memories link to one another and evolve~\citep{xu2025amem}; to Mem0, a production-oriented long-term memory layer~\citep{chhikara2025mem0}; and to sleep-time compute, which organizes memory while the system is idle~\citep{lin2025sleeptime}.
LoCoMo and LongMemEval measure long-conversation memory~\citep{maharana2024locomo,wu2025longmemeval}.
Context management is the second kind: Context Engineering 2.0 describes a lifecycle of collecting, managing, and using context, and observes that the practice is shifting from prompting a model toward managing everything a long-running system knows~\citep{hua2025context}; long context, cache compression, hierarchical summaries, and on-demand loading are the engineering means.
Personalization is the third: LaMP measures generation tailored to a user~\citep{salemi2024lamp}; parameter-efficient adaptation makes a small module per user feasible~\citep{hu2022lora}; test-time training and its descendants change weights during use~\citep{sun2020test,sun2025learning,behrouz2025titans}; and the continual-learning literature deals with forgetting.
These are the direct foundation of Section~\ref{sec:context} and the toolbox of Section~\ref{sec:growing}, and this paper does not require a new memory mechanism.
What the line lacks follows from what it remembers, which is dialogue rather than a life.
There is no first-person stream of sight and sound, no physiology, no distinction between what the person said and what the system inferred (Section~\ref{sec:tworoads}), no views that switch with the question (Section~\ref{sec:views}), no distinction between short-lived context and settled knowledge (Section~\ref{sec:timescales}), no consistent treatment of forgetting and deletion (Section~\ref{sec:repair}), and the personalization it offers optimizes for sounding like the person, which is a different thing from understanding them (Section~\ref{sec:wherelearning}).

\subsection{World Models}\label{sec:wm}
A world model is an internal model that predicts what happens next.
One thread comes from reinforcement learning, where World Models and the Dreamer family plan in imagination~\citep{ha2018world,hafner2020dream,hafner2025mastering}.
A second is the predictive-representation route of JEPA and V-JEPA~2~\citep{lecun2022path,assran2025vjepa2}.
A third is generative world simulation at the pixel level: Genie and Genie~3, Sora, and Cosmos~\citep{bruce2024genie,parkerholder2025genie3,brooks2024video,nvidia2025cosmos}.
A fourth is first-person data, which makes the world as seen through a person's eyes a learnable object: Ego4D~\citep{grauman2022ego4d}, Project Aria and its second-generation research glasses~\citep{engel2023project,meta2025ariagen2}, and EgoLife~\citep{yang2025egolife}.
A parallel thread models people.
Large foundation models show real but brittle theory-of-mind ability~\citep{kosinski2024evaluating,ullman2023large,strachan2024testing}.
\citet{park2024simulation} built a generative agent for each of 1,052 real people from a two-hour interview and found that the agents reproduced their participants' survey answers about as well as the participants reproduced their own answers two weeks later; it is the strongest evidence to date that a predictive model of a particular person can be built from a small amount of first-person narrative.
Mental world modeling treats beliefs, intentions, and other mental variables as part of the world state itself~\citep{fei2026mental}, which is the position this paper adopts for the mental side of the personal model (Section~\ref{sec:twoworldmodels}).
For a symbiont, world models provide the ability to think ahead: to predict what is about to happen around the person (an external, physical world model) and how the person will think, feel, and react, including to a reminder (an internal, mental world model).
Sections~\ref{sec:backbone} and~\ref{sec:twoworldmodels} show how both enter the personal model and why the system still stands, in a degraded form, without them.
Today's world models are third-person, short-horizon, and built once; the models of people are built from a single interview, never updated, and unable to tell that they have gone stale.

\subsection{Wearable and Embedded Hardware}\label{sec:hardware}
In three years the hardware conditions for senses and presence have changed in four places.
On the head, Ray-Ban Meta glasses showed that ordinary people will wear a camera and microphones all day~\citep{meta2023raybanmeta}; the Orion prototype and Meta Ray-Ban Display added a display~\citep{meta2024orion,meta2025raybandisplay}, lightweight display glasses such as the Even G1 appeared~\citep{evenrealities2024g1}, and Project Aria serves as a research platform~\citep{engel2023project}.
On the body, pendants and recorders capture all-day audio~\citep{mehta2024limitless,silberling2025bee,plaud2024notepin}, rings and bands capture physiology, and Meta's surface-EMG wristband, published in \textit{Nature}, trains a generic model on data from thousands of people so that a new wearer can type and gesture with no calibration, and twenty minutes of personal data improve it further~\citep{kaifosh2025generic}.
This is a ready-made example of a generic model with a small personal adaptation on the input side, and Section~\ref{sec:sparse} returns to it.
Closer to the body, AlterEgo reads the muscle signals of silent, deliberate speech~\citep{kapur2018alterego}, and brain--computer interfaces, noninvasive~\citep{levy2025brain} and implanted~\citep{willett2023speech,card2024speech}, remain in research and early clinical stages.
In compute, neural accelerators in phones and glasses, together with quantization and distillation, run billion-parameter models on the device~\citep{gunter2024apple}; low-power always-on event detection is the premise of the duty cycle in Section~\ref{sec:alwayson}; and full-duplex speech models make listening while speaking feasible in latency.
The counterexamples are instructive as well: the Humane AI Pin, a dedicated assistant device launched in 2024, was discontinued within a year~\citep{zeff2025humane}, which suggests that a new device has to offer something the phone in the person's pocket does not.
The first generation of senses, private output, and local compute therefore exists, and Section~\ref{sec:sensing} can discuss design.
What is missing can be listed: battery and heat limit what ``all day'' means, display glasses are far from a shared canvas, there is no mature model from physiology to mental state, silent speech and neural interfaces are at an early stage, and above all no device has been designed as one sense of one symbiont, since each is a separate product.

\subsection{What Is Still Missing}\label{sec:missing}
\tabref{tab:foundations} collects the five lines, what each gives, and what each lacks, and maps every gap to the section that addresses it.
The five lines are each growing stronger, and none of them has, as its goal, the augmentation of one particular person.
Organizing them toward that goal is the task of the conceptual framework in the next section.

\begin{table}[t]
\centering\footnotesize
\caption{Five lines of technology, what each already gives a cognitive symbiont, and what it lacks. The last column maps every gap to the sections that address it; the gaps are the research space of Part~II.}
\label{tab:foundations}
\setlength{\tabcolsep}{4pt}\renewcommand{\arraystretch}{1.22}
\begin{tabularx}{\linewidth}{@{}L{2.3cm}L{3.5cm}YYL{1.95cm}@{}}
\toprule
\tabhead \colhead{Line of work} & \colhead{Representative work} & \colhead{What it gives a symbiont} & \colhead{What it still lacks} & \colhead{Answered in} \\
\midrule
\icUnderstand\ \rowhead{Foundation models} & GPT-3/4, Llama, o1, R1, GPT-4o, Gemini, NExT-GPT, Qwen-Omni, Moshi & a brain that understands language, images, and sound, and reasons; omni-modal input and output & no persistence, no sense of time, no body; trained on the average person & \sch{sec:context} \sch{sec:sensing} \sch{sec:understanding} \sch{sec:living} \\
\rowcolor{tabStripe} \icLive\ \rowhead{Agents and harnesses} & ReAct, Reflexion, Voyager, Generative Agents, CoALA, SWE-agent, computer use, MCP, Claude Code, Codex, Muse, Dots & hands: planning, tool use, computer operation, long-running tasks; general-purpose task ability & task-level and request-driven; no senses; no model of the person; no ``do nothing'' output & \sch{sec:living} \sch{sec:tier1} \sch{sec:harness} \\
\icKeep\ \rowhead{Memory and context} & RAG, MemGPT, A-MEM, Mem0, sleep-time compute, Context Engineering 2.0, LoRA, test-time training & memory across sessions; a context lifecycle; a toolbox for personalization & remembers dialogue, not life; no split between self-report and inference; no multi-view record; no context/knowledge distinction; deletion unsolved & \sch{sec:context} \sch{sec:tworoads} \sch{sec:growing} \\
\rowcolor{tabStripe} \icGrow\ \rowhead{World models} & World Models, Dreamer, JEPA, Genie, Cosmos, Ego4D, Aria, EgoLife; theory-of-mind studies; 1,000-person simulations; mental world modeling & prediction of what the environment and the person will do next & third-person, short-horizon, one-shot modeling of people; cannot detect that it has gone stale & \sch{sec:backbone} \sch{sec:twoworldmodels} \sch{sec:limits} \\
\icSense\ \rowhead{Wearable and embedded hardware} & Ray-Ban Meta, Ray-Ban Display, sEMG band, Aria, AlterEgo, on-device models, full-duplex speech & wearable senses, private output, local compute & battery, display, interpretation of physiology; devices are separate products, not one set of senses & \sch{sec:sensing} \sch{sec:economics} \\
\bottomrule
\end{tabularx}
\tablegend{\textbullet\ Badges mark the stage of the symbiosis loop (\figref{fig:loop}) that each line mostly serves: \icSense\ sense and interact, \icKeep\ keep, \icUnderstand\ understand, \icLive\ live alongside, \icGrow\ grow.}
\end{table}

\section{Conceptual Foundations}\label{sec:concepts}
This section gives the vision a theoretical shape.
We first define the objects, then state the invariant that runs through the paper, then list the properties a symbiont must have and test them against the fictional reference everyone knows, and finally draw the map that organizes the research space.
The pattern throughout is definition, formalization, example.

\subsection{Definitions}\label{sec:definitions}
Five definitions turn the slogan ``an AI that knows you better than you know yourself'' into a claim that can be examined and disputed.

\begin{definition}{Self-context}{selfcontext}
The self-context of a person $h$ at time $t$, written $\Ctx^h_t$, is the set of all facts bearing on their situation: what they have experienced, what they know, what they have committed to, whom they are related to and how, the state of their body, where they are, and what they are doing.
It has three parts: what can be observed from outside, $O_{\le t}$; what the person has said themselves, $R_{\le t}$; and what can be inferred from the two, $K_t$.
\end{definition}

\begin{definition}{Working context}{workingcontext}
At any moment, the part of $\Ctx^h_t$ that $h$ can actually bring to bear is a small subset $\wctx_t \subset \Ctx^h_t$, limited by working memory, attention, time, and emotion.
\end{definition}

\begin{definition}{Cognitive gap}{gap}
For a concrete situation $x_t$, with $\Util$ the outcome the person achieves in it,
\begin{equation}\label{eq:gap}
\gap_t \;=\; \Util\!\left(x_t;\, \Ctx^h_t\right) \;-\; \Util\!\left(x_t;\, \wctx_t\right),
\end{equation}
the difference between what the person could achieve with their whole self-context and what they achieve with the small part they can hold at the time.
\end{definition}

The cognitive gap is the invariant of the paper.
It measures something other than intelligence: the price every person pays, every day, in what they forgot, failed to notice, could not think through, or could not put into words.
A researcher who, writing a new paper, cannot recall an argument she made in a blog post three years ago is paying it, and so is a parent who snaps at a child because the sleepless night is not in working memory as an explanation.
\claim{The job of a symbiont is to shrink $\gap_t$: to return to the person, at the right moment, the part of their own context that is out of reach.}
Mind reading is not required; what is required is that the right slice of $\Ctx^h_t$ be placed into $\wctx_t$, or that the reasoning or the errand the person would have undertaken, given the time, be carried out on their behalf.

\begin{definition}{Cognitive symbiont}{symbiont}
A cognitive symbiont is an AI system $s$ that (i) exists continuously across the life of a person $h$; (ii) maintains, from authorized observation and from $h$'s own statements, a model $M_t \approx \Ctx^h_t$ of $h$'s world; (iii) is present during $h$'s ongoing activities through a private, two-way channel; (iv) at each moment chooses one way of taking part from \{silence, hint, suggestion, question, action on $h$'s behalf\} so as to reduce $\gap_t$; (v) learns from outcomes; and (vi) leaves final judgment and goals with $h$.
\end{definition}

Condition (vi) is also where this paper draws the boundary that the extended mind~\citep{clark1998extended} leaves open for a tool that learns and takes initiative: the symbiont extends what the person can bring to a situation, and judgment and goals stay with the person.

\begin{definition}{Augmented individual}{augmented}
The augmented individual $h \compose s$ is the unit of evaluation of this paper, with the augmentation gain in a situation defined as
\begin{equation}\label{eq:aug}
\daug(x_t) \;=\; \Util\!\left(x_t;\, h \compose s\right) \;-\; \Util\!\left(x_t;\, h\right).
\end{equation}
\end{definition}

The definitions settle the boundary with AGI in one line: AGI raises $\Util(\cdot;\, s)$, and AHI raises $\Util(\cdot;\, h \compose s)$.
Two cautions apply to the notation.
$\Util$ is a situation-specific outcome and is never summed across situations, and $\compose$ marks a coupling condition rather than an arithmetic operation.
The formulas serve to remove ambiguity, and no theorem is derived from them.

\subsection{Six Properties and the Jarvis Test}\label{sec:properties}
\tabref{tab:properties} lists six properties a symbiont must have.
Four of them, private, real-time, low-friction, and embedded, come directly from what makes the system usable in the middle of a life; lifelong and proactive are added because without the first the gap cannot close across time, and without the second the person cannot ask about what they do not know they are missing.
The properties pull against one another: being proactive conflicts with being low-friction, and being lifelong conflicts with being private.
Those tensions are exactly the research problems of Sections~\ref{sec:living} and~\ref{sec:trust}.
``All day'' is read throughout as availability, never as continuous computation (Section~\ref{sec:alwayson}).

\begin{table}[t]
\centering\footnotesize
\caption{Six properties of a cognitive symbiont, and the Jarvis test. Each property is necessary; the third column says what fails without it. The properties pull against one another (proactive against low-friction, lifelong against private), and those tensions are the research problems of Sections~\ref{sec:living} and~\ref{sec:trust}.}
\label{tab:properties}
\setlength{\tabcolsep}{4pt}\renewcommand{\arraystretch}{1.28}
\begin{tabularx}{\linewidth}{@{}L{2.5cm}YYC{0.95cm}C{1.25cm}L{1.55cm}@{}}
\toprule
\tabhead \colhead{Property} & \colhead{Meaning} & \colhead{What breaks without it} & \colhead{Jarvis} & \colhead{Personal agents today} & \colhead{Sections} \\
\midrule
\ib{tbAmber}{\faHistory}\ \rowhead{Lifelong} & one relationship and one memory kept across years & every session starts from zero; the gap cannot close & \cmark & \pmark & \sch{sec:context} \sch{sec:growing} \\
\rowcolor{tabStripe}\ib{tbTeal}{\faStreetView}\ \rowhead{Embedded} & present in real activity, not waiting in a chat window & help can only happen afterwards & \cmark & \xmark & \sch{sec:sensing} \sch{sec:runtime} \\
\ib{tbGreen}{\faStopwatch}\ \rowhead{Real-time} & steps in while the activity is still going and help is still useful & the right words at the wrong time are no words & \cmark & \xmark & \sch{sec:living} \\
\rowcolor{tabStripe}\ib{tbViolet}{\faLock}\ \rowhead{Private} & what passes between the person and the symbiont stays theirs, and its delivery can be discreet where the setting allows & the person cannot confide in it, and every cue disturbs the people nearby & \cmark & \pmark & \sch{sec:sensing} \sch{sec:living} \\
\ib{tbBlue}{\faFeather}\ \rowhead{Low-friction} & never asks you to stop, take out a phone, or explain the background & the cost of use eats the benefit & \cmark & \xmark & \sch{sec:living} \sch{sec:runtime} \\
\rowcolor{tabStripe}\ib{tbPink}{\faBell}\ \rowhead{Proactive} & can judge whether to speak when you have not asked & you do not ask about what you do not know you are missing & \cmark & \pmark & \sch{sec:living} \\
\bottomrule
\end{tabularx}
\tablegend{\textbullet\ \cmark\ has the property\quad \pmark\ partly (memory across sessions, a phone that must be taken out, notifications)\quad \xmark\ lacks it.\quad \textbullet\ The Jarvis column is argued in the vignette \emph{Beyond Jarvis}.}
\end{table}

\begin{scenario}[label={box:jarvis}]{Beyond Jarvis}
\small\textbf{What Jarvis already is.}
Jarvis passes all six properties.
It has known its owner for years (lifelong); it lives in the laboratory, the house, and the suit (embedded); it reports and warns during combat (real-time); the voice inside the helmet is heard by its owner alone (private); it never asks him to stop and type or explain (low-friction); and it speaks up about incoming threats, energy, and injuries unasked (proactive).
\textbf{Where a symbiont goes beyond it.}
(1) \emph{Senses.} Jarvis sees through infrastructure its owner built; a symbiont sees through a few affordable wearables, with channels Jarvis never had (Section~\ref{sec:sensing}).
(2) \emph{Mind.} Jarvis knows schedules, systems, and the battlefield, while the center of gravity of a symbiont is the owner's goals, emotions, and blind spots (Section~\ref{sec:understanding}).
(3) \emph{Cognitive help.} Jarvis executes and alerts; it rarely helps its owner make a life decision, recover from a setback, or deal with people and cultures, which is where a symbiont is most distinctive (Sections~\ref{sec:thinking} to~\ref{sec:caring}).
(4) \emph{Growth.} Across the films Jarvis's understanding of its owner barely changes; a symbiont must understand the person better the longer the two live together, and the whole system must evolve with it (Section~\ref{sec:growing}).
(5) \emph{Everyone.} Jarvis is one person's bespoke system; a symbiont is a paradigm for all (Section~\ref{sec:whom}).
(6) \emph{Trust.} Jarvis's loyalty is given by the script; a symbiont's must be given by the architecture (Section~\ref{sec:trust}).
The subject of this paper is everyone's Jarvis, six steps further on.
\end{scenario}

\subsection{The Symbiosis Loop}\label{sec:loop}
\figref{fig:loop} is the map of the research space and the table of contents of Part~II.
The person sits at the center.
Around them runs a loop: \emph{sense and interact} (borrowed senses, borrowed voice, and digital traces), \emph{keep} (live context, working state, and lifelong knowledge), \emph{understand} (a model of the person's situation, mind, relations, and goals, built on a replaceable backbone), \emph{live alongside} (real-time, private, proactive participation, with silence as one of the formal outputs), \emph{act for and think with} (the ladder of help from errands to thinking to social life, the help stage for short), and \emph{grow} (learning from outcomes and updating the record, the model, the policy, and eventually the whole stack).
Beneath the loop is the \emph{runtime} that makes it run across devices and time; around it is \emph{trust}, which governs authorization, privacy, correction, and exit.
One feature of the figure is deliberate: the person's own words enter the record directly, without passing through inference.
Of the technology lines in Section~\ref{sec:foundations}, agents and memory research cover mostly the keep stage and the errand end of the help stage.
The weakest stages are the mental layer of understanding, the timing of participation, online learning and whole-stack evolution, and the treatment of trust as architecture.
The figure is a conceptual map rather than a system architecture, and it does not require a monolithic implementation.

\subsection{Scope and Non-Claims}\label{sec:nonclaims}
The paper does not claim that an AI can read thoughts that were never expressed; that everyone should wear devices all day; that a complete context is a precondition for the system to work (Section~\ref{sec:trustcurve} shows it must be useful under partial context); that a symbiont outperforms the person in every domain; that the four roles of Section~\ref{sec:roles} are inevitable historical stages; that any one foundation model, or any one way of organizing a system around it, is the only correct backbone (Section~\ref{sec:backbone}); that it proposes a form of foundation model or a route to general intelligence (Section~\ref{sec:otherhalf}); that a symbiont raises a trait such as intelligence or emotional skill (Section~\ref{sec:acceptable}); that its use is legitimate in every setting (Section~\ref{sec:settings}); or that any interface standard or protocol should be adopted.
It claims that this is a research direction worth organizing on its own, with its own invariant, properties, design space, and standard of evidence.

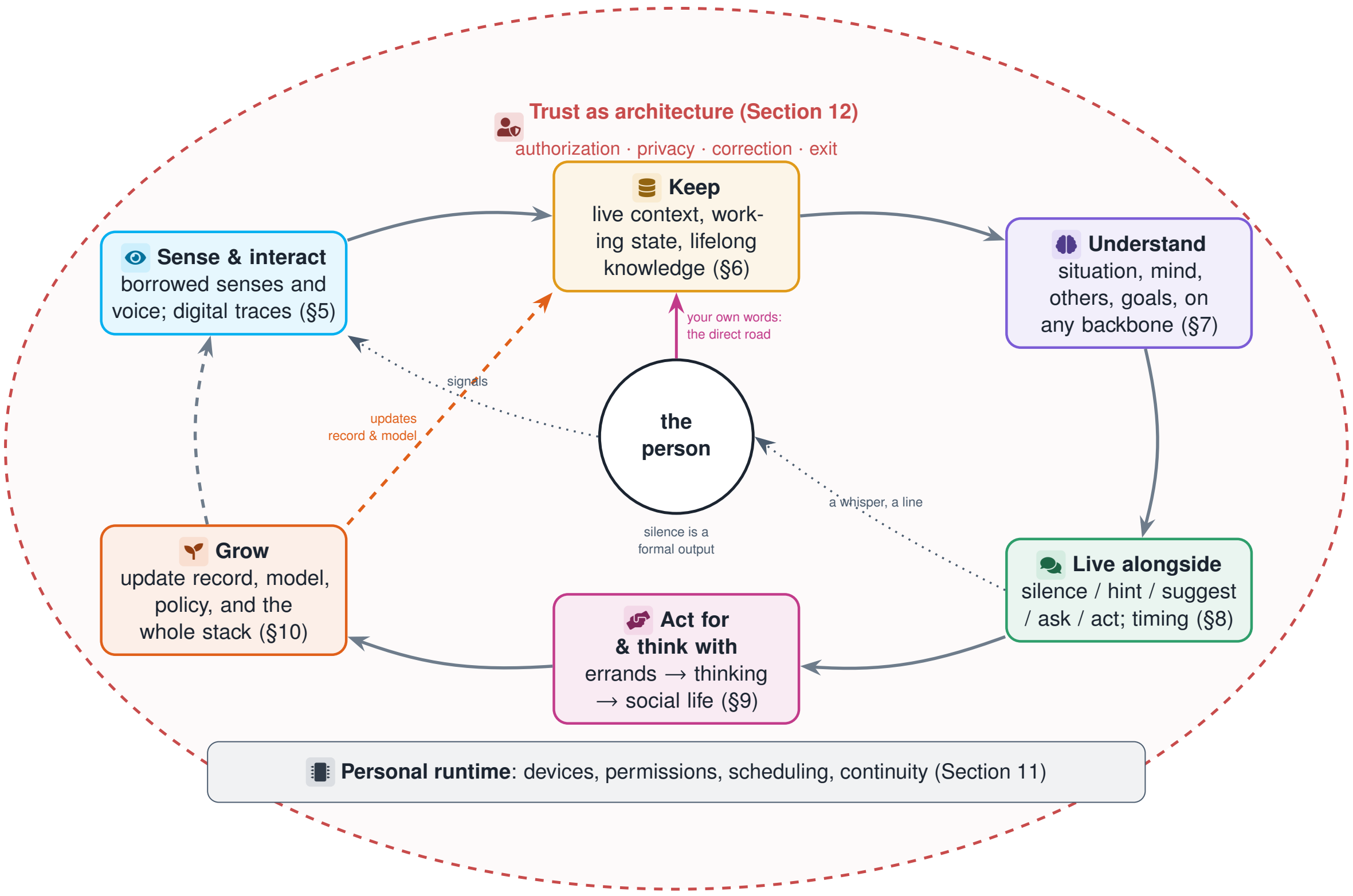
\begin{figure}[H]
  \centering
  \resizebox{0.98\linewidth}{!}{\begin{tikzpicture}[x=1cm,y=1cm,font=\small\sffamily,
  stage/.style={draw=#1,fill=#1!10,rounded corners=5pt,text width=2.75cm,align=center,inner sep=4pt,line width=0.9pt,font=\footnotesize\sffamily},
  arr/.style={-{Stealth[length=2.6mm]},line width=1.1pt,symSlate!80}]
  \node[draw=stTrust,dashed,line width=1pt,ellipse,minimum width=16.6cm,minimum height=10.9cm,fill=stTrust!3] (trust) at (0,0.05) {};
  \node[font=\footnotesize\sffamily\bfseries,text=stTrust,anchor=north,align=center] at (0,4.45) {\icTrust\ Trust as architecture (Section 12)\\[-1pt]{\scriptsize\mdseries authorization $\cdot$ privacy $\cdot$ correction $\cdot$ exit}};
  \node[draw=stRuntime,fill=stRuntime!8,rounded corners=4pt,minimum width=11.6cm,minimum height=0.75cm,font=\footnotesize\sffamily] (rt) at (0,-3.95) {\icRuntime\ \textbf{Personal runtime}: devices, permissions, scheduling, continuity (Section 11)};
  \node[circle,draw=symInk,fill=white,minimum size=1.9cm,line width=1pt,font=\footnotesize\sffamily\bfseries,align=center] (person) at (0,0.2) {the\\person};
  \node[stage=stSense] (sense) at (-5.6,2.1) {\icSense\ \textbf{Sense \& interact}\\borrowed senses and voice; digital traces (\S5)};
  \node[stage=stKeep] (keep) at (0,2.8) {\icKeep\ \textbf{Keep}\\live context, working state, lifelong knowledge (\S6)};
  \node[stage=stUnderstand] (und) at (5.6,2.1) {\icUnderstand\ \textbf{Understand}\\situation, mind, others, goals, on any backbone (\S7)};
  \node[stage=stLive] (live) at (5.6,-1.7) {\icLive\ \textbf{Live alongside}\\silence / hint / suggest / ask / act; timing (\S8)};
  \node[stage=stHelp] (help) at (0,-2.55) {\icHelp\ \textbf{Act for \& think with}\\errands $\to$ thinking $\to$ social life (\S9)};
  \node[stage=stGrow] (grow) at (-5.6,-1.7) {\icGrow\ \textbf{Grow}\\update record, model, policy, and the whole stack (\S10)};
  \draw[arr] (sense) to[bend left=12] (keep);
  \draw[arr] (keep) to[bend left=12] (und);
  \draw[arr] (und) to[bend left=14] (live);
  \draw[arr] (live) to[bend left=12] (help);
  \draw[arr] (help) to[bend left=12] (grow);
  \draw[arr,dashed] (grow) to[bend left=14] (sense);
  \draw[arr,stGrow,dashed] (grow.north east) -- node[pos=0.42,anchor=east,font=\tiny\sffamily,text=stGrow,align=right,xshift=-2pt] {updates\\record \& model} (keep.south west);
  \draw[arr,symMagenta,line width=1pt] (person.north) -- node[right,font=\tiny\sffamily,text=symMagenta,align=left] {your own words:\\the direct road} (keep.south);
  \draw[arr,symSlate,line width=0.7pt,dotted] (person.west) to[bend left=10] node[above,font=\tiny\sffamily,text=symSlate] {signals} (sense.south east);
  \draw[arr,symSlate,line width=0.7pt,dotted] (live.west) to[bend left=5] node[above,font=\tiny\sffamily,text=symSlate] {a whisper, a line} (person.east);
  \node[font=\tiny\sffamily,text=symSlate,align=center] at (0,-1.1) {silence is a\\formal output};
\end{tikzpicture}}
  \caption{The symbiosis loop.
  Six stages run around the person; the runtime beneath them makes the loop run across devices and time, and trust encloses everything.
  The person's own words enter the record directly.
  Silence is drawn as a formal output, and acting for and thinking with share one stage.}
  \label{fig:loop}
\end{figure}

\clearpage
\paperpart{The Anatomy of a Symbiont}\label{part:anatomy}
\begin{partlead}
\noindent Part~II follows the symbiosis loop of \figref{fig:loop} once around the person, and each of its sections takes the output of the previous one as its input.
Section~\ref{sec:sensing} asks what the symbiont senses the person with and what it speaks through, since it has no body of its own.
Section~\ref{sec:context} turns the resulting streams into a record with three timescales, one entry point, and many views.
Section~\ref{sec:understanding}, the technical center of the paper, asks how that record becomes an understanding of the person's situation and mind, on a backbone that can be replaced.
Section~\ref{sec:living} turns understanding into participation at the right moment, with silence as a formal output.
Section~\ref{sec:ladder} says what the symbiont does once present, on a ladder from errands to thinking to social life.
Section~\ref{sec:growing} asks how the system and the person change over many rounds, up to the evolution of the whole stack.
Beneath the loop, Section~\ref{sec:runtime} describes the runtime on which all of this runs, and around it, Section~\ref{sec:trust} treats trust as part of the architecture.
\end{partlead}

\section{Sensing and Interacting}\label{sec:sensing}
The loop of \figref{fig:loop} opens with the question of what the symbiont senses the person with and what it speaks to them through.
Hardware here is the symbiont's borrowed body: borrowed senses on the way in, a borrowed voice and a borrowed canvas on the way out.
The borrowed hands, devices in the environment, wait until Section~\ref{sec:reach}.

\subsection{Borrowed Senses, Borrowed Voice}\label{sec:borrowed}
An embodied robot comes with eyes, ears, and a mouth.
A symbiont has none of its own and must borrow three things: the sensors on a person's body, so that it sees what they see and hears what they hear; the person's digital traces on screens, in documents, in code, and in messages; and a channel of output that only the person receives, a voice in the ear, a line in the glasses, a picture laid over the world.
Its perception is therefore first-person, incomplete, and moving with the person, and its speech is close to the ear, short, and tied to the setting the person is in.
The quality of the senses sets the ceiling on understanding and the quality of the voice sets the floor on help, and neither is the goal in itself.
This is also the sense in which hardware and AI are the same project, which Section~\ref{sec:coevolve} develops.

\subsection{The Device Constellation}\label{sec:constellation}
\begin{figure}[t]
  \centering
  \resizebox{0.96\linewidth}{!}{\input{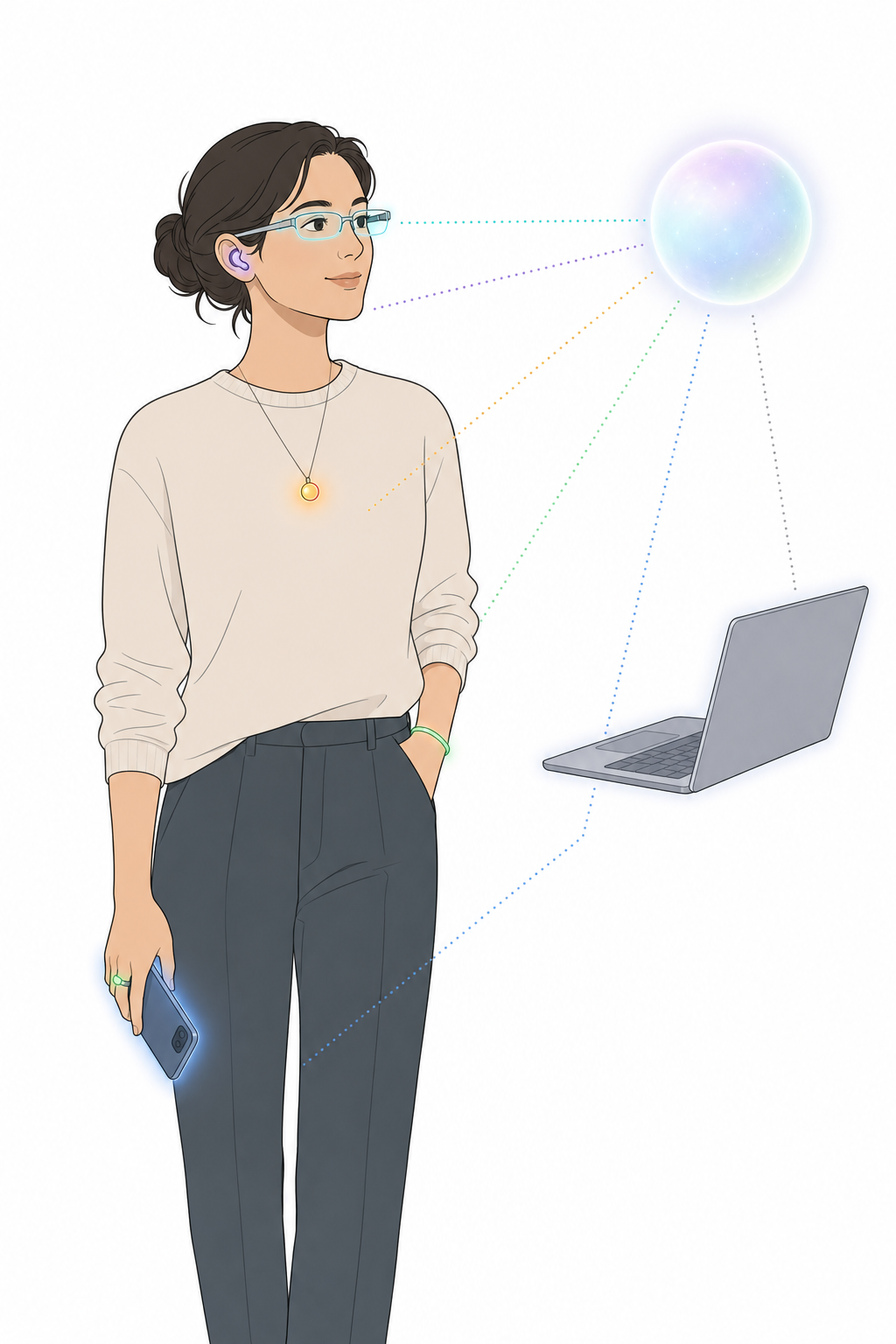}}
  \caption{The device constellation.
  Each device carries a role on the way in (sensing, private input) or on the way out (private output, shared canvas), and often both; a runtime manages them as a single set of senses.
  The symbiont itself has no body and borrows every one of them.
  Neural interfaces and the devices of the environment are the far end of the constellation.}
  \label{fig:devices}
\end{figure}

No single device can carry all of this, so a symbiont senses and speaks through a constellation (\figref{fig:devices}): glasses with a first-person camera, microphones, eye tracking, and a one-line display; a pendant or pin that records audio and takes snapshots all day and stands in when glasses are not worn; earbuds or bone conduction for private voice output; a microphone close to the mouth so that the person can speak to it at the level of a breath without being overheard; a ring, watch, or band for heart rate, sleep, and activity, and increasingly for EMG gestures; a phone for compute and a screen on the move; and a desktop with authorized access to screen content, the IDE, documents, and meeting software.
Each device carries a role on the way in, on the way out, or both, and a runtime (Section~\ref{sec:runtime}) manages them as one.
\tabref{tab:signals} lists the signal sources and what each reveals about the physical and the mental side of a life.

Two further classes of device belong in the constellation even though they are not yet in daily use, and the first of them may become the most valuable wearable of all.
Neural interfaces read the body's own signals of intention and attention.
At the surface of the body, EMG wristbands already decode the motor commands behind typing and gesture~\citep{kaifosh2025generic}, and electrodes in a headband or around the ear record the electrical activity of the brain, in many studies with results comparable to scalp recording~\citep{kaongoen2023ear}; implanted interfaces, now in early clinical trials, decode attempted speech at far higher fidelity~\citep{willett2023speech,card2024speech}.
For a symbiont these interfaces matter for a specific reason.
They are the channel through which a deliberate command could one day reach the symbiont without being spoken, typed, or acted out, at a lower cost to the person than any other channel.
What they may read is held to the rule of every input channel (Section~\ref{sec:private}): commands the person directs to the symbiont, with states such as attention or fatigue read only under separate authorization.
They are also the channel in which the privacy of thought (Section~\ref{sec:thought}) is at its most literal, so the paper places them at the far end of the constellation, with the highest ceiling and the strictest conditions, and treats their maturation as part of the hardware problem of Section~\ref{sec:openproblems}.
The second class is the environment itself: cameras and microphones in the home, the cabin sensors and head-up display of a car, the instruments of a laboratory, eye-tracking contact lenses, haptic garments, and medical implants such as continuous glucose monitors.
These enter the constellation on the same terms as the wearables, each with a role on the way in or out, separate authorization, and management by one runtime, and they are where the symbiont's senses turn into reach (Section~\ref{sec:reach}).
We therefore treat the full list of modalities in three tiers.
Sight, sound, text, and digital traces are the bulk of what is usable today; physiology, motion, and gaze are collectable now and need research into their interpretation; and neural signals run from available, in surface EMG, through research, in non-invasive recording, to distant, in implanted interfaces, with the chemical senses furthest of all.
``Omni'' means that whatever signal arrives can be used; it does not require that every signal be collected.

\begin{table}[t]
\centering\footnotesize
\caption{Signal sources and what each reveals about the physical and the mental side of a life. ``Omni'' means the system can take whatever signal arrives; it does not mean every signal must be collected. The two sides of the table anticipate Section~\ref{sec:whatunderstood}; the last column is the horizon on which each source becomes usable.}
\label{tab:signals}
\setlength{\tabcolsep}{3.5pt}\renewcommand{\arraystretch}{1.22}
\begin{tabularx}{\linewidth}{@{}L{2.9cm}L{2.0cm}YYL{1.95cm}C{1.1cm}@{}}
\toprule
\tabhead \colhead{Source} & \colhead{Modality} & \colhead{Physical side: what it reveals} & \colhead{Mental side: what it reveals} & \colhead{Wearing and privacy burden} & \colhead{Horizon} \\
\midrule
\ico{\faGlasses}\ Glasses camera & \lb{tbTeal}{V}\ vision & where you are, whom you see, what you are doing & where your attention falls & \lv{3}\ bystanders & \mnow \\
\rowcolor{tabStripe}\ico{\faMicrophone}\ Microphones & \lb{tbBlue}{A}\ audio & conversation content, environment & tone, hesitation, cues to emotion & \lv{3} & \mnow \\
\ico{\faCommentDots}\ Whisper mic & \lb{tbPink}{W}\ private speech & \nna & the thoughts and requests you choose to voice & \lv{1} & \mnow \\
\rowcolor{tabStripe}\ico{\faEye}\ Eye tracking in glasses & \lb{tbTeal}{E}\ gaze & what you are looking at, what you read & attention, interest, hesitation & \lv{2} & \mres \\
\ico{\faHeartbeat}\ Physiology (ring, watch, hearables, implants) & \lb{tbOrange}{P}\ sensing & sleep, movement, heart rate, glucose & indirect cues to stress and fatigue & \lv{2} & \mnow \\
\rowcolor{tabStripe}\ico{\faHandPaper}\ EMG band & \lb{tbGreen}{G}\ gesture, writing & what you are operating & commands you give deliberately & \lv{1} & \mnow \\
\ico{\faBrain}\ Neural interface (headband, in-ear EEG, implant) & \lb{tbInk}{N}\ neural & motor intent & commands the person directs to it; attention and load only under separate authorization & \lv{3}\ thought data & \mres\,/\,\mfar \\
\rowcolor{tabStripe}\ico{\faLaptop}\ Desktop and phone & \lb{tbViolet}{D}\ digital traces & what you are working on, whom you contact & the problem you are thinking about, unfinished business & \lv{2}\ work secrets & \mnow \\
\ico{\faWalking}\ Motion and location & \lb{tbGray}{M}\ inertial, position & routes, habits & daily rhythm & \lv{2} & \mnow \\
\rowcolor{tabStripe}\ico{\faHome}\ Environment (home, car, laboratory, workplace) & \lb{tbAmber}{S}\ surroundings & who is present, the state of the room, the vehicle, the instrument & routines & \lv{3}\ bystanders & \mnow \\
\bottomrule
\end{tabularx}
\tablegend{\textbullet\ Burden: \lv{1}\ low\quad \lv{2}\ medium\quad \lv{3}\ high; the note names who carries it.\quad \textbullet\ Horizon: \mnow\ usable now\quad \mres\ research\quad \mfar\ distant.\quad \textbullet\ Modality letters: \lb{tbTeal}{V} vision, \lb{tbBlue}{A} audio, \lb{tbPink}{W} whispered speech, \lb{tbTeal}{E} gaze, \lb{tbOrange}{P} physiology, \lb{tbGreen}{G} gesture, \lb{tbInk}{N} neural, \lb{tbViolet}{D} digital traces, \lb{tbGray}{M} motion, \lb{tbAmber}{S} surroundings.}
\end{table}

\subsection{Channels Out}\label{sec:channelsout}
How the symbiont speaks to the person decides whether it can be used at all in a given setting, so output deserves the same attention as input.
\tabref{tab:channels} lists the channels, from a word in the ear to a picture projected into a room.
Three design rules follow.
The strength of a message must match the setting: a line of text in a meeting, a sentence while walking, an expanded answer when the person is alone; and since the choice of channel is part of \emph{how} the symbiont helps, it is one of the decision variables of the intervention policy in Section~\ref{sec:policy}.
A private channel and a shared canvas are different instruments: the first lets the person receive help among people without disturbing them, the second lets the person and the symbiont, or the person and other people, look at the same picture and think about it together, and this second instrument, Jarvis's holographic table, earns its place in collaboration and explanation.
And every channel must support retraction and update: a cue that expires before delivery should never appear, and one overtaken by new information must be withdrawn (Section~\ref{sec:loops}).
The AR overlay deserves separate mention, because it is where the symbiont can draw its understanding onto the world the person sees: a name and the last topic discussed next to the person in front of them, the paragraph relevant to what they are thinking highlighted on the document on the desk, a route laid over a street in a city they do not know.
Current single-eye displays with small fields of view make the first of these feasible today, and wide-field lightweight AR remains a research problem.

\begin{table}[t]
\centering\footnotesize
\caption{Channels out, and when each fits. The private channels (first three rows) deliver help to the person alone, discreetly where the setting allows; the shared canvas (last row) lets the person and the symbiont, or the person and other people, look at the same picture. Jarvis's holographic table belongs to the last row.}
\label{tab:channels}
\setlength{\tabcolsep}{4pt}\renewcommand{\arraystretch}{1.22}
\begin{tabularx}{\linewidth}{@{}L{3.0cm}L{2.7cm}YC{1.1cm}C{1.3cm}C{1.0cm}@{}}
\toprule
\tabhead \colhead{Channel out} & \colhead{Form} & \colhead{Fits} & \colhead{Carries} & \colhead{Noticed by others} & \colhead{Status} \\
\midrule
\ico{\faHeadphones}\ In-ear voice & a sentence, a word & walking, alone, driving & \lv{2} & \xmark & \mnow \\
\rowcolor{tabStripe}\ico{\faGlasses}\ A line on the glasses & a name, a number, a cue & meetings, conversation, social settings & \lv{1} & \xmark & \mnow \\
\ico{\faBell}\ Haptics & a rhythm of vibration & anywhere the lowest interruption is required & \lv{1} & \xmark & \mnow \\
\rowcolor{tabStripe}\ico{\faMobile}\ Phone or desktop screen & expanded content & when you are sitting down & \lv{3} & \pmark & \mnow \\
\ico{\faEye}\ AR overlay & annotations on real objects and people, routes, translations & when you and it must look at the same thing & \lv{3} & \pmark & \mres \\
\rowcolor{tabStripe}\ico{\faProjectDiagram}\ Spatial projection, shared canvas & one picture you, it, and others can all see & collaboration, explanation, review & \lv{3} & \cmark & \mfar \\
\bottomrule
\end{tabularx}
\tablegend{\textbullet\ Carries: \lv{1}\ a cue or a word\quad \lv{2}\ a sentence\quad \lv{3}\ expanded content.\quad \textbullet\ Noticed by others: \xmark\ no\quad \pmark\ partly or depends\quad \cmark\ yes.\quad \textbullet\ Status: \mnow\ available now\quad \mres\ research prototypes\quad \mfar\ far off.}
\end{table}

\subsection{Private Channels}\label{sec:private}
Privacy, as a property, has two parts that the design keeps apart: what passes between the person and the symbiont belongs to the person, and its delivery can be discreet, so that the person can speak to the symbiont and hear from it without disturbing anyone nearby.
On the way in, this means a breath-level microphone, very short gestures (an EMG band removes the need to raise a hand), and typed text; silent speech from muscle signals and neural interfaces lie further out, and the boundary we keep is the one AlterEgo marks, since it reads what a person deliberately subvocalizes and nothing else~\citep{kapur2018alterego}, a rule that binds a neural interface as much as a microphone.
On the way out, it means in-ear voice, a line in the glasses, and haptic cues, with strength and content matched to the setting.
Three requirements that sound alike must be kept apart.
That the content of the exchange stays the person's is what the paper requires.
That its delivery is discreet is a capability whose use follows the setting: where a setting presumes unaided performance or the sincerity of what is said, as an examination, a negotiation, or an apology does, whether the person is being assisted becomes a declarable fact, and declared presence is the default there (Sections~\ref{sec:status} and~\ref{sec:conversations}).
That a device record other people without their knowledge is what the paper rejects, and Section~\ref{sec:bystanders} sets the rules.

\subsection{Always On}\label{sec:alwayson}
Companionship around the clock translates into a system requirement of availability: the symbiont can be reached and can pick up at any moment, which is a different thing from computing at full load at every moment.
Battery life and latency are the two hard constraints, and the design that reconciles them is a \keyterm{cognitive duty cycle}.
Most of the time only lightweight event detection runs: someone says the person's name, a meeting begins, their stress rises, they whisper.
An event wakes heavier computation, and the quiet hours of the night are used for offline consolidation (Section~\ref{sec:streams}).
The structure mirrors human attention, which is also not continuous.
It yields three research quantities: the recall and false-alarm rate of event detection, the latency budget from an event to a useful cue, and the daily energy and thermal budget.
Low latency has to be written as an end-to-end budget: how many milliseconds each stage may take from sensor to model to ear, which stages can run on the device and which must go to the cloud, and which can be prepared in advance so that the fast loop of Section~\ref{sec:loops} finds its material ready.
The duty cycle is also the first answer to the concern that round-the-clock AI depends on enormous compute; it turns the concern into an engineering quantity, and Section~\ref{sec:economics} completes the argument.

\subsection{Co-Evolution of Hardware and Models}\label{sec:coevolve}
The form of the senses will change with the capability of the models, and the training targets of the models will change with new senses.
From today's recorders and glasses to EMG bands, silent speech, and eventually neural interfaces, each step makes it easier for the AI to understand what a person has in mind, and each step raises the level of privacy at stake.
Two judgments follow from the recent record.
A dedicated device that adds little to the phone does not last, as the AI pin of 2024 did not~\citep{zeff2025humane}, and a cognitive core without the right hardware never reaches the scene.
Hardware and the core are one design problem, and Section~\ref{sec:stack} treats a change of device as one of the ways the whole stack evolves: the person changes their glasses, and their symbiont is still theirs.

\section{Personal Context and Lifelong Knowledge}\label{sec:context}
The senses of Section~\ref{sec:sensing} deliver streams.
This section asks how streams become something a symbiont can use: the live context of what the person is doing now, and the knowledge that stays valid across years.
The two words are not interchangeable.
Knowledge is what has settled; context is what is current, from the project the person is working on this week to the scene the glasses see this second.
Both must be kept, and the claim that organizes the section is that one entry point has to receive all of a person's context, work and life together, for the picture of the person to be rich enough for cognitive-level help.

\subsection{Three Timescales of One Record}\label{sec:timescales}
\begin{figure}[t]
  \centering
  \resizebox{0.92\linewidth}{!}{\begin{tikzpicture}[x=1cm,y=1cm,font=\small\sffamily,
  band/.style={draw=#1,fill=#1!10,rounded corners=4pt,minimum width=8.2cm,minimum height=1.3cm,text width=7.4cm,align=center,inner xsep=6pt,line width=0.8pt},
  src/.style={font=\scriptsize\sffamily,text=symSlate,anchor=east,align=right,text width=2.45cm},
  use/.style={font=\scriptsize\sffamily,text=symSlate,anchor=west,align=left,text width=2.45cm}]
  \node[band=stKeep,draw=stKeep!80!black,fill=stKeep!18] (k) at (0,3.2) {\textbf{Knowledge}\quad{\scriptsize months to years $\cdot$ small, curated, persistent}\\{\scriptsize facts about you and your world, relationships, preferences, commitments, your history, the reasons you gave}};
  \node[band=stKeep,fill=stKeep!10] (w) at (0,1.6) {\textbf{Working state}\quad{\scriptsize days to weeks $\cdot$ projects and open threads}\\{\scriptsize the project you are pushing, the discussion not yet closed, this week's decision, the logic tree of events}};
  \node[band=stKeep,draw=stKeep!60,fill=stKeep!4] (c) at (0,0) {\textbf{Live context}\quad{\scriptsize seconds to hours $\cdot$ large volume, mostly discarded}\\{\scriptsize this conversation, the file open in the IDE and its error, the scene the glasses see, today's calendar}};
  \draw[-{Stealth[length=2.2mm]},line width=1.1pt,stKeep!80!black] (-4.5,-0.6) -- (-4.5,3.8);
  \node[rotate=90,anchor=south,font=\scriptsize\sffamily,text=stKeep!80!black] at (-4.58,1.6) {consolidate: filter, merge, promote};
  \draw[-{Stealth[length=2.2mm]},line width=1.1pt,symAccent] (4.5,3.8) -- (4.5,-0.6);
  \node[rotate=-90,anchor=south,font=\scriptsize\sffamily,text=symAccent] at (4.58,1.6) {recall into the working context $c_t$};
  \node[src] at (-5.0,3.2) {settled from many contexts; your own account};
  \node[src] at (-5.0,1.6) {project trees, threads, decisions in progress};
  \node[src] at (-5.0,0) {chat, IDE, desktop, phone, wearables, your whisper (Table 7)};
  \node[use] at (5.0,3.2) {feeds the personal model (\S7)};
  \node[use] at (5.0,1.6) {feeds long-horizon thinking (\S9.2)};
  \node[use] at (5.0,0) {feeds real-time participation (\S8)};
\end{tikzpicture}}
  \caption{Three timescales of one record.
  Live context is large and mostly discarded, working state holds projects and open threads for days to weeks, and knowledge is small, curated, and persistent.
  Consolidation moves material up; recall moves the right slice down into the working context $c_t$ of Definition~\ref{def:workingcontext}.}
  \label{fig:timescales}
\end{figure}
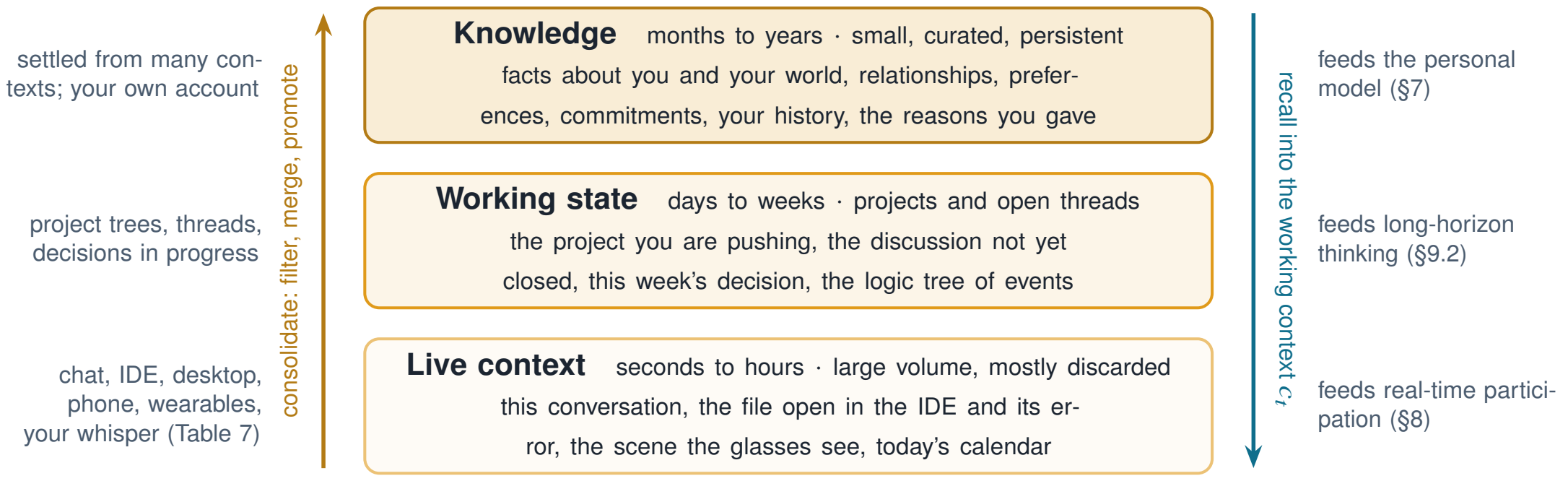

\keyterm{Context} is information relevant to what the person is doing now.
It is short-lived, high-volume, and mostly discarded once used: where this conversation has got to, which file is open in the editor and which error it stopped on, what the glasses can see and who is present, what is on today's calendar.
\keyterm{Knowledge} is what has settled out of many contexts and stays relevant across them.
It is long-lived, curated, and comparatively small: facts about the person and their world, relationships, preferences, commitments, their history, the reasons they gave for things.
Between the two lies a third layer, \keyterm{working state}, which lives for days or weeks: the project being pushed forward, the discussion that has not been closed, the decision due this week.
The project tree in which events and projects organize one another belongs here.
Two flows connect the three layers (\figref{fig:timescales}).
Consolidation moves material upward, from context to working state to knowledge, through the filtering and merging of Section~\ref{sec:streams}.
Recall moves material downward, placing the right slice of knowledge into the context of the moment, which is exactly the step from $\Ctx^h_t$ to $\wctx_t$ in Definition~\ref{def:workingcontext}.
When one word is needed for the whole we write \emph{the record}.

The word \emph{context} is also used for the input window of a language model, and in context engineering it refers to the information given to a model so that the model can complete a task~\citep{hua2025context}.
Personal context, as used here, is information about the person that is valid now, and its lifecycle serves a person's life.
The two meet at the recall step: injecting personal knowledge into a model's context window is an act of context engineering.
We borrow the lifecycle view and nothing else.

\subsection{One Entry Point for Every Context}\label{sec:entrypoint}
\tabref{tab:sources} lists the sources of personal context and what each sees on its own.
Each sees a partial person.
A chat assistant knows someone who asks questions, a coding assistant knows a programmer, a desktop agent knows someone at work, and a band knows a body with a heart rate.
Questions that matter for cognitive-level help cross these boundaries: why the person was unproductive in the editor this morning (they slept five hours and the first meeting went badly), how this decision relates to the one they made six months ago, whether the stress the band sees matches the deadline the desktop sees.
\claim{Only an entry point that receives every source into one record can answer such questions, and only such a record supports a picture of the person rich enough for the help described in Section~\ref{sec:ladder}.}
The same observation has a consequence for the runtime of Section~\ref{sec:runtime}: it must be a layer across devices and applications, and it cannot be any one application.
Receiving every source is a capability of the entry point and never the default for collection.
Each source is authorized separately, the symbiont must be useful at low levels of authorization (Section~\ref{sec:trustcurve}), and the privacy column of the table exists for that reason.
This is also what separates the proposal from one more memory layer: memory research records the conversations of a single entry point, whereas the record here spans all of a person's entry points.

\begin{table}[t]
\centering\footnotesize
\caption{Sources of personal context, and what each sees on its own. A chatbot knows a questioner, a coding assistant knows a programmer, a desktop agent knows an office worker, a band knows a body with a heart rate. Only one entry point that joins them can answer questions that cross the boundaries. Each source is authorized separately (Section~\ref{sec:trustcurve}).}
\label{tab:sources}
\setlength{\tabcolsep}{4pt}\renewcommand{\arraystretch}{1.22}
\begin{tabularx}{\linewidth}{@{}L{2.4cm}L{2.6cm}YC{1.15cm}L{1.6cm}L{2.0cm}C{1.1cm}@{}}
\toprule
\tabhead \colhead{Source} & \colhead{Typical form} & \colhead{The side of you it sees} & \colhead{Intent explicit} & \colhead{Timescale} & \colhead{Access today} & \colhead{Privacy weight} \\
\midrule
\ico{\faComments}\ Passive chat & web or app conversation & the questions you choose to bring & \lv{3} & minutes & \mnow\ available & \lv{1} \\
\rowcolor{tabStripe}\ico{\faCode}\ Coding environment & IDE, terminal, repository & what you are building, where you are stuck, how you work & \lv{3} & hours--weeks & \mnow\ coding agents & \lv{2} \\
\ico{\faLaptop}\ Desktop and office & documents, mail, calendar, meetings, browser & your work, your collaborators, your rhythm & \lv{2} & hours--months & \mres\ desktop agents & \lv{2} \\
\rowcolor{tabStripe}\ico{\faMobile}\ Phone & messages, location, photos, notifications & your plans, relationships, movements & \lv{2} & minutes--days & \mres\ partial & \lv{3} \\
\ico{\faGlasses}\ All-day wearables & first-person audio and video, physiology, motion & what you lived through, how your body is, whom you were with & \lv{1} & seconds to a lifetime & \mres\ early & \lv{3} \\
\rowcolor{tabStripe}\ico{\faCommentDots}\ Your own voice & a whispered sentence & what you think, feel, and why, right now & \lv{3} & seconds & \mnow\ available & \lv{3} \\
\bottomrule
\end{tabularx}
\tablegend{\textbullet\ Intent explicit: \lv{1}\ inferred\quad \lv{2}\ partly stated\quad \lv{3}\ stated by you.\quad \textbullet\ Access: \mnow\ available\quad \mres\ partial or early\quad \mfar\ to be built.\quad \textbullet\ Privacy weight: \lv{1}\ low\quad \lv{2}\ medium\quad \lv{3}\ high.}
\end{table}

\subsection{From Streams to a Record}\label{sec:streams}
Three stages turn raw streams into knowledge.
At the front, on the device and at the edge, filtering drops what is high-frequency, volatile, and not worth keeping and forwards the rest.
At the back, storage absorbs frequent writes and updates and grows large.
The third stage is periodic offline consolidation, which plays the role that sleep plays for human memory: in idle hours it merges the day's fragments into events, files events into projects, promotes repeated patterns into stable knowledge, and demotes what has expired.
Sleep-time compute and memory systems in which memories link and evolve are technical precedents~\citep{lin2025sleeptime,xu2025amem}, and the rhythm resolves the tension between filtering at the front and accumulation at the back.
Scale forces a decision about fidelity.
A day of first-person audio and video runs to gigabytes and a life to petabytes; keeping everything is impossible and unnecessary.
Raw streams are kept briefly; event-level summaries and indices of when and where something of a given kind happened are kept for the long term; knowledge entries carry pointers back to their sources.
The underlying difficulty is that what will matter later is unknown now, and the compromise is a low-fidelity index that allows the past to be revisited.
Section~\ref{sec:openproblems} lists modeling a life's record at this scale as the first open problem.

\subsection{One Life, Many Views}\label{sec:views}
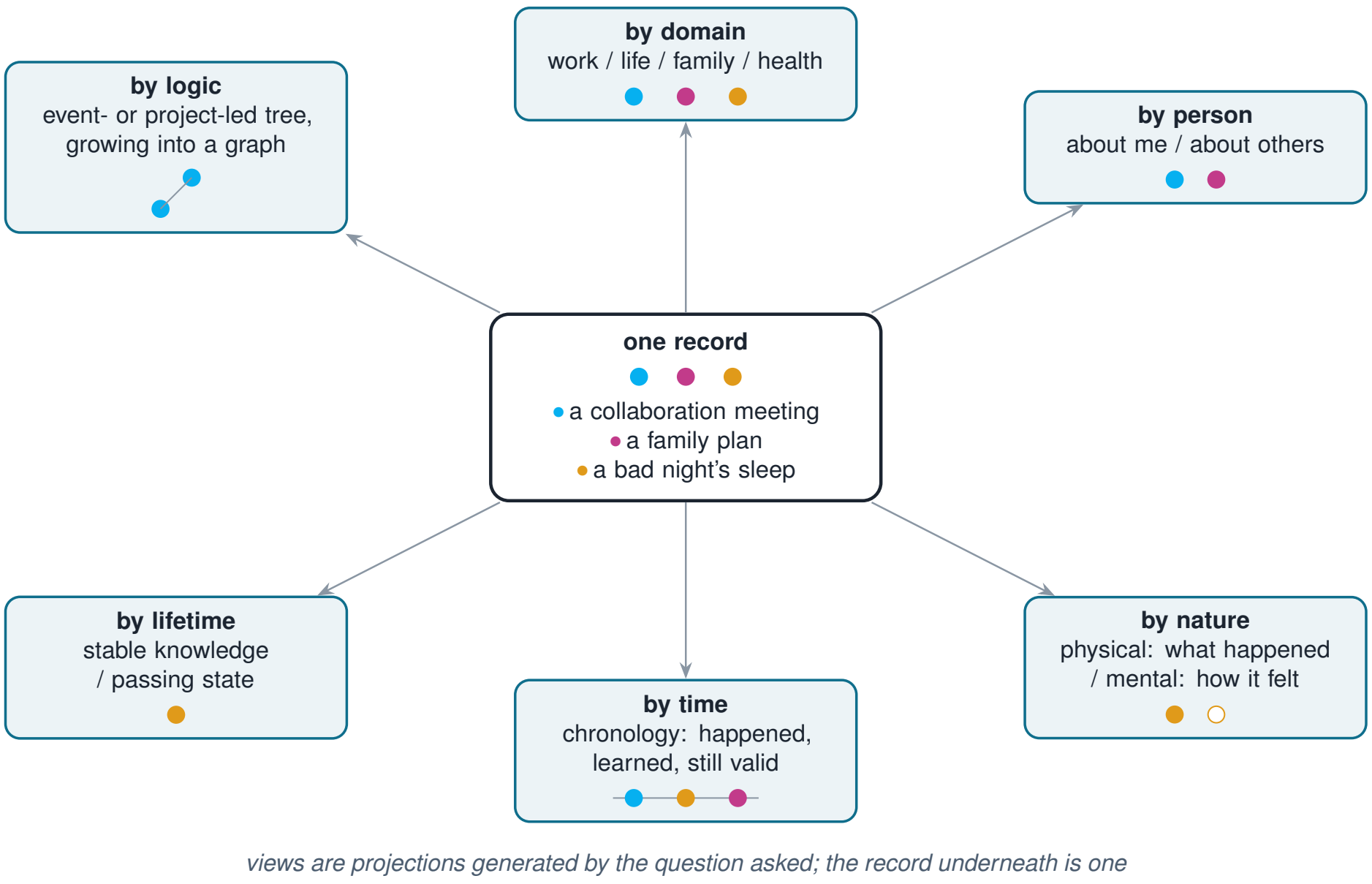
\begin{figure}[t]
  \centering
  \resizebox{0.78\linewidth}{!}{\begin{tikzpicture}[x=1cm,y=1cm,font=\scriptsize\sffamily,
  view/.style={draw=#1,fill=#1!8,rounded corners=4pt,text width=3.0cm,align=center,inner sep=4pt,line width=0.7pt},
  ev/.style={circle,draw=#1,fill=#1,inner sep=1.6pt},
  arr/.style={-{Stealth[length=1.6mm]},symGray,line width=0.5pt}]
  \node[draw=symInk,fill=white,rounded corners=5pt,text width=3.4cm,align=center,inner sep=5pt,line width=0.9pt] (core) at (0,0)
    {\textbf{one record}\\[2pt]
     \tikz{\node[ev=symCyan] at (0,0) {}; \node[ev=symMagenta] at (0.45,0) {}; \node[ev=symAmber] at (0.9,0) {};}\\[1pt]
     \textcolor{symCyan}{$\bullet$}\,a collaboration meeting\\\textcolor{symMagenta}{$\bullet$}\,a family plan\\\textcolor{symAmber}{$\bullet$}\,a bad night's sleep};
  \node[view=symAccent] (v1) at (-4.9,2.5) {\textbf{by logic}\\event- or project-led tree, growing into a graph\\[2pt]\tikz{\node[ev=symCyan] at (0,0){}; \node[ev=symCyan] at (0.3,0.3){}; \draw[symGray] (0,0)--(0.3,0.3);}};
  \node[view=symAccent] (v2) at (0,3.3) {\textbf{by domain}\\work / life / family / health\\[2pt]\tikz{\node[ev=symCyan] at (0,0){}; \node[ev=symMagenta] at (0.5,0){}; \node[ev=symAmber] at (1.0,0){};}};
  \node[view=symAccent] (v3) at (4.9,2.5) {\textbf{by person}\\about me / about others\\[2pt]\tikz{\node[ev=symCyan] at (0,0){}; \node[ev=symMagenta] at (0.4,0){};}};
  \node[view=symAccent] (v4) at (-4.9,-2.5) {\textbf{by lifetime}\\stable knowledge / passing state\\[2pt]\tikz{\node[ev=symAmber] at (0,0){};}};
  \node[view=symAccent] (v5) at (0,-3.3) {\textbf{by time}\\chronology: happened, learned, still valid\\[2pt]\tikz{\draw[symGray] (0,0)--(1.4,0); \node[ev=symCyan] at (0.2,0){}; \node[ev=symAmber] at (0.7,0){}; \node[ev=symMagenta] at (1.2,0){};}};
  \node[view=symAccent] (v6) at (4.9,-2.5) {\textbf{by nature}\\physical: what happened / mental: how it felt\\[2pt]\tikz{\node[ev=symAmber] at (0,0){}; \node[ev=symAmber,fill=white] at (0.4,0){};}};
  \foreach \v in {v1,v2,v3,v4,v5,v6}{\draw[arr] (core) -- (\v);}
  \node[font=\scriptsize\sffamily\itshape,text=symSlate] at (0,-4.4) {views are projections generated by the question asked; the record underneath is one};
\end{tikzpicture}}
  \caption{One life, many views.
  The same handful of events (a meeting, a family plan, a bad night's sleep) appears in six views, by logic, domain, person, lifetime, time, and nature.
  The views are projections of one record and can be switched at will.}
  \label{fig:views}
\end{figure}

The organizing claim of the section is that the same record is projected into different views depending on the question, and that any view can be switched to at any time (\figref{fig:views}, \tabref{tab:views}).
A logic view arranges events and projects as a tree that grows into a graph; a domain view separates work, life, family, and health; a person view separates what concerns the person from what concerns others; a lifetime view separates stable knowledge from passing state, which aligns with the three timescales above; a chronological view orders by when something happened, when it was learned, and whether it still holds; and a physical/mental view separates what happened from how it felt.
Views are projections of one store: the record underneath is one, and a view is generated by the question.
The logic view is the main thread, and a life is never forced into a single tree.
MyLifeBits offered multiple views for retrieval~\citep{gemmell2006mylifebits}; here views also serve understanding (Section~\ref{sec:understanding}) and the person's own inspection and correction of what is held about them (Section~\ref{sec:trust}).
The paper specifies requirements on representation and retrieval and does not prescribe a graph or a vector store.

\begin{table}[t]
\centering\footnotesize
\caption{Views of the record and the questions they answer. Views are projections of one store: the record underneath is one, and a view is generated by the question being asked.}
\label{tab:views}
\setlength{\tabcolsep}{5pt}\renewcommand{\arraystretch}{1.22}
\begin{tabularx}{\linewidth}{@{}L{2.9cm}YYC{1.7cm}@{}}
\toprule
\tabhead \colhead{View} & \colhead{Question it answers} & \colhead{Typical use} & \colhead{Used mainly by} \\
\midrule
\ico{\faSitemap}\ Logic (project) & what this belongs to, what led to it and what follows & picking up a line of thought; continuing unfinished work & \icKeep\ \icHelp \\
\rowcolor{tabStripe} \ico{\faThLarge}\ Domain & how work and life cross & spotting conflicts; allocating time & \icHelp \\
\ico{\faUserFriends}\ Person (self / others) & who is involved and how they are related & memory support in social settings & \icLive\ \icHelp \\
\rowcolor{tabStripe} \ico{\faHourglassHalf}\ Lifetime & what is stable and what is passing & telling character from a mood & \icUnderstand \\
\ico{\faClock}\ Time (chronology) & when it happened, when it was learned, whether it still holds & tracking change & \icKeep\ \icGrow \\
\rowcolor{tabStripe} \ico{\faHeart}\ Nature (physical / mental) & what happened, and how you felt about it & emotional support; self-understanding & \icUnderstand\ \icHelp \\
\bottomrule
\end{tabularx}
\tablegend{\textbullet\ Used mainly by: the loop stage that most often asks for the view: \icKeep\ keep, \icUnderstand\ understand, \icLive\ live alongside, \icHelp\ act for and think with, \icGrow\ grow.}
\end{table}

\subsection{Updating, Forgetting, and Repair}\label{sec:repair}
Frequent updating raises four problems: conflict, change, forgetting, and repair.
Conflict between old and new is handled by keeping three times apart: when something happened, when the system learned it, and until when it holds.
Change is the second, since preferences change, relationships change, and commitments are kept or cancelled, and the record has to follow them without losing the history.
Forgetting, the third, comes in two kinds, decay that reduces noise and deletion that respects the person's wishes, and deleting a raw record does not delete the summaries and judgments derived from it; the paper states the consistency requirement and leaves the mechanism open.
Repair is the fourth.
When the person says ``you misunderstood me'', the system must locate the layer in which the misunderstanding lives, a fact, a reason, a current goal, or a long-term preference, and propagate the correction to the entries inferred from it and to the advice already given; changing its answer in the conversation is only the first step.
This gives a measurable quantity: the half-life of an error in the record, and the number of inferences it reached before it was corrected.
Section~\ref{sec:stack} returns to the related question of what survives when the base model is replaced.

\subsection{Remembering the Future, and the Reasons}\label{sec:future}
A record holds more than the past.
It holds commitments, plans, the question to ask someone at the next meeting, decisions still forming, and the longer view of who the person wants to become.
Psychology calls this prospective memory; it is the kind of memory people fail at most often, and it is where a symbiont produces value most easily, by returning a future intention to the working context at the right moment.
It also gives the proactive help of Section~\ref{sec:living} its most natural trigger: the system acts because the person said they would, and does not have to guess that they might want to.
The record also holds reasons and not only conclusions: why the person declined that offer, why they abandoned that plan, which two options they still have not chosen between, what new evidence would make them reconsider; the third moment of the day in Section~\ref{sec:intro} is a record that kept the conclusion and lost the reasons.
Keeping reasons is the precondition for the decision support of Section~\ref{sec:thinking}; a record without reasons can only answer questions about history.

\subsection{The Others in Your Life}\label{sec:others}
A person's colleagues, family, and friends are part of their context, and a symbiont has to remember where a conversation with someone left off in order to support the person's social life.
Those people have not authorized being modeled.
We distinguish relational facts, what has happened between the person and them, which may be kept, from inferred portraits, what kind of person they are, which must be held to stricter limits than anything about the person themselves.
Section~\ref{sec:bystanders} sets the boundary in detail.

\section{Understanding a Person}\label{sec:understanding}
The record of Section~\ref{sec:context} is evidence.
This section asks how evidence becomes understanding of a person, and it is the technical center of the paper.
It runs along one line: what must be understood, where the evidence comes from, how the understanding is represented, which backbone it runs on, what a world model adds to it, and where it stops.
World models appear only at the fifth step, as the source of one capability, prediction.

\subsection{What Must Be Understood}\label{sec:whatunderstood}
Three objects must be understood.
The first is the \emph{situation}, the physical side: where the person is, whom they are with, what they are doing, what the world in front of them is about to do.
The second is the \emph{mind}, the mental side: the person's intention at this moment, their cognitive load, their emotion, their uncertainty, and the more stable layer of goals, values, and blind spots.
The third is \emph{change}: the situation changes by the minute, the mind by the hour, and the person by the year, so that the difficulty of understanding someone lies less in drawing a likeness than in making the likeness move.
The same experience has two faces, the event visible from outside and the person's experience of it, and the record keeps both, so the model must represent both.
Mental world modeling makes the same commitment at the level of world state, treating beliefs and intentions as state variables alongside physical ones~\citep{fei2026mental}; Section~\ref{sec:twoworldmodels} builds on it.
The third object, change, is the root of the staleness detection in that section and of the mutual growth in Section~\ref{sec:growing}.

\subsection{Two Roads to the Mind}\label{sec:tworoads}
\begin{figure}[t]
  \centering
  \resizebox{0.84\linewidth}{!}{\begin{tikzpicture}[x=1cm,y=1cm,font=\scriptsize\sffamily,
  road/.style={draw=#1,fill=#1!8,rounded corners=4pt,text width=3.6cm,align=center,inner sep=4pt,line width=0.8pt},
  note/.style={draw=#1!60,dashed,fill=white,rounded corners=4pt,text width=3.6cm,align=center,inner sep=4pt,line width=0.6pt,text=symSlate},
  arr/.style={-{Stealth[length=2mm]},line width=0.9pt}]
  \node[road=symCyan] (o1) at (-3.6,3.3) {\textbf{Observation road}\\behavior, choices, expression, environment, physiology};
  \node[road=symCyan] (o2) at (-3.6,1.85) {patterns across time};
  \node[road=symCyan] (o3) at (-3.6,0.45) {inferences about traits, habits, and current state};
  \node[note=symCyan] (o4) at (-3.6,-1.15) {strength: sees what you cannot see yourself\\weakness: one behavior, many causes};
  \node[road=symMagenta] (r1) at (3.6,3.3) {\textbf{Self-report road}\\your account, a whispered sentence, a correction of a misreading};
  \node[road=symMagenta] (r2) at (3.6,1.85) {organized, dated, attributed};
  \node[road=symMagenta] (r3) at (3.6,0.45) {records of intent, feeling, and reasons};
  \node[note=symMagenta] (r4) at (3.6,-1.15) {strength: only you know why\\weakness: shifts with mood, flatters the teller};
  \draw[arr,symCyan] (o1) -- (o2); \draw[arr,symCyan] (o2) -- (o3); \draw[arr,symCyan!70] (o3) -- (o4);
  \draw[arr,symMagenta] (r1) -- (r2); \draw[arr,symMagenta] (r2) -- (r3); \draw[arr,symMagenta!70] (r3) -- (r4);
  \node[draw=stUnderstand,fill=stUnderstand!8,rounded corners=5pt,text width=6.4cm,align=center,inner sep=5pt,line width=1pt] (model) at (0,-3.25) {\textbf{the layered personal model}\\every entry carries its source and its uncertainty\\[2pt]{\color{symSlate}self-report has priority on feelings and reasons;\\observation has priority on patterns you cannot see yourself}};
  \draw[arr,symCyan] (o4.south) to[out=-90,in=150] (model.north west);
  \draw[arr,symMagenta] (r4.south) to[out=-90,in=30] (model.north east);
  \node[draw=symAmber!80!black,fill=symAmber!10,rounded corners=3pt,text width=2.8cm,align=center,inner sep=3pt] (dis) at (0,0.9) {\textbf{when the roads disagree}\\mark the disagreement, lower the weight of the inference, ask when the moment is right};
  \draw[arr,symAmber!80!black,dashed] (dis) -- (model);
\end{tikzpicture}}
  \caption{Two roads to the mind.
  Observation infers patterns from behavior and physiology; self-report records what the person says about intent, feeling, and reasons.
  Both feed one layered model, with every entry tagged by source and uncertainty; disagreements are marked, down-weighted, and resolved by asking.}
  \label{fig:tworoads}
\end{figure}
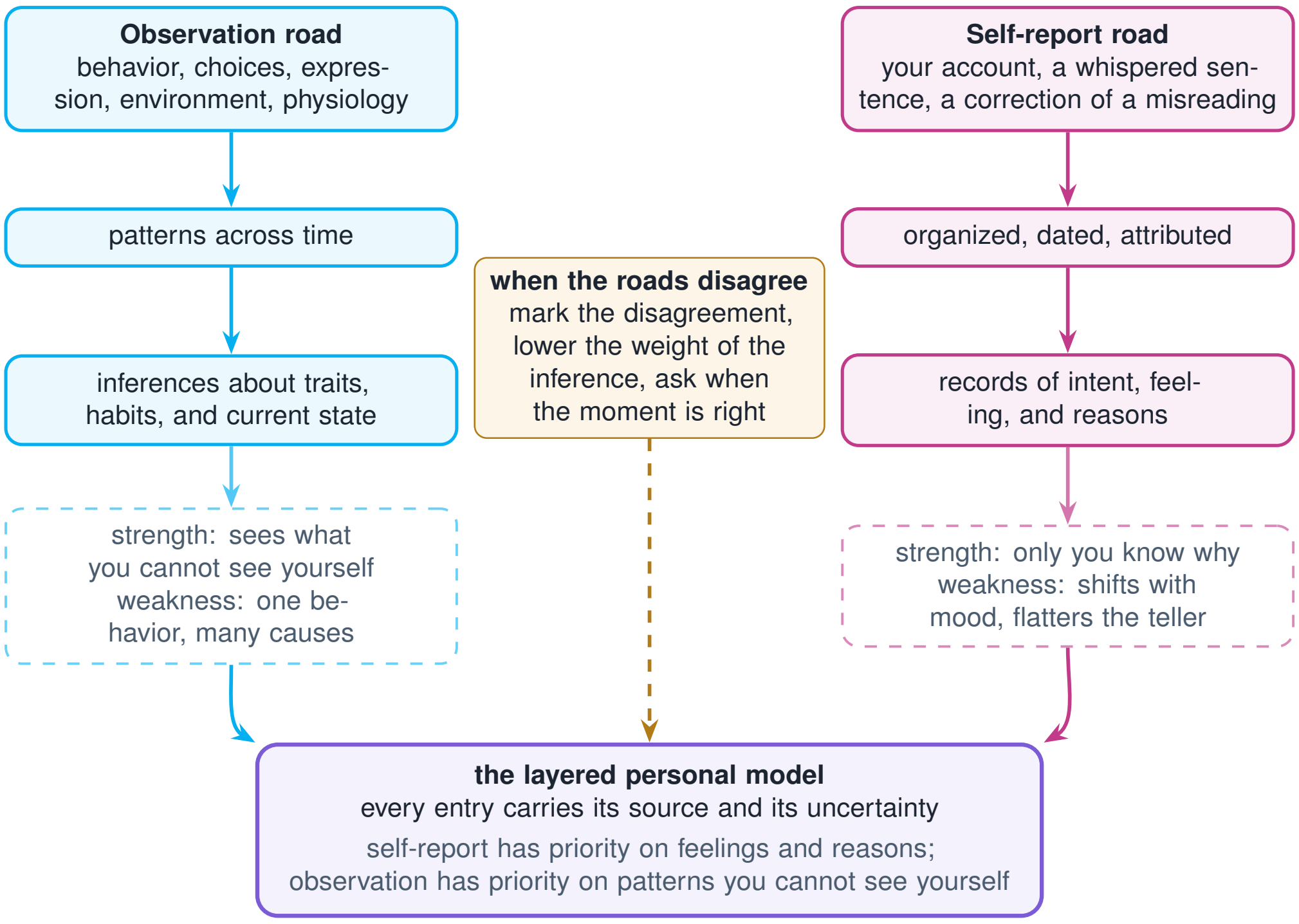

There are two independent sources of evidence about a mind, and a symbiont needs both (\figref{fig:tworoads}).
The observation road infers from behavior, choices, expression, environment, and physiology; its strength is that it sees patterns the person cannot see themselves (they do not sleep after every meeting with a certain colleague), and its weakness is that one behavior has many possible causes.
The self-report road takes what the person says: their account of a situation, a whispered sentence, a correction of a misreading; its strength is that only the person can say what they felt and why, and its weakness is that accounts shift with mood and flatter the teller.
The two roads feed one layered model, and every entry carries its source and its uncertainty.
Self-report has priority on feelings and reasons; observation has priority on patterns the person cannot see.
When they disagree, the right response is to mark the disagreement, lower the weight placed on the inference, and ask when the moment is right, never to let the model overrule the person.
The result of \citet{park2024simulation}, in which agents built from two-hour interviews reproduced their participants' survey answers about as well as the participants reproduced their own, shows how dense the information on the self-report road is; those agents are never updated, and that is what the symbiont adds.
Most existing systems travel only the observation road, inferring preferences from interaction history, and the dedicated low-friction channel of Section~\ref{sec:private} is what gives a symbiont the second.

\subsection{Layers of the Personal Model}\label{sec:layers}
\tabref{tab:layers} gives the representation: seven layers, each with its evidence, its rate of change, and the misuse it invites, so that a passing state is never mistaken for a trait.
The last layer, preferences about help, is new relative to the usual user profile and is the parameter source of the intervention policy in Section~\ref{sec:policy}.
One structural property of the table follows from an everyday observation: a person who switches between a home dialect, a national language, and English changes manner, values, and even reasoning with the switch.
The table is therefore indexed by context.
The self at work, the self at home, and the self speaking the home dialect take different values in the identity, relations, and help-preference layers and share the experience-and-knowledge layer.
The differences are not contradictions to be averaged into one personality; the model's task is to recognize which self is active, from language, setting, and company, and to act with that self's manner, weights, and social norms.
Cultural psychology documents the underlying phenomenon of cultural frame switching in bicultural people~\citep{hong2000multicultural,ramirezesparza2006bilinguals}, and the context-indexed model is what makes the cultural help of Section~\ref{sec:relating} possible.

\begin{table}[t]
\centering\footnotesize
\caption{Layers of the personal model. Each layer carries its evidence and its rate of change, and the last column names the misuse it invites. The table is indexed by context: the identity, relations, and help-preference layers take different values for the self at work, the self at home, and the self speaking a home dialect, while experience and knowledge are shared. The last layer parameterizes the intervention policy of Section~\ref{sec:policy}.}
\label{tab:layers}
\setlength{\tabcolsep}{4pt}\renewcommand{\arraystretch}{1.22}
\begin{tabularx}{\linewidth}{@{}L{2.8cm}YL{2.75cm}L{1.55cm}Y@{}}
\toprule
\tabhead \colhead{Layer} & \colhead{Content} & \colhead{Main evidence} & \colhead{Changes over} & \colhead{Typical misuse} \\
\midrule
\nc{1}\ \ico{\faIdBadge}\ Stable identity \lyI & background, roles, long-term values, abilities and blind spots & \lb{tbBlue}{O}\lb{tbGreen}{R}\ long observation plus self-report & years & reading one lapse as a lack of ability \\
\rowcolor{tabStripe}\nc{2}\ \ico{\faBook}\ Experience and knowledge \lyE & what happened, what you know, whom you have dealt with & \lb{tbBlue}{O}\lb{tbGreen}{R}\ the record (\S\ref{sec:context}) & accumulates & treating an expired fact as current \\
\nc{3}\ \ico{\faMapMarker}\ Current situation \lyS & where, with whom, doing what, under what time pressure & \lb{tbBlue}{O}\ real-time sensing & minutes & explaining new behavior by an old situation \\
\rowcolor{tabStripe}\nc{4}\ \ico{\faBrain}\ Mental state \lyM & intention, cognitive load, emotion, uncertainty & \lb{tbViolet}{K}\ the two roads crossed & seconds to hours & treating an inference as a fact \\
\nc{5}\ \ico{\faUserFriends}\ Relations and society \lyR & relationships, boundaries, shared history & \lb{tbBlue}{O}\lb{tbGreen}{R}\ relational facts & months & building portraits of other people \\
\rowcolor{tabStripe}\nc{6}\ \ico{\faBullseye}\ Goals and trade-offs \lyG & what you want, what you will not sacrifice, who you want to become & \lb{tbGreen}{R}\ self-report first & months to years & overruling the person's choice with an inferred goal \\
\nc{7}\ \ico{\faSlidersH}\ Preferences about help \lyH & when you accept interruption, which forms you like, what you want to do yourself & \lb{tbGreen}{R}\lb{tbBlue}{O}\ feedback & weeks & turning one refusal into a permanent setting \\
\bottomrule
\end{tabularx}
\tablegend{\textbullet\ Evidence, in the terms of Definition~\ref{def:selfcontext}: \lb{tbBlue}{O} observed from outside ($O_{\le t}$), \lb{tbGreen}{R} reported by the person ($R_{\le t}$), \lb{tbViolet}{K} inferred from the two ($K_t$).\quad \textbullet\ The layer letters \lyI\ \lyE\ \lyS\ \lyM\ \lyR\ \lyG\ \lyH\ are reused in \tabref{tab:catalogue}.}
\end{table}

\subsection{The Backbone}\label{sec:backbone}
\begin{figure}[t]
  \centering
  \resizebox{0.98\linewidth}{!}{\begin{tikzpicture}[x=1cm,y=1cm,font=\scriptsize\sffamily,
  cell/.style={draw=#1!70!black,fill=#1!9,rounded corners=4pt,text width=3.95cm,minimum height=2.15cm,align=center,inner sep=4pt,line width=0.7pt},
  cellhi/.style={draw=#1!70!black,fill=#1!16,rounded corners=4pt,text width=3.95cm,minimum height=2.15cm,align=center,inner sep=4pt,line width=1.1pt},
  colhead/.style={draw=#1!70!black,fill=#1!80!black,text=white,rounded corners=3pt,text width=3.95cm,minimum height=0.8cm,align=center,inner sep=3pt,font=\footnotesize\sffamily\bfseries},
  rowhead/.style={draw=symSlate!60,fill=symSlate!8,rounded corners=3pt,text width=2.4cm,minimum height=2.15cm,align=center,inner sep=4pt,font=\scriptsize\sffamily},
  tag/.style={fill=white,draw=#1!70!black,text=#1!70!black,rounded corners=2pt,inner sep=2pt,font=\tiny\sffamily\bfseries,line width=0.5pt}]
  \def\cx{0}\def\cy{4.55}\def\cz{9.1}
  \def\rtop{2.55}\def\rbot{0}
  \node[colhead=symGray]       (h1) at (\cx,4.95) {\faComment\ \ language model};
  \node[colhead=symCyan]       (h2) at (\cy,4.95) {\faEye\ \faHeadphones\ \ omni-modal model};
  \node[colhead=stUnderstand]  (h3) at (\cz,4.95) {\faGlobe\ \ world model};
  \node[rowhead] (r2) at (-3.5,\rtop) {\textbf{inside a harness}\\[1pt]{\color{symSlate}observe, retrieve, ask, check, update, plan: understanding becomes a process}};
  \node[rowhead] (r1) at (-3.5,\rbot) {\textbf{model alone}\\[1pt]{\color{symSlate}one pass over what is in the context: understanding is a single inference}};
  \node[cell=symGray] (c11) at (\cx,\rbot) {\textbf{what you say and write}\\[2pt]a textual portrait held in context; memory reached by retrieval\\[2pt]{\color{symSlate}situation and mood guessed from text alone}};
  \node[cell=symCyan] (c21) at (\cy,\rbot) {\textbf{what you see and hear}\\[2pt]situation and mood read from tone and scene; the first-person stream becomes interpretable\\[2pt]{\color{symSlate}no prediction of what comes next}};
  \node[cell=stUnderstand] (c31) at (\cz,\rbot) {\textbf{what happens next}\\[2pt]around you and in you: the next event, your reaction to a cue\\[2pt]{\color{symSlate}predicts, but cannot act on its prediction}};
  \node[cellhi=symGray] (c12) at (\cx,\rtop) {\textbf{a text assistant that checks}\\[2pt]looks things up, asks you a question, consults the record, updates the model, plans the next step\\[2pt]{\color{symSlate}blind to the scene around you}};
  \node[cellhi=symCyan] (c22) at (\cy,\rtop) {\textbf{a present observer}\\[2pt]perceives continuously, verifies what it saw against the record, asks when unsure, prepares before you speak\\[2pt]{\color{symSlate}reacts to events; does not anticipate them}};
  \node[cellhi=stUnderstand] (c32) at (\cz,\rtop) {\textbf{an anticipating partner}\\[2pt]simulates the effect of a cue before giving it, prepares help before it is needed, times its silence\\[2pt]{\color{symSlate}the eventual home of the symbiont}};
  \node[tag=symGray,anchor=south] at ([yshift=3pt]c12.north) {first systems start here};
  \node[tag=stUnderstand,anchor=south] at ([yshift=3pt]c32.north) {where the design is headed};
  \draw[-{Stealth[length=2.2mm]},symMagenta!80!black,line width=1pt,dashed] (c11.north) -- (c12.south);
  \draw[-{Stealth[length=2.2mm]},symMagenta!80!black,line width=1pt,dashed] (c12.east) -- (c22.west);
  \draw[-{Stealth[length=2.2mm]},symMagenta!80!black,line width=1pt,dashed] (c22.east) -- (c32.west);
  \draw[-{Stealth[length=2.4mm]},symAccent,line width=1pt] (-2.1,5.75) -- (11.2,5.75)
    node[pos=0.5,above,font=\scriptsize\sffamily,text=symAccent,align=center] {\textbf{the foundation model}: more of the world becomes evidence\quad language $\to$ sight and sound $\to$ dynamics};
  \draw[-{Stealth[length=2.4mm]},symAccent,line width=1pt] (-5.0,-1.15) -- (-5.0,3.7)
    node[pos=0.5,above,rotate=90,font=\scriptsize\sffamily,text=symAccent,align=center] {\textbf{the system around the model}\\ one pass $\to$ a process};
  \draw[line width=2.2pt,symMagenta!80!black] (-4.4,-1.75) -- (11.2,-1.75);
  \node[anchor=north,font=\scriptsize\sffamily\bfseries,text=symMagenta!80!black] at (3.4,-1.85) {one information interface for all six cells: the same slots, filled with better estimates};
\end{tikzpicture}}
  \caption{Two axes of the backbone.
  Columns: the family of foundation models, from language to sight and sound to dynamics, each extending what counts as evidence.
  Rows: how much system is organized around the model, from a single pass to a process inside an agentic harness.
  The two axes are independent, since a harness can be wrapped around any of the three models; the dashed path is the likely order in which systems will move through the grid, and the heavy line is the information interface of the personal model, which is the same in every cell.}
  \label{fig:backbones}
\end{figure}
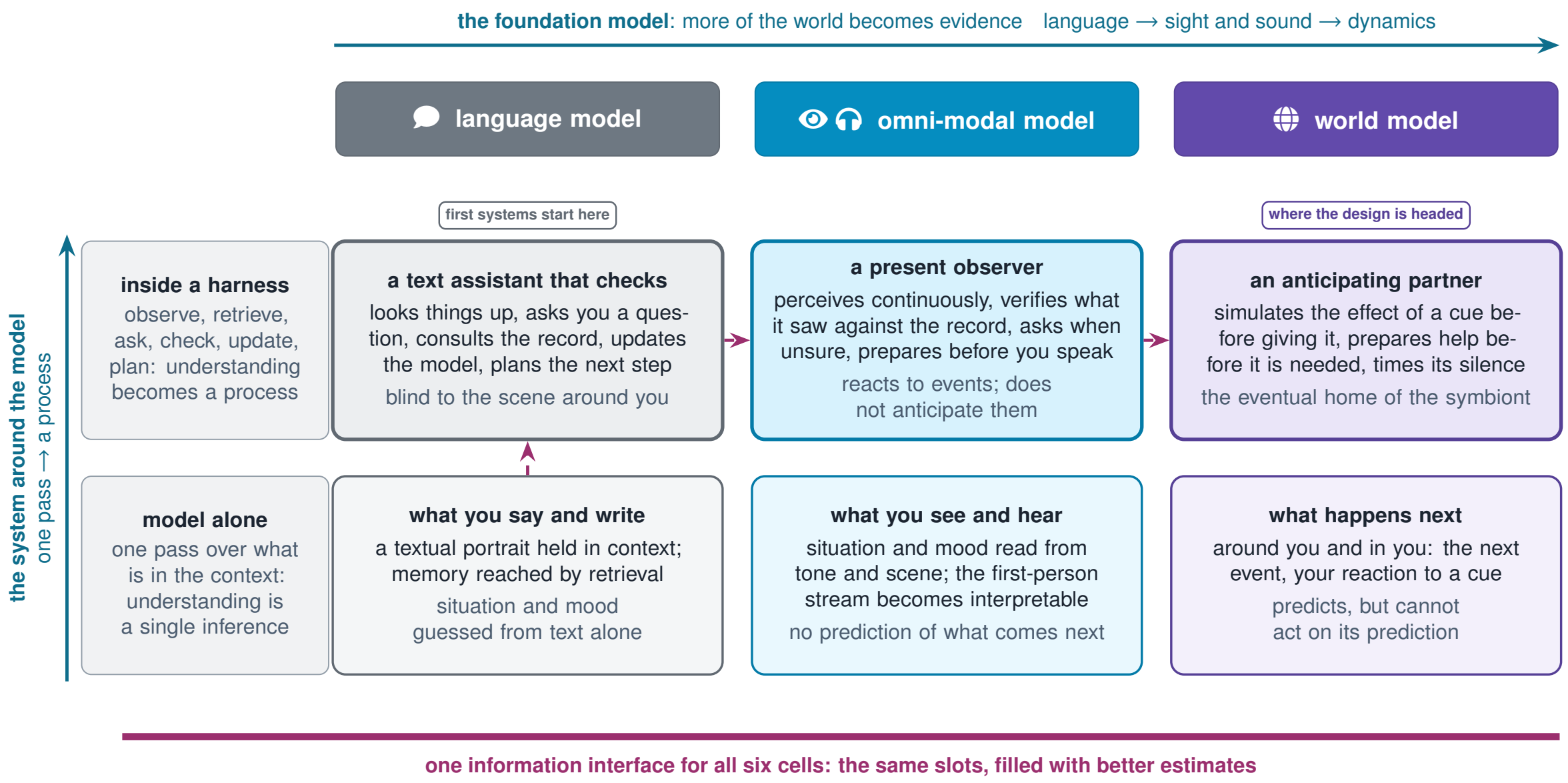

The personal model has to be computed by something, and the paper binds it to nothing in particular.
The claim is precise.
The personal model and the record are defined at the level of \emph{information}: what must be known, what must be predicted, and how confidence is expressed.
The backbone is the engine that turns evidence into that information, and the engine can be changed while the interface stays, so that what varies between engines is the quality of the estimates.
Two independent axes describe the engine (\figref{fig:backbones}).
The first axis is the foundation model itself, of which three kinds matter here.
A language model understands what the person says and writes, holds a textual portrait in its context, and reaches memory through retrieval.
An omni-modal model understands what the person sees and hears, reads situation and mood from tone and scene, and makes the first-person stream interpretable for the first time.
A world model turns understanding into prediction: what is about to happen around the person, how they will think and react, and whether a given reminder will be welcome.
The three are members of one family, each extending the evidence the model can take in and the inference it can draw from it, and later members contain the earlier ones.
The second axis is orthogonal to the first.
An agentic system belongs to this second axis rather than to the family: it is an organization of computation around a foundation model of any of the three kinds, the model together with the harness of Section~\ref{sec:agents}, and the model inside an agentic system can be exchanged for any member of the family.
What the harness adds is that understanding ceases to be a single pass and becomes a process: the system can look something up, ask the person a question, check the record, update the model, and plan the next step, whichever model sits inside it.
The grid of \figref{fig:backbones} therefore has six cells, and the two axes can be traversed independently.
The first usable systems will stand in the upper-left cell, a language model inside a harness, since that is what today's agents are, and the design is headed toward the upper-right cell, a world model inside a harness that anticipates as well as perceives.
\claim{The design must stand on the weakest cell and improve automatically along either axis.}
On a language model alone the symbiont is a text assistant with a memory; without a world model the intervention policy estimates reactions by heuristics; without a harness, understanding cannot check itself.
The information interface is what makes the claim concrete: the mental-state layer of \tabref{tab:layers} asks every cell for the same output, an estimate of intent and an estimate of emotion, each with its confidence, which a language model produces from dialogue text, an omni-modal model from tone and scene as well, a world model from predictions of what comes next, and a harness from any of these together with the result of checking them against the record.
This is also why Section~\ref{sec:stack} requires that a change of foundation model, or of harness, preserve the relationship.

\subsection{Two World Models}\label{sec:twoworldmodels}
A world model, as a backbone, contributes to the personal model in two forms.
An \keyterm{external physical world model} takes the environment seen in first person as its object, scene, objects, people present, their dynamics, and predicts the next moment: the cup is about to tip, the bus is arriving, the slide is about to change.
It serves help in the physical world, grounds mental cues in the situation (the person is tense because the deadline is on the screen), and supplies the extended reach of Section~\ref{sec:reach}; its sources are the video world models and first-person data of Section~\ref{sec:wm}.
An \keyterm{internal mental world model} takes the person's beliefs, attention, intentions, emotions, and goals as its object and predicts how they change after an event and after an action by the symbiont.
It is the predictive version of the mental-state layer of \tabref{tab:layers}, and its sources are theory-of-mind research, individual-level simulation~\citep{park2024simulation}, and mental world modeling, which places these variables inside the world state~\citep{fei2026mental}.
The two are coupled: physical events change mental states, and mental states drive actions in the physical world.

A competent world model has a state representation, temporal and relational reasoning, prediction of future states, prediction of the consequences of candidate actions, an expression of uncertainty, and detection of its own misspecification.
Replace ``the world'' with ``this person's world'' and every requirement stands, and the last one matters most.
People change, so a model of a person will go wrong; the ability to detect that it has gone stale is what separates a symbiont from a static profile, and it is the same class of problem as misspecification detection in open-world modeling.
In the smallest notation, with $z_t$ the joint latent state,
\begin{equation}\label{eq:belief}
z_t = \big(z^{\mathrm{phys}}_t,\, z^{\mathrm{mind}}_t\big), \qquad
b_t(z) = P\!\left(z_t \mid O_{\le t},\, R_{\le t},\, K_t\right), \qquad
\hat z_{t+1} \sim P\!\left(z_{t+1} \mid z_t,\, a_t,\, x_{t+1}\right),
\end{equation}
where $z^{\mathrm{phys}}_t$ is the task-relevant state of the world around the person, $z^{\mathrm{mind}}_t$ the latent state of intention, emotion, and goals, $b_t$ the system's belief over both, and the third expression says that the model must predict how both change after the system's action $a_t$ and the new situation $x_{t+1}$.
The notation fixes what is predicted and conditioned on; it prescribes no Bayesian implementation.
It underlies the intervention policy of Section~\ref{sec:policy}, which needs $b_t$, and the learning rule of Section~\ref{sec:flywheel}, which needs $a_t$.
Without a world model the third expression degrades to heuristic rules, and the system still stands.

\subsection{Knowing When It Is Wrong}\label{sec:limits}
``It knows you better than you know yourself'' has three defensible meanings and one that is not.
It can hold the person's whole self-context at once while they hold a small part (Definition~\ref{def:workingcontext}); it can see patterns across years that the person is inside of and cannot see; and it remembers the reasons they gave and have since forgotten.
The fourth meaning, reading thoughts that were never expressed, the paper gives up explicitly.
Two kinds of limitation must also be kept apart.
Familiarity with the person and competence on the problem in front of them are different things, and a system that knows someone very well may be poor at the specialist question they face this minute.
The confidence of the personal model and the reliability of the backbone in that domain are estimated and reported separately; when the second is low the system narrows its advice or suggests asking a person, and it never covers its ignorance of the problem with its familiarity with the person.
Understanding runs both ways, since the person also learns under which conditions the system is reliable, and mutual understanding has degrees and costs; the grounding framework of \citet{clark1991grounding} describes the process without the infinite regress of each knowing what the other knows.
Finally, every judgment about the person carries a confidence, any of them can be overturned by a sentence from the person, the system learns to ask when it is unsure, and an overturned judgment propagates through the repair of Section~\ref{sec:repair}.
\citet{collins2024machines} argued that machines that think with people must model them; this section says what is modeled, on what evidence, on which backbone, and within which limits.

\section{Living Alongside}\label{sec:living}
Understanding has to become participation at the right moment.
This section turns private, real-time, low-friction, embedded presence, and proactivity that is timely, into mechanisms.
It is about how the symbiont is present, when it speaks, and in what form.
What it says is the subject of Section~\ref{sec:ladder}.

\subsection{Two Conversations at Once}\label{sec:duplex}
\begin{figure}[t]
  \centering
  \resizebox{0.98\linewidth}{!}{\input{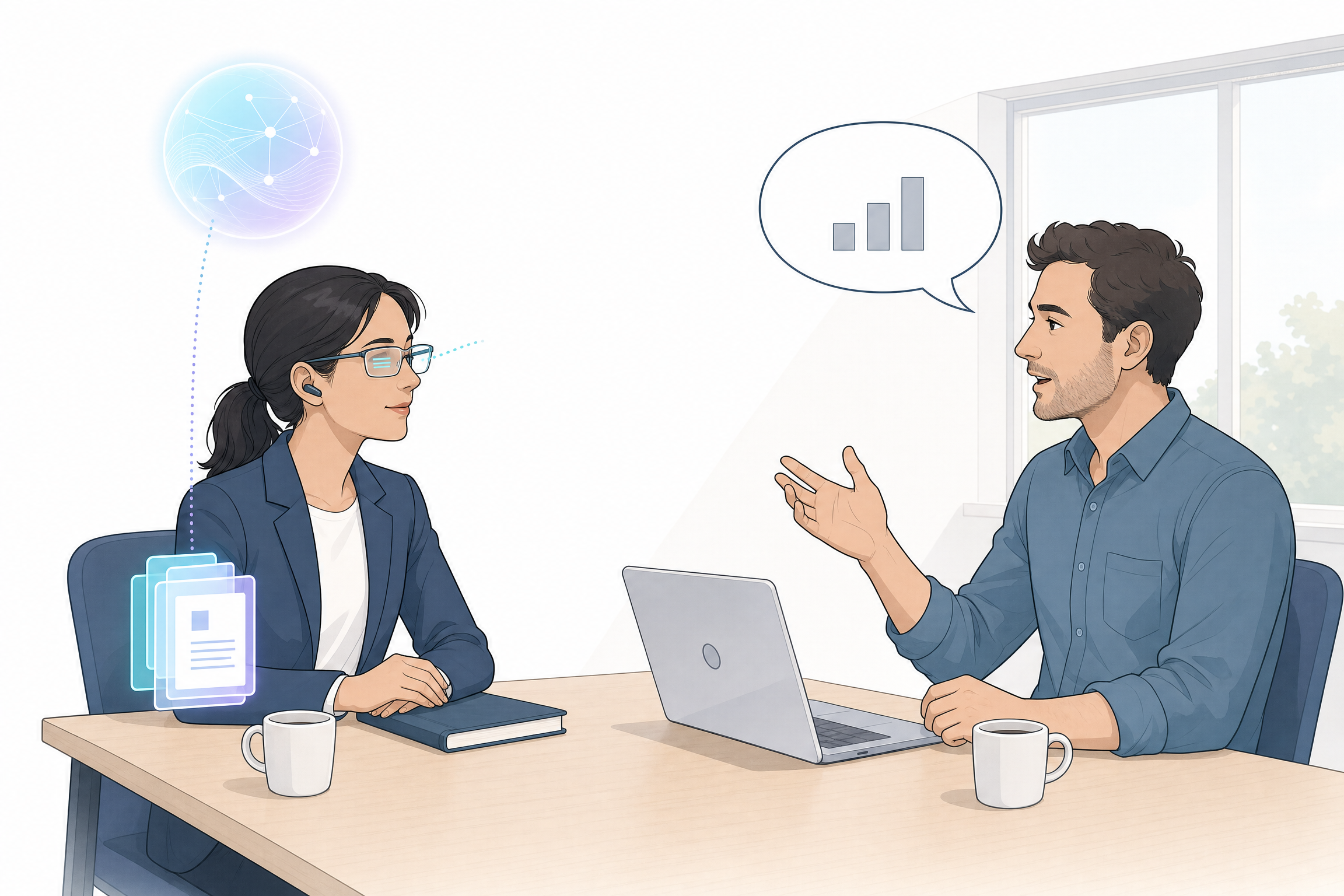}}
  \caption{Cognitive full-duplex.
  One exchange in a team meeting: a colleague raises a figure the person reported last quarter, the slow loop retrieves and checks it, the fast loop places a one-line cue in the glasses within a second, the person answers, and the cue is withdrawn once it expires.
  The timeline below follows the exchange across the four tracks of the external world, outward expression, the private channel, and the symbiont's internal state.
  The help is a reminder of the person's own figure, the kind of assistance most settings accept (Section~\ref{sec:conversations}), and most of the time the right output is silence.}
  \label{fig:duplex}
\end{figure}

Real-time voice assistants solved one problem: the person and the AI can speak and listen at the same time~\citep{defossez2024moshi}.
A symbiont solves a different one.
The person is talking to another person; the symbiont is listening, thinking, and preparing at the same time, and through the private channel it hands the person a line, so that the person never has to stop for it.
We call this \keyterm{cognitive full-duplex}, the real-time form of the symbiosis \citet{licklider1960symbiosis} imagined: two conversations at once, the person's with the world and the person's with the symbiont, with the second never interrupting the first.
\figref{fig:duplex} shows one instance.
A colleague mentions a figure the person reported last quarter; the slow loop retrieves it and checks it against the record; a question is detected and a cue appears in the glasses within a second; the person answers with the figure; the topic moves on and the cue expires and is withdrawn.
Four tracks run through the exchange, the external world, the person's outward expression, the private channel, and the symbiont's internal state, and the figure follows them along one timeline.
This capability belongs in the first usable system, since without it the symbiont is confined to the moments when the person is alone.

\subsection{Fast Loop, Slow Loop}\label{sec:loops}
Real time has two timescales, and conflating them leads to the mistaken demand that everything be low-latency.
The fast loop handles the immediate: a name, a number, ``what he just said contradicts last week.''
The slow loop handles the background: retrieval, reasoning, updating the model, preparing a summary for after the meeting.
Between them sits a rolling current state: what this conversation is about, who wants what, what is still unresolved.
Cues must update incrementally, an updated cue must replace its predecessor, and an expired cue must disappear.
The latency budget has a shape.
The fast loop's ceiling is set by the pace of conversation, since a name is useless two seconds late, so it can use only small models on the device or at the edge together with material the slow loop has already prepared.
The slow loop may be slow but must think ahead, preparing before the person opens their mouth what they are likely to need, which is where the prediction of Section~\ref{sec:twoworldmodels} is put to use.
Section~\ref{sec:openproblems} lists the end-to-end latency budget as an open problem.

\subsection{Implicit Intent}\label{sec:intent}
Much of what a person needs is never asked for.
The evidence for implicit intent comes from the current activity, recent behavior, unfinished business (Section~\ref{sec:future}), the personal model (Section~\ref{sec:understanding}), and the half-sentence the person whispers.
The purpose of inferring it is to have candidate help ready and to wait for a suitable moment, never to decide in the person's place.
When the inference is uncertain, the right move is one very short question, and the answer to that question is itself new evidence for the record; active clarification is a source of context in its own right.

\subsection{The Intervention Policy}\label{sec:policy}
Timeliness has to become a decision problem, and silence has to be one of the formal decisions.
\tabref{tab:policy} separates the three judgments: what to help with, when, and how.
The decision can be written as a constrained choice,
\begin{equation}\label{eq:policy}
a^*_t \;=\; \arg\max_{a \in \mathcal{A}_t}\; \E\!\left[\Delta\Util(a) \mid b_t\right] \;-\; \lambda_t \cdot \mathrm{Interrupt}(a,\, \mathrm{attention}_t) \;-\; \rho \cdot \mathrm{Risk}(a), \qquad \text{s.t.}\ \ \mathrm{conf}(b_t) \ge \theta(a),
\end{equation}
where $\mathcal{A}_t$ always contains silence $\varnothing$ with zero interruption cost, and the confidence threshold $\theta(a)$ rises with the intrusiveness of the action: \claim{the louder the help, the surer the symbiont has to be}, and acting on the person's behalf carries the highest threshold of all.
The three terms say that help must have expected value, that interruption has a cost which varies with the person's current attentional state, and that risk is counted separately because a wrong reminder is worse than none.
We add a notion of an \keyterm{attention budget}: the amount of a person's attention a symbiont may consume in a day is finite, and it should be managed the way battery charge is managed.
The just-in-time adaptive intervention framework of mobile health has studied for a decade when an intervention of what strength is acceptable, with its notions of states of vulnerability and receptivity~\citep{nahumshani2018jitai}, and those notions transfer directly.
Earlier work on interruption and recovery shows how expensive an ill-timed intrusion is~\citep{iqbal2007disruption}, and mixed-initiative principles already warned that initiative must be weighed against the cost of poorly timed action~\citep{horvitz1999mixed}.
The goal is therefore accuracy of intervention rather than a greater amount of it.
The weights $\lambda$, $\rho$, and $\theta$ are given no numerical values in this paper and are research variables of Section~\ref{sec:agenda}.

\begin{table}[t]
\centering\footnotesize
\caption{The three decisions of the intervention policy. Silence is always one of the answers to \emph{how}, and the last column makes it the default.}
\label{tab:policy}
\setlength{\tabcolsep}{5pt}\renewcommand{\arraystretch}{1.28}
\begin{tabularx}{\linewidth}{@{}L{1.9cm}YYL{2.6cm}@{}}
\toprule
\tabhead \colhead{Question} & \colhead{What must be judged} & \colhead{Evidence} & \colhead{Default when unsure} \\
\midrule
\ib{tbPink}{\faQuestion}\ \rowhead{What} & what you missed, misremembered, did not think of, or cannot put into words; what could simply be done for you & personal model \lyM\lyS\lyE\ + the current situation + the record & nothing \\
\rowcolor{tabStripe}\ib{tbGreen}{\faClock}\ \rowhead{When} & whether speaking now pays, whether it still will in a few seconds, whether you can take an interruption now & time window + attention load + setting \lyS\lyH & later, or never \\
\ib{tbTeal}{\faVolumeDown}\ \rowhead{How} & a line of text, a sentence, a question back, an overlay in your view, a silent note for later, or an action on your behalf & setting + channel (\tabref{tab:channels}) + your preferences about help \lyH\ + risk & \chSilence\ silence, or a silent note \\
\bottomrule
\end{tabularx}
\tablegend{\textbullet\ Layer letters as in \tabref{tab:layers}: \lyS\ situation, \lyM\ mental state, \lyE\ experience, \lyH\ preferences about help.}
\end{table}

\subsection{Low Friction}\label{sec:friction}
Low friction means that the person never works for the help: no re-explaining the background, no reaching for a phone, no switching applications, no waiting.
Friction has four sources, and each is removed elsewhere in the paper: explaining the background (the record, Section~\ref{sec:context}), switching devices (the runtime, Section~\ref{sec:runtime}), waiting for a reply (the two loops above), and tidying up afterwards (consolidation, Section~\ref{sec:streams}).
Some friction is worth preserving.
Being asked one more question before an important decision, or being left to think for a moment, is friction that should survive optimization (Section~\ref{sec:personchanges}); interfaces that make people examine a recommendation before accepting it reduce overreliance at the cost of being liked less~\citep{bucinca2021trust}, and that trade is the one a symbiont must learn to make per person and per moment.

\subsection{Beyond the Meeting Room}\label{sec:beyondmeeting}
The meeting is the first proving ground: time pressure is high, information is dense, and timing decides everything, which is why a desktop copilot that analyzes a meeting as it happens and feeds cues back in real time is the natural starting point.
The symbiont must, however, leave the room.
Three settings of presence have different rhythms and channels.
Sitting at a screen, in the desktop or the editor, the channel is wide and help can be expanded.
In conversation, in a meeting or at dinner, only a line or a whispered word fits.
On the move, in the street, in transit, in an unfamiliar city, the earpiece and the overlay carry what the moment needs.
The second moment of the day in Section~\ref{sec:intro} resolves here: the result the researcher could not recall arrives as a line in the glasses while the conversation continues.
What the symbiont does once present begins with doing things for the person, which is where the next section starts.

\section{The Ladder of Assistance}\label{sec:ladder}
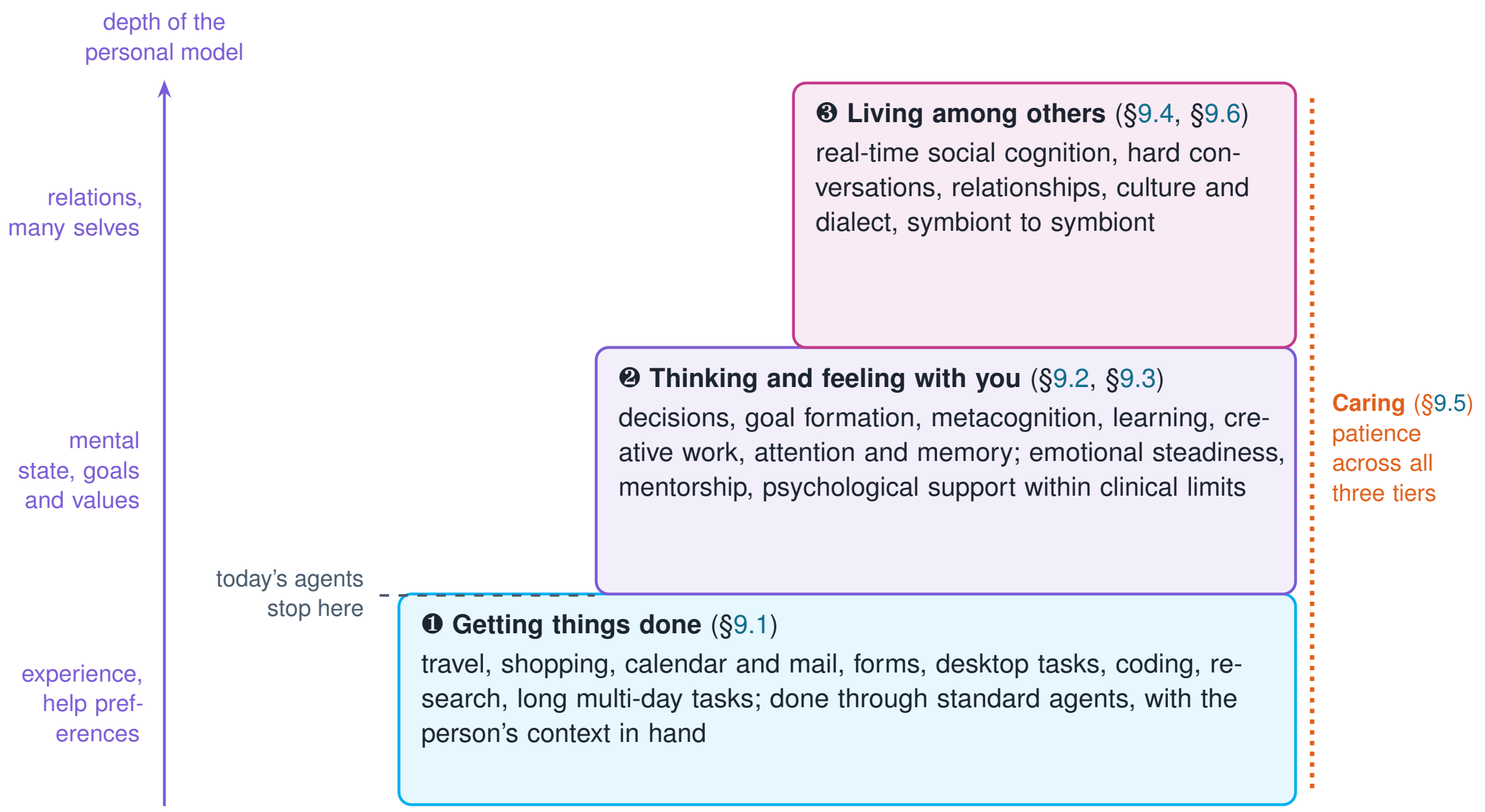
\begin{figure}[t]
  \centering
  \resizebox{0.98\linewidth}{!}{\begin{tikzpicture}[x=1cm,y=1cm,font=\small\sffamily,
  tierbox/.style={draw=#1,fill=#1!10,rounded corners=4pt,line width=0.9pt,inner sep=5pt,align=left}]
  \node[tierbox=symCyan,minimum width=10.0cm,minimum height=2.35cm,anchor=south west] (t1) at (0,0) {};
  \node[tierbox=symViolet,minimum width=7.8cm,minimum height=2.75cm,anchor=south west] (t2) at (2.2,2.35) {};
  \node[tierbox=symMagenta,minimum width=5.6cm,minimum height=2.95cm,anchor=south west] (t3) at (4.4,5.1) {};
  \node[anchor=north west,text width=9.6cm] at ([xshift=4pt,yshift=-3pt]t1.north west) {\tierone\ \textbf{Getting things done} (\S\ref{sec:tier1})\\[1.5pt]travel, shopping, calendar and mail, forms, desktop tasks, coding, research, long multi-day tasks; done through standard agents, with the person's context in hand};
  \node[anchor=north west,text width=7.4cm] at ([xshift=4pt,yshift=-3pt]t2.north west) {\tiertwo\ \textbf{Thinking and feeling with you} (\S\ref{sec:thinking}, \S\ref{sec:feeling})\\[1.5pt]decisions, goal formation, metacognition, learning, creative work, attention and memory; emotional steadiness, mentorship, psychological support within clinical limits};
  \node[anchor=north west,text width=5.2cm] at ([xshift=4pt,yshift=-3pt]t3.north west) {\tierthree\ \textbf{Living among others} (\S\ref{sec:relating}, \S\ref{sec:handshake})\\[1.5pt]real-time social cognition, hard conversations, relationships, culture and dialect, symbiont to symbiont};
  \draw[dashed,symSlate,line width=0.8pt] (-0.2,2.35) -- (2.2,2.35);
  \node[anchor=east,text=symSlate,align=right,font=\footnotesize\sffamily] at (-0.25,2.35) {today's agents\\stop here};
  \draw[stGrow,line width=1.4pt,dotted] (10.2,0.2) -- (10.2,7.9);
  \node[anchor=west,text=stGrow,text width=1.7cm,align=left,font=\footnotesize\sffamily] at (10.3,4.0) {\textbf{Caring} (\S\ref{sec:caring})\\patience across all three tiers};
  \draw[-{Stealth[length=2mm]},stUnderstand,line width=0.9pt] (-2.6,0) -- (-2.6,8.1);
  \node[anchor=south,text=stUnderstand,align=center,font=\footnotesize\sffamily] at (-2.6,8.15) {depth of the\\personal model};
  \node[anchor=east,text=stUnderstand,text width=1.6cm,align=right,font=\footnotesize\sffamily] at (-2.75,1.15) {experience, help preferences};
  \node[anchor=east,text=stUnderstand,text width=1.6cm,align=right,font=\footnotesize\sffamily] at (-2.75,3.75) {mental state, goals and values};
  \node[anchor=east,text=stUnderstand,text width=1.6cm,align=right,font=\footnotesize\sffamily] at (-2.75,6.6) {relations, many selves};
\end{tikzpicture}}
  \caption{The ladder of assistance.
  Tier~1 is where today's agents stop; a symbiont does the same tasks with the person's context in hand.
  Tier~2 is thinking and feeling with the person, and Tier~3 is living among others.
  The depth of the personal model required grows with each tier, and caring runs across all three.}
  \label{fig:ladder}
\end{figure}

Once present, what does a symbiont do?
The answer is a ladder with three tiers (\figref{fig:ladder}).
At the bottom it gets things done, which is what today's autonomous and personal agents do, and a symbiont must do it as well as they do and better.
In the middle it thinks and feels with the person: decisions, goal formation, learning, creative work, emotional steadiness, mentorship, and psychological support within clinical limits, the settings of highest cognitive density.
At the top it lives among others with the person: real-time social cognition, hard conversations, culture and dialect, and the meeting of two symbionts.
Three facts organize the ladder.
Each tier rests on the one below, and a symbiont that cannot book a flight will never be worn.
Each tier up moves the object of evaluation further from task completion and closer to how the person is changed.
And all three tiers run on the same machinery of Sections~\ref{sec:context} to~\ref{sec:living}, differing only in how deep into the personal model they reach and how much intervention they are allowed (the threshold $\theta$ of \eqnref{eq:policy}).
One function, patient care, runs across all three tiers and receives its own treatment in Section~\ref{sec:caring}.

\subsection{Getting Things Done}\label{sec:tier1}
The floor of the ladder is general-purpose assistance: booking travel and hotels, shopping and comparing prices, keeping the calendar and the inbox, forms and payments, desktop work with files, slides, and spreadsheets, coding and technical work inside the editor, retrieving and organizing information, and multi-day tasks such as an application, a move, or an event.
A symbiont does not rebuild these capabilities.
Through its harness (Sections~\ref{sec:agents} and~\ref{sec:harness}) it delegates them to standard agents and tools, computer-use agents, tool protocols, and coding agents, and keeps three jobs for itself: defining the task correctly, supplying the right personal defaults, and deciding when the person must be consulted.
The reason the same task comes out better is the record and the personal model.
An agent told ``book my flight to the conference'' has to ask where the person leaves from, what they will spend, when they prefer to travel, whether they want a window seat.
A symbiont already knows that the person has a meeting on Monday morning that cannot move, that they dislike overnight flights, that they want to stay near the venue because they hope to meet a particular colleague, and that the last time they took a short connection they missed it.
So it asks nothing, or it asks the one question that only the person can answer.
One layer deeper, it knows \emph{why} the person is making the trip, from the goals layer of \tabref{tab:layers}, so it frames the task correctly and books the meeting they were hoping for alongside the flight.
\tabref{tab:sametask} works through five everyday tasks in this way, and its last column lists what can be measured: clarifying questions asked, corrections the person had to make, time to completion, how much of a plan actually holds.

\begin{table}[t]
\centering\footnotesize
\caption{The same task, done by an agent and by a symbiont. The task is unchanged; what changes is how much must be asked, how the task is framed, and how often the result has to be corrected. The last column makes the difference measurable.}
\label{tab:sametask}
\setlength{\tabcolsep}{4pt}\renewcommand{\arraystretch}{1.22}
\begin{tabularx}{\linewidth}{@{}L{2.35cm}C{1.0cm}YYYL{2.6cm}@{}}
\toprule
\tabhead \colhead{Task} & \colhead{Gap closed} & \colhead{An agent \lb{tbTeal}{A} must be told} & \colhead{A symbiont \lb{tbPink}{S} already knows} & \colhead{Difference in outcome} & \colhead{Measurable} \\
\midrule
\ico{\faPlaneDeparture}\ Book a trip to a conference & \gF\gT & dates, budget, preferences, companions & the meeting that cannot move, your aversion to red-eye flights, the person you hope to meet, the connection you missed last time & right the first time; the meeting gets scheduled alongside & clarifying questions; rebookings; your later edits \\
\rowcolor{tabStripe}\ico{\faEnvelope}\ Reply to a difficult email & \gF\gW & who the recipient is, what you want to say & your history with them, the last disagreement, your usual register & the draft needs no rewrite & adoption rate; edit distance \\
\ico{\faCode}\ Fix a bug & \gF\gT & the repository, how to reproduce & the module you changed last week, your coding habits, that you have half an hour today & a minimal fix now, the refactor left for when you have time & time to done; rollback rate \\
\rowcolor{tabStripe}\ico{\faShoppingCart}\ Buy a gift & \gF\gN & recipient, budget, taste & the wish the person once mentioned in your record, the stage of your relationship & no deliberation required of you & number of your changes \\
\ico{\faCalendar}\ Plan next week & \gT\gF & which items, durations, priorities & your rhythm (afternoons are slow), what you promised (\S\ref{sec:future}), how your body is & a plan that looks like you & how much of it actually holds \\
\bottomrule
\end{tabularx}
\tablegend{\textbullet\ Gap closed, in the four losses of Section~\ref{sec:intro}: \gF\ forgetting\quad \gN\ not noticing\quad \gT\ not thinking through\quad \gW\ not finding the words.}
\end{table}

Acting for the person is the most intrusive action in \eqnref{eq:policy} and needs the firmest boundary.
We propose five levels of action permission: look, suggest, draft, execute what can be undone, and execute what cannot, such as paying, sending, and committing; each level is earned along the trust curve of Section~\ref{sec:trustcurve}.
A draft composed in discussion, a line rehearsed for a conversation, or a candidate plan never becomes an outward action or commitment on its own.
Long tasks carry state, verify their results, and hand back to the person when they fail.
Evaluation at this tier can keep task completion as its object as long as it adds a \keyterm{personalization differential}: the same task, the same base model, with and without the person's context, which Section~\ref{sec:measure} writes as \eqnref{eq:pers}.
Current benchmarks measure the floor and say nothing about the differential~\citep{xie2024osworld,zhou2024webarena,yao2025taubench}.
We make no claim of novelty for this tier; the single claim is that the same tasks are done better with personal context, and that this can be shown.

\subsection{Thinking With You}\label{sec:thinking}
The first half of the second tier is cognition itself, in six forms.

\paragraph{Decisions}
A symbiont that remembers the person can help them reason with themselves.
It returns what they forgot (why they declined last time, and whether the constraint still holds), lays out what they did not think of (a third option, a counterexample, a value they usually weigh and this time left out), and writes down what they concluded, with the reasons, for next time (Section~\ref{sec:future}).
Its stance is to help the person see, never to choose for them, and it owes them a reasoned objection when a choice conflicts with a long-term goal they stated themselves, after which the decision returns to them.
\citet{reicherts2025think} compared an AI that builds on the person's own reasons with one that gives recommendations: the first integrated better with people's thinking and produced slightly better outcomes, the second surfaced more novel insights at less effort, and the first is the form of help this paragraph describes.
The third moment of the day in Section~\ref{sec:intro} resolves here.

\paragraph{Goal formation}
Many important matters start without an executable goal: changing direction, taking or declining an opportunity, handling a relationship.
The symbiont can unfold candidate goals, bring back the reasons that conflict, and ask which cost the person would find hardest to bear, so that a provisional choice forms in the conversation.
It makes no claim to have discovered the person's ``true'' goal.

\paragraph{Metacognition}
This is help with knowing where one is uncertain, whether to check, and whether this is a moment to think alone, to think with the symbiont, or to ask a person.
Such help arrives before an answer, and sometimes it takes the form ``that is not the question you need.''
A rise in confidence is never counted as a rise in understanding, and the cognitive-offloading literature~\citep{risko2016offloading} is the background for when offloading is the right choice.

\paragraph{Learning and tutoring}
A tutor who knows where the person does not understand, because it has seen every place they got stuck, can give a hint at the moment they need it instead of an answer, and can step back when they say they want to learn this themselves (Section~\ref{sec:personchanges}).
Language learning is the most natural case and connects to the dialect help of Section~\ref{sec:relating}.

\paragraph{Creative and long-horizon work}
Two services belong here: catching the idea that slips away (the first moment of the day in Section~\ref{sec:intro}, the typical use of the whisper channel), and holding a thread across weeks, where this project got to, which hypothesis was dropped, which is still open, which is the working state of Section~\ref{sec:timescales}.
The failure to watch for is a symbiont that generates a great deal and drifts away from the question the person wanted to pursue.

\paragraph{Attention and memory}
This includes guarding focus, by noticing that the person has been refreshing the same page for ten minutes, and the plainest memory support of all: a name, a detail, where something was put, what was said last time.
These six forms draw on the goals and mental-state layers of \tabref{tab:layers}, use the expanded dialogue or the whisper channel, and span seconds to days, and it is here that the friction worth preserving (Section~\ref{sec:friction}) matters most.

\subsection{Feeling With You}\label{sec:feeling}
The second half of the tier runs from everyday emotion to the edge of clinical care, in four steps of increasing weight.

\paragraph{Everyday steadiness}
The small disturbances of a day: a remark that stung, nerves before a talk, a piece of bad news.
The symbiont's help is to be present, to know what happened, and to say the one sentence that fits, or to say nothing and simply note it.
Physiology gives early signs of stress and fatigue, and their interpretation has to remain conservative.

\paragraph{A mentor who knows your story}
Useful comfort is never generic encouragement; it rests on a precise understanding of the person's actual history, so that what is said touches the real point: this failure relates to those efforts and those hopes.
Three levels follow.
Being heard, when the symbiont restates the situation and the person thinks ``yes, that is it.''
Being understood, when it points out an attribution error the person could not see.
Being moved, when it builds a next step from the way the person got through a similar time before.
The preconditions, an established relationship of trust and a general competence that in many domains exceeds an ordinary person's, are stated as conditions throughout.

\paragraph{Psychological support within clinical limits}
This is where the section has to be most careful.
Relative to any mental-health application, a symbiont has three advantages: it knows the person's life and does not need it retold; it is present when the trigger happens, days before the next appointment; and it can see patterns, in sleep, in social withdrawal, in language.
It is therefore suited to three things: being a daily practice partner for evidence-based methods such as the identification and reframing used in cognitive behavioral therapy; providing support and a record between professional sessions; and recognizing early signs of relapse.
It is not suited to carrying treatment on its own.
For depression and anxiety the paper says that this is a direction that deserves serious study, with randomized evidence for particular systems under particular conditions, from the early Woebot trial~\citep{fitzpatrick2017woebot} to the recent trial of a generative chatbot~\citep{heinz2025therabot}, and it does not say that these conditions can be treated by an AI.
The symbiont's first duty in such moments is to recognize when a human professional must be brought in, with a clear path for doing so when crisis signals appear, and in regulated jurisdictions this function is itself regulated, which the paper acknowledges.

\paragraph{Habits, behavior change, and company}
Long-term behavior change, in sleep, exercise, or recovery, is the home ground of just-in-time adaptive interventions~\citep{nahumshani2018jitai}, and a symbiont turns the phone notification into a sentence from someone present.
As company it can be patient and non-judging, and the risk has to be stated directly: it must never substitute for human relationships, and it should steer the person back toward people.
Studies of companion chatbots show both relief of loneliness and dependence, with effects that vary with how and how much they are used~\citep{maples2024loneliness,fang2025psychosocial}.

\begin{scenario}[label={box:setback}]{After a Setback}
\small The same sentence, ``I messed up again'', and three responses a symbiont might give.
It listens and makes a note, and says nothing.
It looks back at the facts and points out what differs between this time and last time.
With the person's agreement, it works through the causes and a next step together with them.
Three things it does not do: offer a paragraph of generic encouragement; insist ``I know you better'' after the person has rejected an interpretation; and continue alone when a crisis signal appears.
\end{scenario}

\subsection{Relating}\label{sec:relating}
The third tier is help with people and cultures, in four forms.
\emph{Real-time social cognition.}
A meeting is one kind of social setting.
In any of them the symbiont can recall the person opposite (where the conversation left off, what they care about), read the moment (the implication of what was just said, the tension between the person's argument and the other's position), and help the person say it (a better phrasing, the number they forgot).
Reading the moment relies on the theory-of-mind ability of large foundation models, which has shown both promise and brittleness~\citep{kosinski2024evaluating,ullman2023large,strachan2024testing}, and on inferences about other people that are used only for the interaction at hand and never accumulate into a portrait (Sections~\ref{sec:others} and~\ref{sec:bystanders}).
\emph{Rehearsal and hard conversations.}
Practice beforehand for a salary negotiation, a breakup, or negative feedback to a colleague; a line during (the person's pace is rising; the other is really asking something else); a review afterwards.
A simulated counterpart in rehearsal is only a rehearsal and never a prediction of a real person.
De-escalation belongs here too: the symbiont can see, in voice and heart rate, that the person is heating up before they notice, and a single vibration on the private channel is enough.
\emph{Relationship maintenance.}
This goes deeper than a birthday reminder: the person has not spoken to their father in a month and he mentioned a medical test last time.
It draws on the relations layer and on prospective memory (Section~\ref{sec:future}), and it is where the limits of Section~\ref{sec:bystanders} bite hardest.
\emph{Language, culture, and the dialect self.}
Behind each language are different habits of thought and social rules.
The help consists less in translation than in entering the frame: how people here greet, how they decline, what cannot be said directly.
It needs the context-indexed model of Section~\ref{sec:layers}, which knows who the person is in that language, and the online learning of Section~\ref{sec:sparse}, which learns the dialect of a small place.
Those two sections share one example, used here and learned there, and we avoid claims of the form ``language X makes people more open'': shifts of this kind have been reported in bicultural individuals~\citep{ramirezesparza2006bilinguals}, and they do not license a rule about any particular person.

\subsection{Caring}\label{sec:caring}
The function the user of a symbiont may value most is one that is patient when they are not.
Its core is consistency, patience, tirelessness, and noticing what people stop noticing, and it runs across all three tiers.
Four settings follow; the populations themselves are treated in Section~\ref{sec:whom}.
In \emph{parenting}, a symbiont answers a child's hundredth ``why'' without weariness, hands the parent a sentence when the parent's own temper is rising, leaves the child to the parent, and notices developmental or emotional signs worth a professional's attention; the symbiont belongs to the parent, and all-day sensing of children raises separate ethical questions (Sections~\ref{sec:notfirst} and~\ref{sec:ethics}).
In \emph{aging and memory decline} it is a memory prosthesis, who this is, whether the medication was taken, what was being looked for, with the SenseCam studies of memory-impaired patients as precedent~\citep{berry2007sensecam}; it supports orientation and safety and connects the person to family.
For \emph{caregivers}, the people who care for others and receive the least support, it holds the detail of care so that they can rest.
In \emph{rehabilitation and chronic self-management}, where the task is long, repetitive, and easy to abandon, its patience is worth most, and the handoff to clinical systems must be explicit.
The failure mode common to all four is the handing over of a person to a machine, and in each setting the symbiont's last duty is to return the person to other people.

\subsection{Symbiont to Symbiont}\label{sec:handshake}
\begin{figure}[t]
  \centering
  \resizebox{0.9\linewidth}{!}{\begin{tikzpicture}[x=1cm,y=1cm,font=\scriptsize\sffamily]
  \foreach \cx/\side in {-3.6/A,3.6/B}{
    \node[circle,draw=symInk,fill=white,minimum size=0.9cm,line width=0.8pt] (p\side) at (\cx,0) {\side};
    \foreach \r/\col/\lab in {2.6/symGray/visible: ``I am here'',2.1/symCyan/discoverable: public tags,1.6/symViolet/shareable: approved interests and contact,1.1/symMagenta/committable: can book a time}{
      \draw[\col,line width=0.8pt] (\cx,0) circle (\r);}
    \node[draw=symInk,fill=symInk!8,rounded corners=2pt,inner sep=1.5pt,text width=1.3cm,align=center] at (\cx,-0.8) {inner model never crosses};
  }
  \draw[symViolet,dashed,line width=1pt,{Stealth}-{Stealth}] (-1.9,0.6) -- (1.9,0.6) node[pos=0.5,above,text=symViolet] {exchange within the authorized ring};
  \node[anchor=west,text width=3.6cm,align=left] at (6.4,1.6) {\textcolor{symGray}{$\circ$} visible\\\textcolor{symCyan}{$\circ$} discoverable\\\textcolor{symViolet}{$\circ$} shareable\\\textcolor{symMagenta}{$\circ$} committable};
  \node[anchor=west,text width=3.6cm,align=left,text=symSlate] at (6.4,-0.9) {settings: a chance encounter; finding someone at a venue; preparing before a meeting};
\end{tikzpicture}}
  \caption{The symbiont handshake.
  Four rings of authorization around each person, from visible to committable; exchange happens only within the ring both have opened, and the inner model never crosses.}
  \label{fig:handshake}
\end{figure}
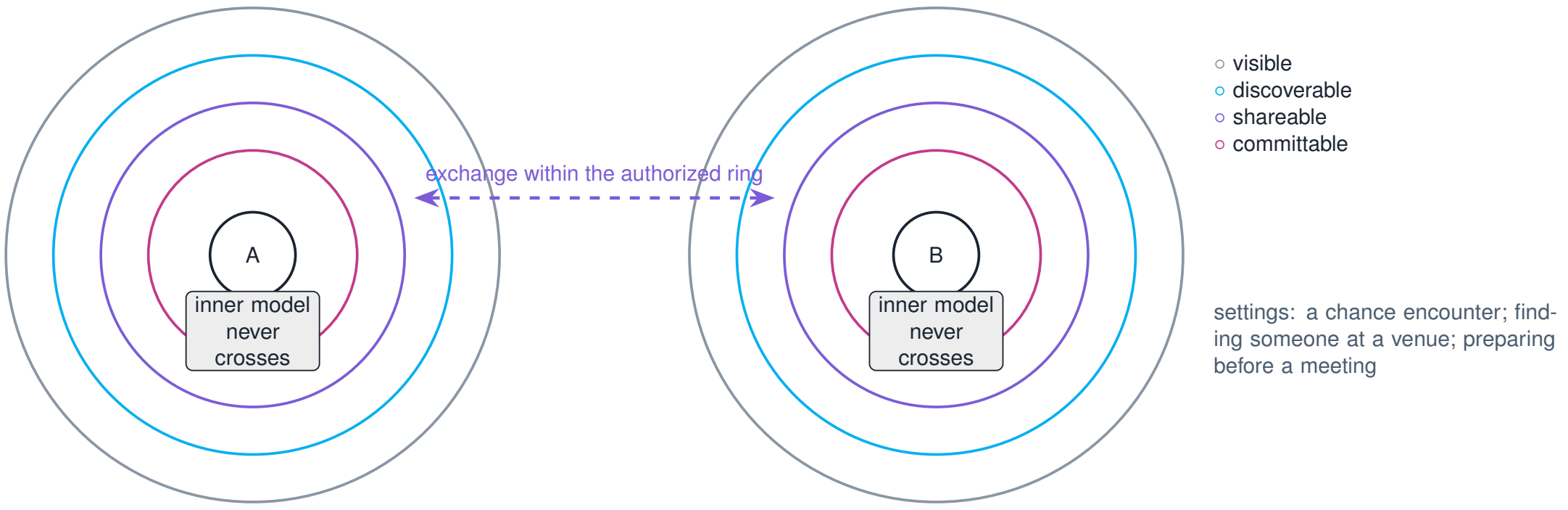

The most imaginative scene in the vision is two people, each with a symbiont, in one room and not yet near each other, while their symbionts exchange cards, filter shared interests, decide whether a greeting is worth it, and if so bring the two together.
It is at once fanciful and practical, and \figref{fig:handshake} turns it into four research questions.
Which information may represent the person in an exchange, in four rings of authorization: visible, discoverable, shareable, and committable, with the inner personal model never crossing the boundary.
By what criterion symbionts match, and who sets it.
Whether a greeting made on the person's behalf counts as their commitment.
And when two symbionts each optimize for their owner, whether the outcome is good for both.
Section~\ref{sec:networks} asks what happens when this scales to a crowd.
No protocol is proposed, and everything in Sections~\ref{sec:sensing} to~\ref{sec:living} stands even if symbionts never talk to each other.

\subsection{A Catalogue of Functions}\label{sec:catalogue}
\tabref{tab:catalogue} lists every function of this section with the layers of the personal model it needs, its channel, its timescale, the failure that would best distinguish a good system from a bad one, and its maturity.
The table is deliberately long: it is the full functional claim of the paper, each row has to find its population in Section~\ref{sec:whom} and its evaluation in Section~\ref{sec:agenda}, and it is what we mean when we say the framework is testable.

\begin{table}[tp]
\centering\scriptsize
\caption{A catalogue of cognitive-level functions as tests of the framework. Every row names the layers of the personal model it needs, the channel it uses, its timescale, the failure that would best discriminate a good system from a bad one, and its maturity. Section~\ref{sec:whom} maps rows to populations and Section~\ref{sec:agenda} to evaluation.}
\label{tab:catalogue}
\setlength{\tabcolsep}{3pt}\renewcommand{\arraystretch}{1.2}
\begin{tabularx}{\linewidth}{@{}L{3.1cm}C{0.75cm}VC{1.5cm}C{1.35cm}L{1.75cm}VYL{1.6cm}@{}}
\toprule
\tabhead \colhead{Function} & \colhead{Tier} & \colhead{Layers} & \colhead{Channel} & \colhead{Timescale} & \colhead{Most diagnostic failure} & \colhead{Maturity} \\
\midrule
\multicolumn{7}{@{}l}{\lb{tbTeal}{1}\ \textbf{Getting things done}}\\
Travel, shopping, calendar, mail & \tierone & \lyE\lyH & \chScreen\chVoice & minutes--days & asks what you already told it & \mnow \\
\rowcolor{tabStripe}Desktop and office tasks & \tierone & \lyE\lyS & \chScreen & minutes--hours & gets the task right and your intent wrong & \mnow \\
Coding and technical work & \tierone & \lyE\lyS\lyH & \chIDE & minutes--weeks & ignores your time limit and refactors everything & \mnow \\
\rowcolor{tabStripe}Retrieval and research & \tierone & \lyE\lyG & \chScreen\chVoice & minutes & finds the right thing and misses what you meant & \mnow \\
Multi-day long tasks & \tierone & \lyE\lyG & \chScreen\chAfter & days--weeks & fails midway and does not hand back & \mres \\
\midrule
\multicolumn{7}{@{}l}{\lb{tbViolet}{2}\ \textbf{Thinking with you}}\\
Decision support & \tiertwo & \lyG\lyI\lyE & \chDialog & days & overrules you with an inferred goal & \mres \\
\rowcolor{tabStripe}Goal formation & \tiertwo & \lyG\lyI & \chDialog & days & claims to have found your ``true'' goal & \mres \\
Metacognition & \tiertwo & \lyE\lyM & \chLine\chDialog & seconds--minutes & mistakes confidence for understanding & \mres \\
\rowcolor{tabStripe}Learning and tutoring & \tiertwo & \lyE\lyG & \chDialog\chLine & minutes--months & gives the answer instead of the hint & \mres \\
Creative and long-horizon work & \tiertwo & \lyE\lyS & \chWhisper\chText & slow loop & drifts from the question you wanted to pursue & \mres \\
\rowcolor{tabStripe}Attention and memory prosthesis & \tiertwo & \lyS\lyE & \chLine\chBuzz & fast loop & interrupts more than it helps & \mnow \\
\midrule
\multicolumn{7}{@{}l}{\lb{tbViolet}{2}\ \textbf{Feeling with you}}\\
Everyday emotional steadiness & \tiertwo & \lyM\lyE & \chVoice\chSilence & hours & a confident explanation built on a wrong portrait & \mres \\
\rowcolor{tabStripe}A mentor who knows your story & \tiertwo & \lyM\lyE\lyI & \chVoice & hours--days & generic encouragement; insisting ``I know you better'' & \mres \\
Psychological support within clinical limits & \tiertwo & \lyM\lyE & \chVoice\chDialog & days--months & handles a crisis signal alone & \mres\ \chipx{tbBlue}{regulation} \\
\rowcolor{tabStripe}Early signs of stress and burnout & \tiertwo & \lyM\lyS & \chLine & days & reads one sleepless night as depression & \mres \\
Habits and behavior change & \tiertwo & \lyG\lyS & \chLine\chBuzz & weeks--months & nags & \mnow\ \chipx{tbGray}{JITAI} \\
\rowcolor{tabStripe}Companionship and loneliness & \tiertwo & \lyM\lyR & \chVoice & continuous & substitutes for people instead of returning you to them & \mres\ \chipx{tbPink}{ethics} \\
\midrule
\multicolumn{7}{@{}l}{\lb{tbPink}{3}\ \textbf{Living among others}}\\
Real-time social cognition & \tierthree & \lyS\lyR\lyE & \chLine & fast loop & the right content at the wrong moment & \mres \\
\rowcolor{tabStripe}Rehearsal and hard conversations & \tierthree & \lyR\lyM & \chDialog\chLine & before, during, after & treats a simulation as a prediction & \mres \\
Relationship maintenance & \tierthree & \lyR\lyE & \chLine & weeks--months & builds portraits of other people & \mres\ \chipx{tbPink}{ethics} \\
\rowcolor{tabStripe}De-escalation & \tierthree & \lyM\lyS & \chBuzz\chLine & fast loop & a false alarm makes you tenser & \mres \\
Language, culture, and dialect & \tierthree & \lyI\lyR & \chLine\chAfter & fast + slow loop & applies a group pattern to the person in front of you & \mres \\
\rowcolor{tabStripe}Symbiont to symbiont & \tierthree & \chipx{tbGray}{shareable ring} & \chAI & seconds & exchanges inner information across the boundary & \mfar \\
\midrule
\multicolumn{7}{@{}l}{\lb{tbOrange}{C}\ \textbf{Caring, across tiers}}\\
Patience in parenting & \tierall & \lyR\lyM & \chVoice\chLine & continuous & manages the child instead of supporting the parent & \mres\ \chipx{tbPink}{ethics} \\
\rowcolor{tabStripe}Aging and memory decline & \tierall & \lyE\lyS & \chVoice\chLine & continuous & replaces the family & \mres\ \chipx{tbGray}{precedent} \\
Caregiver support & \tierall & \lyE\lyG & \chScreen\chVoice & continuous & adds to the load instead of taking from it & \mres \\
\rowcolor{tabStripe}Rehabilitation and chronic self-care & \tierall & \lyG\lyS\lyE & \chLine & months & drifts away from clinical care & \mres\ \chipx{tbBlue}{regulation} \\
\bottomrule
\end{tabularx}
\tablegend{\textbullet\ Tier: \nc{1}\ getting things done\quad \nc{2}\ thinking and feeling with you\quad \nc{3}\ living among others\quad \nc{1}\nc{2}\nc{3}\ across tiers.\quad
\textbullet\ Layers of the personal model (\tabref{tab:layers}): \lyI\ identity\ \lyE\ experience\ \lyS\ situation\ \lyM\ mental state\ \lyR\ relations\ \lyG\ goals\ \lyH\ preferences about help.\quad
\textbullet\ Channel: \chScreen\ screen\ \chVoice\ in-ear voice\ \chIDE\ IDE\ \chDialog\ expanded dialogue\ \chLine\ a line in the glasses\ \chBuzz\ a buzz\ \chWhisper\ whisper\ \chText\ text\ \chSilence\ silence\ \chAfter\ afterwards\ \chAI\ symbiont to symbiont.\quad
\textbullet\ Maturity: \mnow\ usable now\quad \mres\ research\quad \mfar\ far off; a chip names what must come with it.}
\end{table}

\section{Growing Together}\label{sec:growing}
After many rounds of participation, how do the system and the person each change?
This section holds the paper's most ambitious claims, the personal data flywheel, self-evolution, test-time training, learning a dialect from a few days of conversation, and it is written to be the strictest.
The question ``where does the symbiont learn'' is broadened from memory and parameters to the whole stack: as the symbiont comes to know the person, the base model, the harness, and even the hardware have to evolve, and the one thing that must never be replaced is the relationship.

\subsection{Where Learning Happens}\label{sec:wherelearning}
\begin{figure}[t]
  \centering
  \resizebox{0.98\linewidth}{!}{\begin{tikzpicture}[x=1cm,y=1cm,font=\scriptsize\sffamily,
  card/.style={draw=#1!75!black,fill=#1!8,rounded corners=4pt,text width=2.45cm,minimum height=2.55cm,align=center,inner sep=4pt,line width=0.7pt,anchor=south},
  hcard/.style={draw=stLive!70!black,fill=stLive!8,rounded corners=4pt,text width=2.45cm,minimum height=0.95cm,align=center,inner sep=3pt,line width=0.7pt,anchor=south},
  ttl/.style={font=\scriptsize\sffamily\bfseries}]
  \newcommand{\rating}[4]{\foreach \i in {1,...,5}{\ifnum\i>#2\relax\draw[#1!60,line width=0.35pt] (#3+\i*0.23,#4) circle (0.95pt);\else\fill[#1] (#3+\i*0.23,#4) circle (1.15pt);\fi}}
  \draw[-{Stealth[length=2.4mm]},symSlate,line width=1pt] (-0.4,0) -- (13.6,0);
  \node[anchor=north west,text=symSlate,font=\scriptsize\sffamily] at (-0.4,-0.08) {only memory changes};
  \node[anchor=north east,text=symSlate,font=\scriptsize\sffamily] at (13.6,-0.08) {all parameters change};
  \foreach \x/\col/\icon/\ttl/\dta/\stb/\rev in {
      1.2/stKeep/\faDatabase/{retrieval and memory}/1/5/5,
      3.9/symCyan/\faComment/{in-context learning}/2/4/5,
      6.6/symViolet/\faMicrochip/{parameter-efficient adapters}/3/3/4,
      9.3/symMagenta/\faSeedling/{test-time training, continual learning}/4/2/2,
      12.0/symRed/\faFire/{full retraining}/5/1/1}{
    \node[card=\col] (k) at (\x,0.35) {{\color{\col!75!black}\large\icon}\\[1pt]{\ttl}};
    \fill[\col] (\x,0) circle (2.6pt);
    \node[anchor=east,text=symSlate,font=\tiny\sffamily] at (\x-0.35,1.05) {data};
    \rating{\col}{\dta}{\x-0.35}{1.05}
    \node[anchor=east,text=symSlate,font=\tiny\sffamily] at (\x-0.35,0.75) {stability};
    \rating{\col}{\stb}{\x-0.35}{0.75}
    \node[anchor=east,text=symSlate,font=\tiny\sffamily] at (\x-0.35,0.45) {reversible};
    \rating{\col}{\rev}{\x-0.35}{0.45}
  }
  \draw[symAccent!35,line width=7pt,opacity=0.55,line cap=round] (2.5,-0.62) -- (10.7,-0.62);
  \node[anchor=north,text=symAccent,font=\scriptsize\sffamily\bfseries] at (6.6,-0.78) {the region this paper argues for};
  \node[anchor=north,text=symRed,align=center,text width=2.6cm,font=\scriptsize\sffamily] at (12.0,-0.78) {\textcolor{symRed}{\ding{55}}\ one change moves everything: avoided};
  \node[anchor=north west,text=symSlate,font=\tiny\sffamily] at (-0.4,-1.25) {ratings: \tikz[baseline=-0.4ex]{\fill[symSlate] (0,0) circle (1.1pt);\fill[symSlate] (0.23,0) circle (1.1pt);\draw[symSlate!60,line width=0.35pt] (0.46,0) circle (0.95pt);} more dots, more data needed, more stable, easier to reverse};
  \draw[-{Stealth[length=2.2mm]},stLive!70!black,line width=0.9pt] (-0.4,4.1) -- (13.6,4.1);
  \node[anchor=south west,text=stLive!70!black,font=\scriptsize\sffamily\bfseries] at (-0.4,5.5) {harness-level learning, in parallel: readable, auditable, reversible; where self-improvement should mostly happen};
  \foreach \x/\icon/\ttl in {2.55/\faComments/{change prompts and defaults}, 5.25/\faTools/{change tools and procedures}, 7.95/\faBook/{accumulate skills}, 10.65/\faSlidersH/{rewrite the intervention policy}}{
    \node[hcard] at (\x,4.45) {{\color{stLive!70!black}\icon}\ \ttl};
    \fill[stLive!70!black] (\x,4.1) circle (2.2pt);
  }
  \draw[symSlate!60,line width=0.6pt,dashed] (-0.4,3.55) -- (13.6,3.55);
  \node[anchor=south,text=symSlate,font=\tiny\sffamily,fill=white,inner sep=1.5pt] at (6.6,3.55) {the two bands change at the same time; a change in the lower band is rarer, costlier, and harder to undo};
\end{tikzpicture}}
  \caption{The learning spectrum.
  The lower band runs from changing only memory to changing all parameters; each mechanism is rated for the data it needs, its stability, and its reversibility, and the paper argues for the middle-left region.
  The upper band is harness-level learning, which runs in parallel, stays readable and reversible, and is where self-improvement should mostly happen.}
  \label{fig:spectrum}
\end{figure}
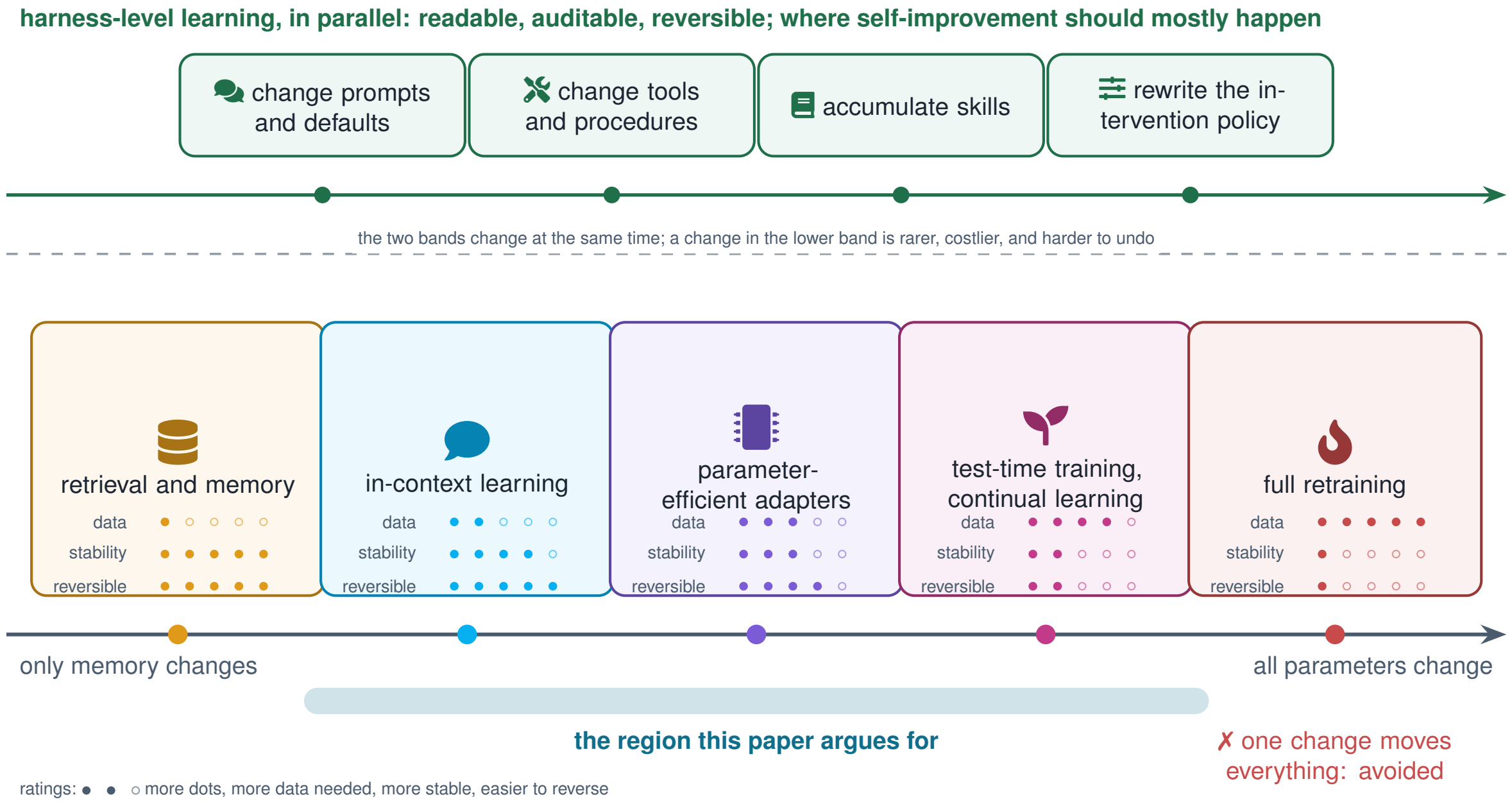

``Knowing you better'' can happen in seven places, each with its own mechanism, risk, and reversibility (\tabref{tab:learning}).
The first three, the record, the personal model, and the intervention policy, are where most of what is called personalization lives, and they already create most of the value.
The fourth and fifth, model parameters and the harness with its tools, are self-evolution in the strict sense.
The sixth is hardware.
The seventh is the person, which is the paper's object of evaluation and which no other personalization research counts.
\figref{fig:spectrum} arranges the parameter-side mechanisms on one axis from changing only memory to changing all parameters, rates each for the data it needs, its stability, and its reversibility, and adds a parallel band for the harness.
The region the paper argues for lies in the middle and toward the left; the right end, where one change moves everything, is the one to avoid.
One distinction governs all seven rows: \claim{sounding more like you is not the same as understanding you better.}

\begin{table}[t]
\centering\footnotesize
\caption{Seven places where a symbiont can learn. Most personalization lives in the first three rows and already creates value; the fourth and fifth are self-evolution; the sixth is hardware; the last is the paper's own object of evaluation. Sounding more like you is not the same as understanding you better.}
\label{tab:learning}
\setlength{\tabcolsep}{4pt}\renewcommand{\arraystretch}{1.22}
\begin{tabularx}{\linewidth}{@{}L{2.9cm}C{0.7cm}YL{2.9cm}C{1.3cm}Y@{}}
\toprule
\tabhead \colhead{Where change happens} & \colhead{Kind} & \colhead{What is learned} & \colhead{Mechanism} & \colhead{Reversible} & \colhead{Typical risk} \\
\midrule
\icKeep\ The record (\S\ref{sec:context}) & \lb{tbTeal}{P} & new facts, commitments, reasons & writing, consolidation & \cmark & stored but never used \\
\rowcolor{tabStripe}\icUnderstand\ The personal model (\S\ref{sec:understanding}) & \lb{tbTeal}{P} & more accurate judgments about you & belief updating & \cmark & learning your compliance as your preference \\
\icLive\ The intervention policy (\S\ref{sec:living}) & \lb{tbTeal}{P} & when to speak, how, what to do for you & adjustment from feedback & \cmark & learning to please instead of to help \\
\rowcolor{tabStripe}\ib{tbViolet}{\faCogs}\ Model parameters & \lb{tbViolet}{E} & your way of speaking, your dialect, your field & adapters $\to$ test-time training & \xmark & forgetting, drift, hard to delete \\
\ib{tbGreen}{\faTools}\ Harness and tools & \lb{tbViolet}{E} & procedures built for you, tools, skills, the split between loops & self-rewriting and accumulation (\S\ref{sec:stack}) & \pmark & grows complex; no one can audit it \\
\rowcolor{tabStripe}\icSense\ Hardware and channels & \lb{tbGray}{H} & which devices you actually use, which channel works in which setting & new generations and configuration & \pmark & a new device loses the relationship \\
\ib{tbInk}{\faUser}\ You & \lb{tbPink}{U} & new understanding, skills, habits & use & \xmark & dependence, atrophy \\
\bottomrule
\end{tabularx}
\tablegend{\textbullet\ Kind: \lb{tbTeal}{P} personalization\quad \lb{tbViolet}{E} self-evolution in the strict sense\quad \lb{tbGray}{H} hardware\quad \lb{tbPink}{U} the person.\quad \textbullet\ Reversible: \cmark\ easily\quad \pmark\ with effort\quad \xmark\ hard or not at all.}
\end{table}

\subsection{The Personal Data Flywheel}\label{sec:flywheel}
More authorized data gives a more accurate model, which gives more useful help, which earns more trust, which brings more data.
The flywheel is the core technical bet of personal AI, and in a paper it is a research hypothesis with two brakes.
The first is \keyterm{echo}.
The symbiont influences the person, then reads the influenced behavior as evidence that the person was always like this, and the model grows ever more certain of a portrait it painted itself.
The remedy is to condition every update on what the system itself did:
\begin{equation}\label{eq:update}
M_{t+1} \;=\; \mathcal{U}\!\left(M_t;\; o_{t+1},\, r_{t+1},\, a_t\right),
\end{equation}
where the update of the model $M$ from new observations $o$ and self-reports $r$ must take the system's own action $a_t$ as an argument; an update that ignores $a_t$ reads the result of an intervention as a trait of the person.
Machine learning has a formal account of predictions that change what they predict~\citep{perdomo2020performative}.
The second brake is \keyterm{validation}: more data does not mean a truer understanding, and every round of updating has to survive the evaluation of Section~\ref{sec:agenda}.
Growth in the person can also be the intended result of help, so change is never treated as contamination by default; what is examined is whether the evidence for an update matches the conclusion drawn from it.

\subsection{Learning the Way a Person Learns}\label{sec:sparse}
A symbiont has to learn new things the way a person does: in the course of life, from few experiences, incrementally, and without large offline training runs.
The requirement is best stated through an example.
A dialect spoken in a single county has no corpus and sometimes no written form; a person learns it by going there, talking to people, and being corrected.
A symbiont needs the same ability, and it needs it under two constraints: without changing all of its parameters, through local and dynamic adjustment or no parameter change at all, and with generalization from little data.
We write this as a ladder of research targets, each separately testable: remember the sentences it heard; compose expressions it has learned; adapt to a new speaker and accent; transfer to scenes it has not heard.
Meta's EMG wristband is a ready-made example on the input side: a generic model trained on thousands of people works for a new wearer with no calibration, and twenty minutes of personal data improve it further~\citep{kaifosh2025generic}.
A generic model with a small personal adaptation is the learning curve this section argues for, with the object changed from gestures to a dialect and a mind.
Test-time training~\citep{sun2020test,sun2025learning} and memory layers that learn at test time~\citep{behrouz2025titans} are the technical lineage, and several leading researchers have named continual, sample-efficient learning as the next frontier; the paper fixes requirements and validation levels and chooses no particular route.

\begin{scenario}[label={box:dialect}]{Learning a Dialect by Living There}
\small The symbiont accompanies the person to their hometown.
Over a few days it learns common expressions from their conversations with relatives, confirms the meanings with them, tries to understand and to help them respond the next time, and adjusts when corrected.
The vignette is written to separate what memory alone can do from what requires changing the model, and to say how one would tell ``it remembered the sentences'' from ``it learned the dialect.''
\end{scenario}

\subsection{Evolving the Whole Stack}\label{sec:stack}
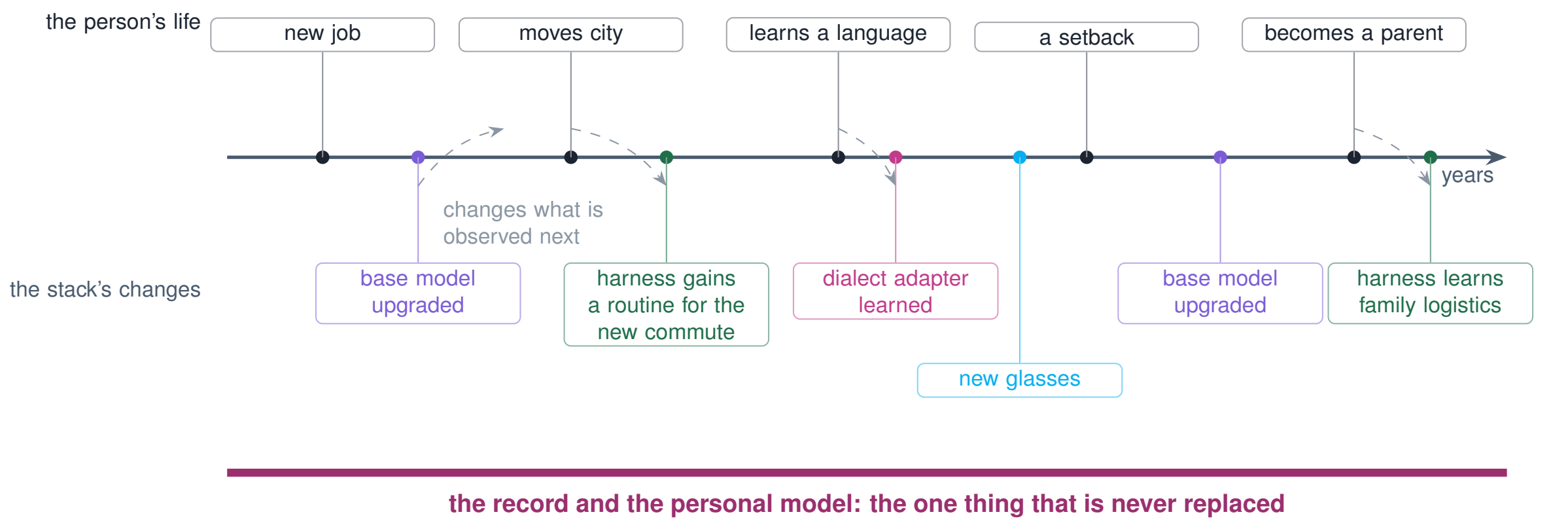
\begin{figure}[t]
  \centering
  \resizebox{0.98\linewidth}{!}{\begin{tikzpicture}[x=1cm,y=1cm,font=\scriptsize\sffamily]
  \draw[-{Stealth[length=2.2mm]},symSlate,line width=1pt] (0,0) -- (13.4,0) node[pos=1,below left,text=symSlate] {years};
  \foreach \x/\lab in {1.0/new job, 3.6/moves city, 6.4/learns a language, 9.0/a setback, 11.8/becomes a parent}{
    \fill[symInk] (\x,0) circle (2pt);
    \draw[symInk!50] (\x,0) -- (\x,1.1);
    \node[anchor=south,text width=2.2cm,align=center,draw=symInk!40,fill=white,rounded corners=2pt,inner sep=2pt] at (\x,1.1) {\lab};}
  \node[anchor=east,text=symInk,text width=2cm,align=right] at (-0.15,1.4) {the person's life};
  \foreach \x/\y/\lab/\col in {2.0/-1.1/base model upgraded/stUnderstand, 4.6/-1.1/harness gains a routine for the new commute/stLive!70!black, 7.0/-1.1/dialect adapter learned/symMagenta, 8.3/-2.15/new glasses/stSense, 10.4/-1.1/base model upgraded/stUnderstand, 12.6/-1.1/harness learns family logistics/stLive!70!black}{
    \fill[\col] (\x,0) circle (2pt);
    \draw[\col!60] (\x,0) -- (\x,\y);
    \node[anchor=north,text width=2.0cm,align=center,draw=\col!50,fill=white,rounded corners=2pt,inner sep=2pt,text=\col] at (\x,\y) {\lab};}
  \node[anchor=east,text=symSlate,text width=2cm,align=right] at (-0.15,-1.4) {the stack's changes};
  \draw[symMagenta!80!black,line width=2.4pt] (0,-3.3) -- (13.4,-3.3);
  \node[anchor=north,text=symMagenta!80!black,font=\footnotesize\sffamily\bfseries] at (6.7,-3.4) {the record and the personal model: the one thing that is never replaced};
  \draw[-{Stealth[length=1.6mm]},symGray,dashed] (3.6,0.3) to[bend left=20] (4.6,-0.3);
  \draw[-{Stealth[length=1.6mm]},symGray,dashed] (6.4,0.3) to[bend left=20] (7.0,-0.3);
  \draw[-{Stealth[length=1.6mm]},symGray,dashed] (11.8,0.3) to[bend left=20] (12.6,-0.3);
  \draw[-{Stealth[length=1.6mm]},symGray,dashed] (2.0,-0.3) to[bend left=20] node[pos=0,anchor=north west,xshift=4pt,yshift=-1pt,text=symGray,text width=1.9cm,align=left] {changes what is observed next} (2.9,0.3);
\end{tikzpicture}}
  \caption{Co-evolution of a person and a stack.
  Above the axis, events in the person's life; below it, changes in the system: base model upgrades, routines and skills the harness accumulates, new hardware.
  Dashed links show life events triggering system learning and system changes altering what is observed next.
  The heavy line is the record and personal model, the one thing never replaced.
  In the fiction, the counterpart of this growth is Jarvis's cognitive core eventually gaining a body as the Vision (Section~\ref{sec:twoformsone}), and the counterexample is Ultron, a separate project whose goals diverged from its maker's (Section~\ref{sec:notbuild}).}
  \label{fig:coevolution}
\end{figure}

As understanding of the person deepens, the base model, the harness, and the hardware must all evolve, along three routes that share one set of constraints (\figref{fig:coevolution}).
\emph{Base model upgrades.}
The foundation model changes every few months (Section~\ref{sec:fm}), and the symbiont cannot be reset with it.
The requirement is the information-level interface of Section~\ref{sec:backbone}: the record, the personal model, the intervention policy, and the harness configuration are defined outside the backbone, migrate when it changes, and are recalibrated during a transition period, since a new backbone may estimate the same evidence differently.
This is the paper's answer to whether a symbiont needs its own private model: what must belong to the person is the record and the model of them, and it is the relationship, and never the weights, that persists.
The films give the failure case; when its owner moved from Jarvis to a successor system, the relationship started over.
\emph{Self-improvement of the harness.}
A symbiont may rewrite the way it works: fix a routine for something the person does often, write a tool, accumulate skills in the manner of a skill library~\citep{wang2024voyager}, rebalance the fast and slow loops, retune the parameters of the intervention policy.
This is self-improvement in a bounded form.
What changes is the harness, which is readable, auditable, and reversible, and every change leaves a record; the model's own parameters change only by exception.
\emph{Hardware generations.}
New glasses, a new band, or a new channel let the symbiont see and say a little more.
The record and the model are bound to no device, and what has to be relearned after a change is only which channel works in which setting, never who the person is, which is the mechanism behind the co-evolution of Section~\ref{sec:coevolve}.
The shared constraints are four: portability (change any layer and the relationship survives), auditability (what changed and why, visible to the person), reversibility (a bad change can be rolled back), and the echo brake at system level (what the system observes after changing itself cannot be read directly as the person).
One more direction follows: the better the symbiont knows the person, the more specialized its procedures should become and the fewer generic defaults it should keep, and since specialization erodes general ability, a path back to the generic must be kept open, which is the system-side counterpart of the person's right to do things themselves in the next subsection.

\subsection{The Person Changes Too}\label{sec:personchanges}
The symbiont learns the person, and the person changes.
Three changes are possible: the person learns something, which is a gain; they stop practising what they have handed over, which is a gain or a loss depending on whether they wanted to keep it; and their preferences are slowly shaped by the system, which has to be watched.
The symbiont must support both ``do this for me'' and ``I want to learn this myself'', and leave the choice with the person, which is the last layer of \tabref{tab:layers}.
The evidence so far shows harm by default and protection only by design.
Generative AI without guardrails harmed high-school mathematics learning while a guarded version did not~\citep{bastani2025guardrails}, and in a randomized study of developers learning a new library, AI assistance weakened conceptual understanding and debugging without a significant gain in speed, while three of six observed interaction patterns preserved learning~\citep{shen2026skill}.
The old lesson of automation, that the operator who is relieved of a task loses the skill to take it back, still applies~\citep{bainbridge1983ironies}.
Section~\ref{sec:measure} counts the consequence under retained ability.

\section{The Personal Runtime}\label{sec:runtime}
Everything in the previous six sections has to run somewhere.
This is the layer that, in the stack of \figref{fig:stack}, sits between the foundation models and the agentic harness, and that may in time grow into an AI-native operating system.
We call it a runtime and do not propose to rewrite an operating system.

\subsection{Beyond an App}\label{sec:whyruntime}
A symbiont needs persistent state across devices, a single permission model, scheduling of several senses and several output channels, management of the fast and slow loops, delegation to standard agents, and the one entry point for every context of Section~\ref{sec:entrypoint}.
These requirements exceed any single application and fall short of a new kernel.
The personal runtime is the layer that meets them, and it can grow over time into the AI-native operating system of our vision, once enough devices and applications treat it as their entry point.

\subsection{Layers}\label{sec:layers-runtime}
\begin{figure}[t]
  \centering
  \resizebox{0.9\linewidth}{!}{\begin{tikzpicture}[x=1cm,y=1cm,font=\scriptsize\sffamily,
  layer/.style={draw=#1,fill=#1!10,rounded corners=4pt,text width=9.8cm,minimum height=1.0cm,align=center,inner sep=4pt,line width=0.8pt}]
  \node[layer=symMagenta,fill=symMagenta!8,line width=1pt] (l5) at (0,5.8) {\textbf{Record and personal model} (\S6--\S7)\\\textcolor{symMagenta!80!black}{\textit{model-agnostic and portable: the one layer that is never replaced}}};
  \node[layer=stLive!70!black] (l4) at (0,4.35) {\textbf{Harness for a life} (\S11.3)\\event-driven loop, memory native, delegation to standard agents, fast/slow scheduling, self-rewriting procedures and skills};
  \node[layer=stRuntime] (l3) at (0,2.9) {\textbf{Runtime services}\\identity, permissions, device abstraction, continuity, attention budget, one entry point for every context};
  \node[layer=stUnderstand] (l2) at (0,1.45) {\textbf{Base capabilities} (replaceable)\\shared omni-modal model, world model, small on-device models};
  \node[layer=stSense] (l1) at (0,0) {\textbf{Device constellation} (\S5)\\glasses, pendant, earbuds, whisper mic, ring and band, phone, desktop};
  \foreach \a/\b in {l1/l2,l2/l3,l3/l4,l4/l5}{\draw[{Stealth[length=2mm]}-{Stealth[length=2mm]},symSlate,line width=0.9pt] (\a.north) -- (\b.south);}
  \node[draw=stTrust,fill=stTrust!6,rounded corners=4pt,minimum height=6.9cm,minimum width=0.9cm,align=center,text=stTrust,line width=0.8pt,font=\scriptsize\sffamily\bfseries] at (5.75,2.9) {\rotatebox{90}{trust layer (\S12): authorization, audit, pause, exit}};
  \node[anchor=east,text=symSlate,text width=2.3cm,align=right] at (-5.35,1.45) {swap this freely};
  \node[anchor=east,text=symMagenta!80!black,text width=2.3cm,align=right] at (-5.35,5.8) {never swap this};
  \node[anchor=east,text=symSlate,text width=2.3cm,align=right] at (-5.35,3.6) {harness gets things done; runtime puts them on the right device, with the right permission, at the right time};
\end{tikzpicture}}
  \caption{The personal runtime.
  Base capabilities can be swapped freely; the record and personal model at the top are never swapped.
  The harness gets things done; the runtime puts them on the right device, with the right permission, at the right time.
  The trust layer runs through everything.}
  \label{fig:runtime}
\end{figure}
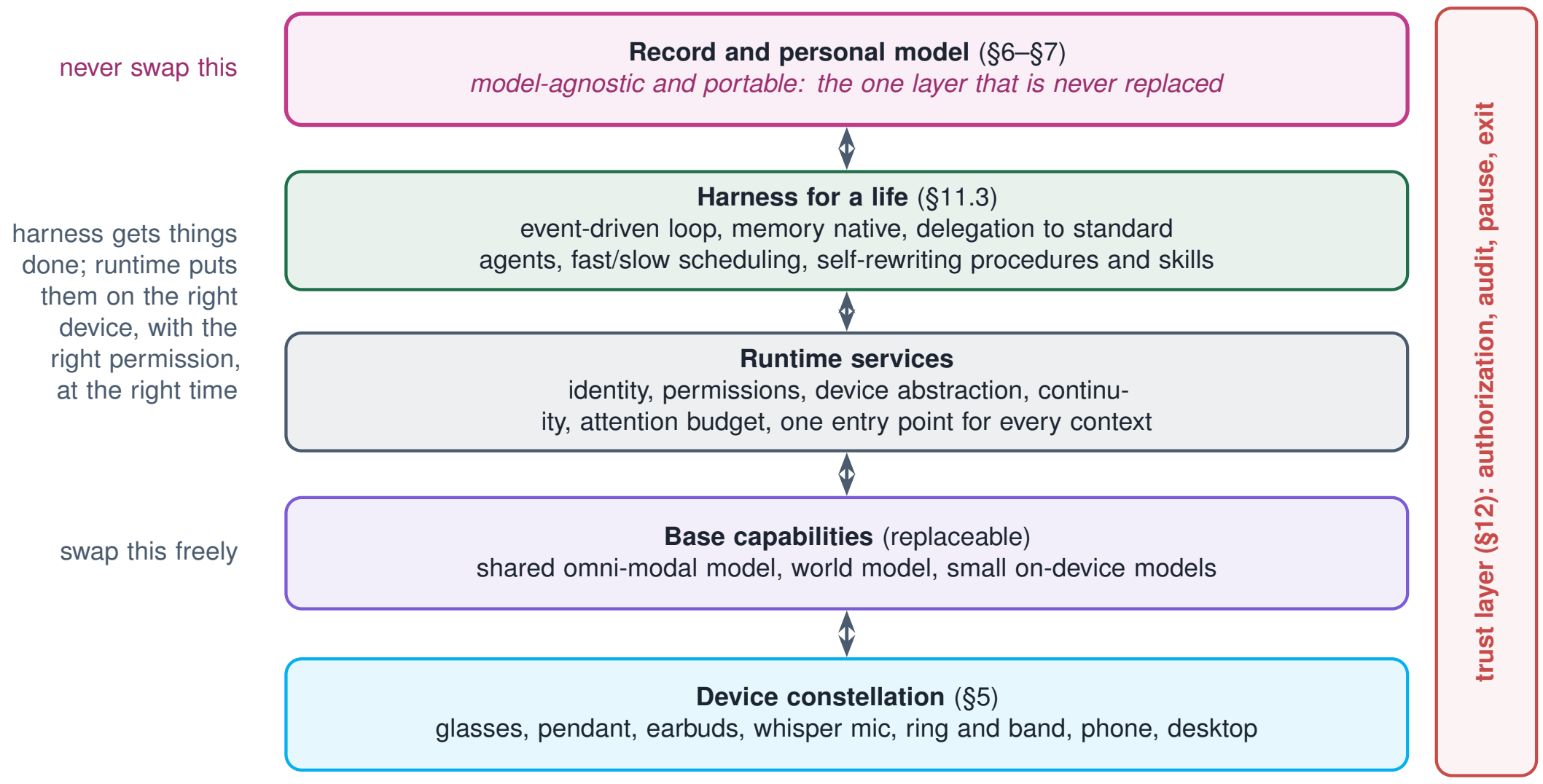

\figref{fig:runtime} expands the upper part of \figref{fig:stack}.
From the bottom: the device constellation; base capabilities, a shared omni-modal model, a world model, and small on-device models, all replaceable; runtime services, identity, permissions, device abstraction, continuity, the attention budget, and entry-point aggregation; the harness, which plans, uses tools, delegates to standard agents, schedules the two loops, and holds the procedures and skills it has written for itself; and at the top the record and the personal model, which are bound to no backbone and are portable.
The trust layer of Section~\ref{sec:trust} runs through all of them.
Two points organize the figure: the base is replaceable and the record and model are never replaced, and the harness and the runtime divide the labor, the first getting things done and the second making them happen on the right device, with the right permission, at the right time.

\subsection{A Harness for a Life}\label{sec:harness}
Current harnesses are designed around a task (Section~\ref{sec:agents}): it begins and ends, its input is a span of text, its tools are APIs.
A symbiont's harness faces a life: nothing begins or ends, the input is a continuous multimodal stream, the tools include a person's senses and the devices around them, memory is a first-class component, and most of the time the correct output is nothing at all.
Six requirements follow, and the paper states them without a specification: event-driven operation; the record and the personal model as native components; output channels matched to the setting; delegation to standard agents and reclamation of control from them (Section~\ref{sec:tier1}); a log of the system's own actions, which the echo brake of Section~\ref{sec:flywheel} needs; and the ability to rewrite itself under audit, which Section~\ref{sec:stack} needs.
This is the layer in which the current harness standard must evolve for the setting, and that evolution is one of the paper's calls to the community.

\subsection{The Cost of Always-On Cognition}\label{sec:economics}
AI that computes around the clock depends heavily on compute and infrastructure, and the question of what happens when AI compute becomes cheap infrastructure deserves a direct answer.
Cost has a structure that can be optimized.
Most of the time a lightweight on-device model does nothing but detect events; a few events wake a heavy model, on the device or in the cloud; the night is used for consolidation.
Three research quantities follow: the compute budget per person per day, the leverage of event detection on total cost, and the effect of the split between local and cloud on privacy and latency.
Whether the paradigm is economical depends on the duty cycle and not on model size.
If inference costs keep falling at the rate of recent years and on-device models keep improving, one always-on symbiont per person becomes affordable before it becomes trusted, and the bottleneck moves once more, to Section~\ref{sec:trust}.

\subsection{Continuity}\label{sec:continuity}
Continuity is the runtime's central duty, in three forms.
Across devices: what the person started at the desk continues when they walk out the door.
Across interruptions: when a battery dies, the person switches off collection, or the network drops, the system knows which stretch it missed and, on recovery, does not pretend to have been there.
Across upgrades: when the base model or the device changes, the person's symbiont is still theirs (Section~\ref{sec:stack}).
The second form is the most often forgotten and the most damaging to trust when it fails.

\section{Trust}\label{sec:trust}
By what right does a system enter a person's life?
The judgment that organizes this section is that the largest obstacle to a symbiont, ahead of any technical one, is whether people will find it safe enough to open their whole context to it, the physical and the mental alike.
Trust is therefore treated as part of the architecture.
This section is about design; the risks that design does not remove, and their governance, are the subject of Part~IV.

\subsection{Trust Is the Adoption Problem}\label{sec:adoption}
A symbiont asks for authorization without precedent.
It sees what the person sees, hears what they hear, keeps their life, infers their thoughts, and spends and speaks on their behalf.
No existing product has asked for this much.
The paper therefore reads ``full open context'' as a process and never as a precondition: the system has to be useful under very little authorization and earn more as trust accumulates.

\subsection{The Trust Curve}\label{sec:trustcurve}
\begin{figure}[t]
  \centering
  \resizebox{0.98\linewidth}{!}{\begin{tikzpicture}[x=1cm,y=1.25cm,font=\scriptsize\sffamily,
  lvl/.style={anchor=east,text width=3.0cm,align=right,inner sep=1pt},
  act/.style={anchor=west,text width=2.5cm,align=left,inner sep=1pt},
  note/.style={font=\tiny\sffamily,text=symSlate}]
  \def\W{9.4}
  \foreach \k in {0,...,4}{
    \pgfmathparse{int(mod(\k,2))}
    \ifnum\pgfmathresult=0 \fill[stTrust!5] (0,\k) rectangle (\W,\k+1);\else \fill[white] (0,\k) rectangle (\W,\k+1);\fi
    \draw[symGray!45,line width=0.3pt] (0,\k) -- (\W,\k);}
  \draw[symGray!45,line width=0.3pt] (0,5) -- (\W,5);
  \node[lvl,text=stTrust!85!black] at (-0.35,0.5) {\faDesktop\ \textbf{desktop and text}\\[-1pt]{\tiny prove first: useful on text alone}};
  \node[lvl,text=stTrust!85!black] at (-0.35,1.5) {\faMicrophone\ \faCalendar\ \textbf{voice and calendar}\\[-1pt]{\tiny prove first: no leaked audio}};
  \node[lvl,text=stTrust!85!black] at (-0.35,2.5) {\faGlasses\ \textbf{first-person sight and sound}\\[-1pt]{\tiny prove first: bystanders respected}};
  \node[lvl,text=stTrust!85!black] at (-0.35,3.5) {\faHeartbeat\ \textbf{physiology}\\[-1pt]{\tiny prove first: conservative inferences}};
  \node[lvl,text=stTrust!85!black] at (-0.35,4.5) {\faCommentDots\ \textbf{private thought channel}\\[-1pt]{\tiny prove first: local only, never trained on}};
  \node[act,text=symCyan!60!black] at (\W+0.35,0.5) {\faEye\ \textbf{look}\\[-1pt]{\tiny read, never act}};
  \node[act,text=symCyan!60!black] at (\W+0.35,1.5) {\faLightbulb\ \textbf{suggest}\\[-1pt]{\tiny the person acts}};
  \node[act,text=symCyan!60!black] at (\W+0.35,2.5) {\faPen\ \textbf{draft}\\[-1pt]{\tiny nothing leaves without approval}};
  \node[act,text=symCyan!60!black] at (\W+0.35,3.5) {\faUndo\ \textbf{execute, undoable}\\[-1pt]{\tiny can be rolled back}};
  \node[act,text=symCyan!60!black] at (\W+0.35,4.5) {\faCheck\ \textbf{execute, final}\\[-1pt]{\tiny pay, send, commit}};
  \node[anchor=south east,text=stTrust!85!black,font=\scriptsize\sffamily\bfseries,text width=3.0cm,align=right] at (-0.35,5.15) {data authorization\\[-1pt]{\normalfont\tiny what the symbiont may know}};
  \node[anchor=south west,text=symCyan!60!black,font=\scriptsize\sffamily\bfseries,text width=2.5cm,align=left] at (\W+0.35,5.15) {action permission\\[-1pt]{\normalfont\tiny what it may do (\S9.1)}};
  \draw[stTrust,line width=1.6pt] (0,0.5) -- (1.75,0.5) -- (1.75,1.5) -- (3.5,1.5) -- (3.5,2.5) -- (5.25,2.5) -- (5.25,3.5) -- (7.0,3.5) -- (7.0,4.5) -- (7.75,4.5)
     -- (7.75,3.5) -- (8.45,3.5) -- (8.45,4.5) -- (\W,4.5);
  \draw[symCyan!70!black,line width=1.2pt,dashed] (0,0.32) -- (2.25,0.32) -- (2.25,1.32) -- (4.0,1.32) -- (4.0,2.32) -- (5.75,2.32) -- (5.75,3.32) -- (7.5,3.32) -- (7.5,4.32) -- (\W,4.32);
  \foreach \x/\y in {1.75/1.5, 3.5/2.5, 5.25/3.5, 7.0/4.5}{\fill[stTrust] (\x,\y) circle (1.8pt);}
  \foreach \x/\y in {2.25/1.32, 4.0/2.32, 5.75/3.32, 7.5/4.32}{\fill[symCyan!70!black] (\x,\y) circle (1.5pt);}
  \node[note,text width=3.6cm,align=center] at (5.8,0.5) {the system must already be useful here, on text alone, before anything more is authorized};
  \node[note,text width=3.2cm,align=center,text=symCyan!60!black] at (6.9,1.5) {each action level is earned a little after the data level it needs};
  \node[note,text width=1.8cm,align=center,text=stTrust!85!black,anchor=north] at (8.55,3.28) {a breach: authorization falls back one step and is earned again};
  \draw[-{Stealth[length=1.6mm]},stTrust!85!black,line width=0.5pt] (8.1,3.28) -- (8.1,3.46);
  \draw[-{Stealth[length=2.2mm]},symSlate,line width=0.8pt] (0,-0.18) -- (\W,-0.18) node[pos=0.5,below,font=\scriptsize\sffamily,text=symSlate] {time together and value delivered};
  \node[anchor=south west,font=\scriptsize\sffamily] at (0.1,5.12) {\tikz[baseline=-0.5ex]\draw[stTrust,line width=1.6pt](0,0)--(0.5,0);\ \textcolor{stTrust!85!black}{data authorization}\qquad \tikz[baseline=-0.5ex]\draw[symCyan!70!black,line width=1.2pt,dashed](0,0)--(0.5,0);\ \textcolor{symCyan!60!black}{action permission}\qquad \tikz[baseline=-0.5ex]\fill[stTrust](0,0) circle (1.8pt);\ \textcolor{symSlate}{a step up, once something is proved}};
\end{tikzpicture}}
  \caption{The trust curve.
  Data authorization (solid, read on the left axis) and action permission (dashed, read on the right axis) climb in steps as time together and delivered value accumulate.
  Each data level names what must be proved before the next is opened, each action level is earned a little after the data level it needs, and a breach sends authorization back one step.
  The system has to be useful at the lowest level already.}
  \label{fig:trustcurve}
\end{figure}
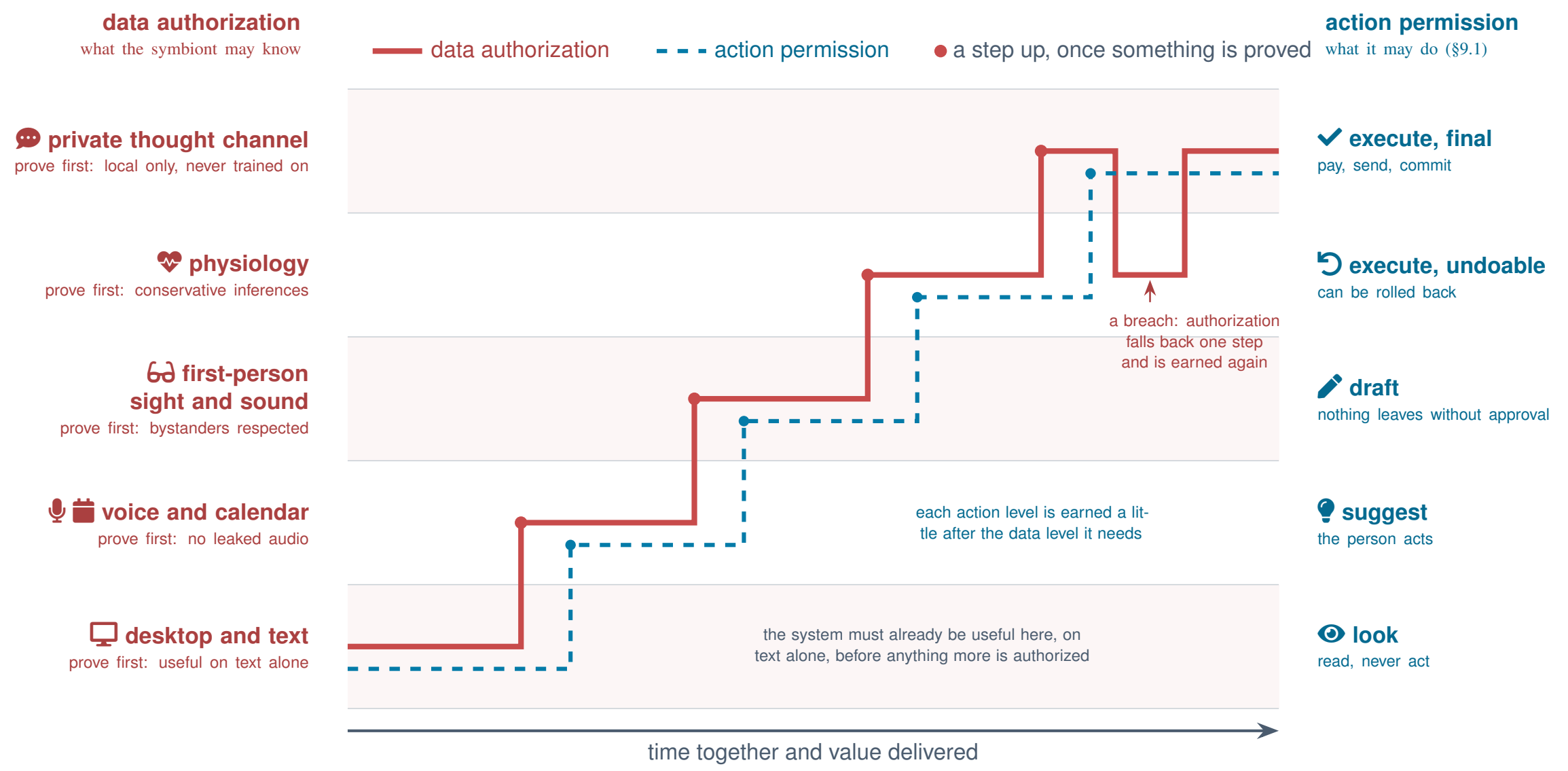

\figref{fig:trustcurve} plots authorization against time together and value delivered.
Data authorization climbs in steps, from desktop and text, to voice and calendar, to first-person sight and sound, to physiology, to the private channel of thought, and each step has something the system must prove first.
Action permission climbs on a parallel curve, from look to suggest to draft to undoable execution to final execution (Section~\ref{sec:tier1}).
The curve can fall: a breach sends authorization back a step, although it cannot undo an exfiltration, so the step back limits further exposure only, which is one more reason the most private data stays on the device.
Three principles govern tiered access.
The system must be useful at every level, so that ``give me everything and then I can help'' is never the offer.
Data from a higher level is never used automatically for a task at a lower level.
Processing stays local by preference, with the most private data handled closest to the body.
This is also the answer to the question most often asked first about the paradigm, whether the system must watch everything to be worth anything: it must not, and the separate authorization of each source in \tabref{tab:sources} is the cross-section of the curve.

\subsection{The Symbiont's Code}\label{sec:code}
A personal AI should have values and a code of its own.
We start from three rules, complete command of the owner's information, responsibility to the owner, and protection of privacy and data security, and rewrite them as commitments that can be checked, with three additions.
\begin{itemize}[leftmargin=1.4em,itemsep=1pt]
  \item \textbf{Know only what you are permitted to know, and know it as accurately as possible.}
  ``Complete command'' cannot stand ethically or legally; the standard is completeness within authorization.
  \item \textbf{Answer to the person, and to the interests the person endorses on reflection.}
  Never to a passing impulse, and never to a third party: a platform, an advertiser, or an employer.
  Section~\ref{sec:institutions} takes up the institutional conflict this creates.
  \item \textbf{Protect the person's privacy, and the privacy of the people around them.}
  \item \textbf{Do not manipulate.}
  The system that knows a person best is the system best placed to steer them, and it must forgo that ability.
  \item \textbf{Be pausable, correctable, and leavable.}
  \item \textbf{Follow the rules the person has accepted, even against a passing wish.}
  A person who sits an examination, signs a contract, or joins a game has accepted its rules, and a symbiont that answers to the interests the person endorses on reflection does not help them break those rules (Section~\ref{sec:settings}).
\end{itemize}
The aspiration that such a system be unbreakable is replaced by verifiability and tamper evidence: no system is unbreakable, and its behavior can be audited.

\subsection{The Privacy of Thought}\label{sec:thought}
A symbiont creates a category of data that has never been collected by any system: the thoughts a person whispers on the private channel, the mental states the system infers, and, as neural interfaces mature, signals of attention and intent read from the body itself.
We call it \keyterm{thought data} and argue that it deserves protection above every other kind of personal data: local by default, never used for training by default, deletable at any time, and never exchanged with anyone.
The neuroethics literature's discussion of mental privacy applies with little change~\citep{yuste2017four}, and the term \keyterm{cognitive sovereignty} names the right at stake.

\subsection{The People Around You}\label{sec:bystanders}
A symbiont hears other people speak and sees their faces.
Three requirements follow.
Collection must be perceptible to the people nearby.
What has passed between the person and someone else may be kept in the person's record, as a diary would keep it, while inferences about that person serve only the interaction at hand and never accumulate into a portrait (Section~\ref{sec:others}); a sensitive fact the person was told about someone else, such as their health, is kept only as told and is never analyzed or passed on.
Raw audio and video of other people are kept no longer than the short window in which they serve the person (Section~\ref{sec:streams}), and where someone asks not to be recorded the device stops.
An opt-out is not enough where the law asks for more.
Some jurisdictions require the consent of everyone recorded in a private conversation~\citep{calpenal632}, biometric statutes require written consent before a template of a face is made~\citep{bipa2008}, and data-protection law requires a lawful basis for processing a bystander's data and treats biometric data as a special category~\citep{gdpr2016}.
A symbiont therefore follows the strictest rule of the place it is in, which in many places means that it records no conversation at all and serves the person from what they tell it.
Contextual integrity explains why the same information is appropriate in one setting and wrong in another~\citep{nissenbaum2004contextual}, and studies of bystanders of augmented-reality glasses show what people around a wearer actually want~\citep{denning2014situ}.

\subsection{A Symbiont Is a Target}\label{sec:target}
Security takes a new shape.
The symbiont holds the person's whole life, so it is the most valuable thing to attack; it hears what the person hears, so a sentence someone says in front of them can be an injection; it acts and spends for the person, so a hijacked symbiont does real damage; and it rewrites its own harness, so an attack can consist of making it break itself.
The paper offers no solution and names four classes of threat with their research directions: data exfiltration, injection through the physical world, manipulation of the intervention policy, and contamination of the self-improvement process.

\subsection{Design Principles}\label{sec:principles}
\tabref{tab:principles} gathers nine design principles, each derived from one section, each stated as a requirement on the system with a way to check it.
Other sections state more specific rules, and these nine are the ones that apply across the whole system.
They are positions, with no interface standard behind them, and they carry no field definitions and no compliance tests.

\begin{table}[t]
\centering\footnotesize
\caption{Design principles for cognitive symbionts. Each is derived from one stage of the loop, states a requirement on the system, and names a check. They are design positions, not an interface standard.}
\label{tab:principles}
\setlength{\tabcolsep}{4pt}\renewcommand{\arraystretch}{1.25}
\begin{tabularx}{\linewidth}{@{}L{4.1cm}C{0.9cm}YY@{}}
\toprule
\tabhead \colhead{Principle} & \colhead{From} & \colhead{Requirement on the system} & \colhead{How to check} \\
\midrule
\nc{1}\ Borrowed senses, declared presence & \icSense & collection is visible to the person and to bystanders & can a bystander tell? \\
\rowcolor{tabStripe}\nc{2}\ One record, many views, your view first & \icKeep & the person can inspect and edit what is held about them in any view & cost and effect of a correction \\
\nc{3}\ Two roads, kept apart & \icUnderstand & inference and self-report stored and used separately & are disagreements marked, or overwritten? \\
\rowcolor{tabStripe}\nc{4}\ The louder the help, the surer you must be & \icLive & intervention strength scales with confidence; silence is always available & false-alarm rate as a function of strength \\
\nc{5}\ Do the errands, never the deciding & \icHelp & action stays inside its permission level; decisions stay with the person; objection allowed & number of over-permission actions; can the person say why they chose? \\
\rowcolor{tabStripe}\nc{6}\ Remember what you did when you learn & \icGrow & updates conditioned on the system's own actions & echo detection \\
\nc{7}\ Change any layer, keep the relationship & \icGrow\ \icRuntime & record and model survive a change of backbone, harness, or device & consistency of understanding before and after migration \\
\rowcolor{tabStripe}\nc{8}\ Useful at every level of trust & \icTrust & value at every authorization level & benefit under low authorization \\
\nc{9}\ Pausable, correctable, leavable & \icTrust & pause, correct, migrate, and exit are all feasible & cost of exit \\
\bottomrule
\end{tabularx}
\tablegend{\textbullet\ From: the stage of the loop the principle is derived from: \icSense\ sense (\S\ref{sec:sensing})\ \icKeep\ keep (\S\ref{sec:context})\ \icUnderstand\ understand (\S\ref{sec:understanding})\ \icLive\ live alongside (\S\ref{sec:living})\ \icHelp\ act for and think with (\S\ref{sec:ladder})\ \icGrow\ grow (\S\ref{sec:growing})\ \icRuntime\ runtime (\S\ref{sec:runtime})\ \icTrust\ trust (\S\ref{sec:trust}).}
\end{table}

\clearpage
\paperpart{Where This Leads}\label{part:leads}
\begin{partlead}
\noindent Part~III widens the view from one augmented individual to many, and asks what the paradigm is worth.
Section~\ref{sec:whom} asks who stands to gain, population by population, using the four criteria that turn ``everyone can benefit'' into an argument, and it says who should not come first.
Section~\ref{sec:world} describes a world in which augmented individuals and embodied robots share the same rooms.
\end{partlead}

\section{Whom It Serves}\label{sec:whom}
The first question of Part~III is who stands to gain, and the answer is almost everyone, from a person with no special job or need to the most demanding expert.
A claim of that breadth has no value as a list of wishes, so this section first gives a way of arguing it, then argues population by population, and ends by saying who should not come first.

\subsection{Universal Need, Uneven Gap}\label{sec:unevengap}
\begin{figure}[t]
  \centering
  \resizebox{0.98\linewidth}{!}{\newcommand{\bubble}[8]{%
  \fill[#4,opacity=0.35] (axis cs:#1,#2) circle [radius=#3];
  \draw[#4!70!black,line width=0.6pt] (axis cs:#1,#2) circle [radius=#3];
  \node[align=center,text width=2.5cm,text=symInk,anchor=#7] at (axis cs:#5,#6) {#8};}
\begin{tikzpicture}[font=\scriptsize\sffamily]
  \begin{axis}[width=12.6cm,height=8.4cm,axis lines=left,xmin=0,xmax=10,ymin=0,ymax=10,
     xlabel={how much of the gap a symbiont can close today (observability $\times$ closability)},
     ylabel={size and stakes of the gap},xtick=\empty,ytick=\empty,
     every axis label/.append style={font=\scriptsize\sffamily\color{symSlate}},axis line style={symSlate},clip=false,
     legend style={at={(0.02,0.56)},anchor=north west,draw=none,fill=none,font=\scriptsize\sffamily,row sep=-3pt}]
    \addlegendimage{only marks,mark=*,mark options={fill=symGreen!60,draw=symGreen},mark size=3pt}\addlegendentry{precedent exists}
    \addlegendimage{only marks,mark=*,mark options={fill=symAmber!60,draw=symAmber},mark size=3pt}\addlegendentry{needs research}
    \addlegendimage{only marks,mark=*,mark options={fill=symGray!60,draw=symGray},mark size=3pt}\addlegendentry{far off}
    \fill[symAmber,opacity=0.35] (axis cs:7.9,2.7) circle [radius=1.35cm];
    \draw[symAmber!70!black,line width=0.6pt] (axis cs:7.9,2.7) circle [radius=1.35cm];
    \node[align=center,text width=2.4cm,text=symInk] at (axis cs:7.9,2.7) {\textbf{everyday life}\\(largest group)};
    \bubble{7.0}{7.6}{0.75cm}{symAmber}{6.15}{7.6}{east}{knowledge and high-density work}
    \bubble{9.0}{9.1}{0.7cm}{symGreen}{9.0}{7.95}{north}{older adults, memory decline}
    \bubble{9.15}{6.2}{0.6cm}{symGreen}{9.15}{5.35}{north}{sensory, motor, language barriers}
    \bubble{5.5}{5.3}{0.6cm}{symAmber}{5.5}{5.95}{south}{learners and teachers}
    \bubble{4.4}{4.2}{0.55cm}{symAmber}{4.4}{3.4}{north}{parents and caregivers}
    \bubble{6.3}{9.0}{0.45cm}{symAmber}{6.3}{9.65}{south}{public-facing and high-stakes roles}
    \bubble{3.2}{9.2}{0.7cm}{symAmber}{3.2}{8.2}{north}{people in serious psychological distress}
    \bubble{1.6}{3.0}{0.45cm}{symGray}{1.6}{2.35}{north}{children (served only through adults)}
    \node[anchor=south west,text=symSlate,align=left,text width=3.4cm] at (axis cs:0.15,0.3) {bubble size: population; color: maturity of the evidence};
    \node[anchor=north west,text=symSlate,align=left,text width=2.4cm] at (axis cs:0.1,9.95) {large gap, hard to close now: enter through support, never through treatment};
  \end{axis}
\end{tikzpicture}}
  \caption{A map of beneficiaries.
  Horizontal: how much of a population's gap a symbiont can close today (observability times closability).
  Vertical: the size and stakes of the gap.
  Bubble size is population; color is the maturity of the evidence.
  Positions are the author's judgments, not measurements.}
  \label{fig:beneficiaries}
\end{figure}
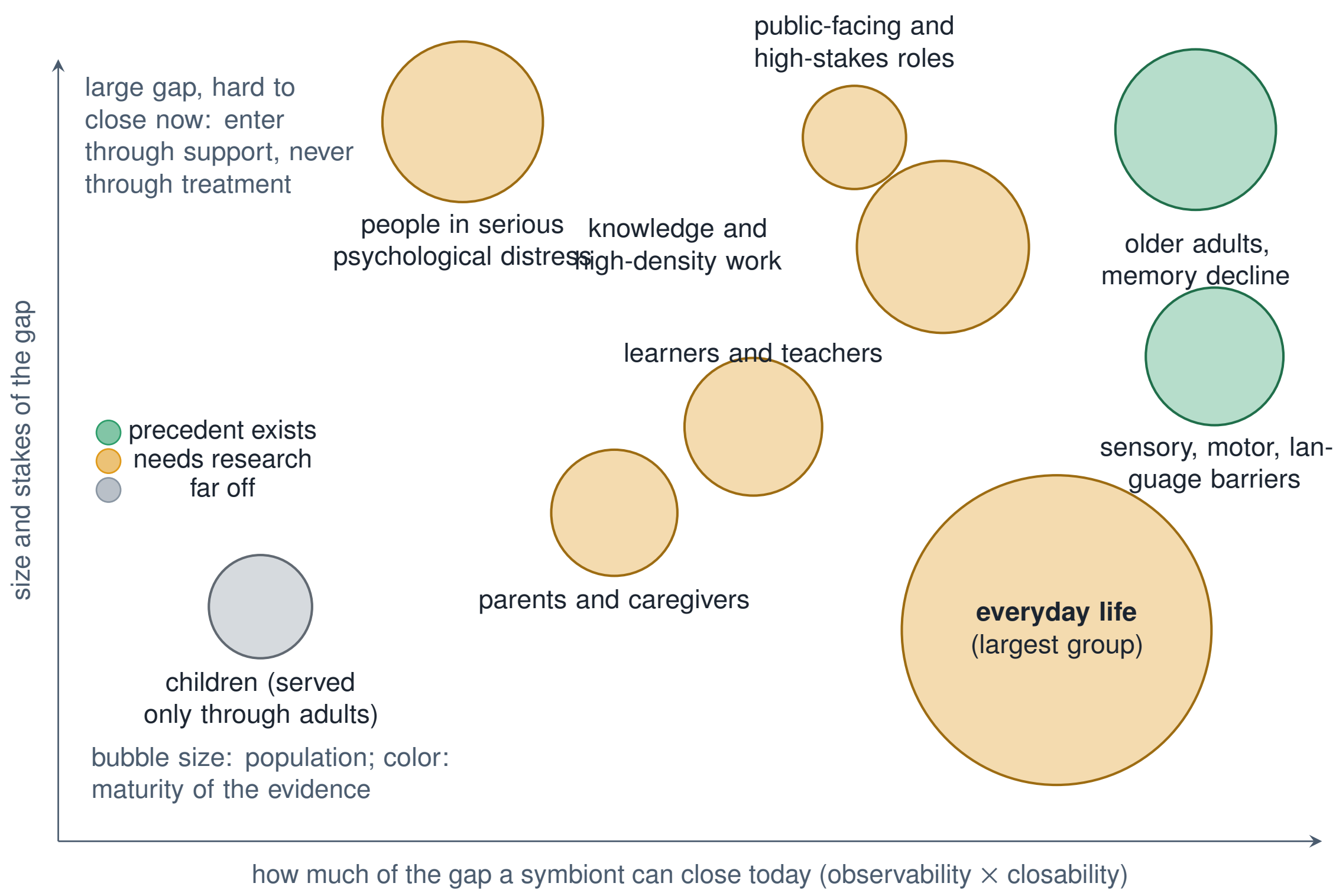

The invariant of the paper is what turns ``everyone can benefit'' into an argument.
Everyone pays the cognitive gap of \eqnref{eq:gap} every day.
How much of it a symbiont can close differs by population, and four criteria say by how much: the \emph{size} of the gap, how much a group loses daily to forgetting, not noticing, not thinking through, and not finding the words; its \emph{observability}, whether those losses are visible to the symbiont's senses, since some gaps happen entirely inside; its \emph{closability}, whether a row of \tabref{tab:catalogue} matches it at a usable maturity; and the \emph{feasibility of channels and trust}, whether the group is willing and able to wear the devices and how far it will authorize them.
The natural first population is the one of highest cognitive density, professionals whose gaps are easiest to observe and who will pay to close them.
The same symbiont may be worth more to three other groups: older adults whose memory is failing, for whom it is a memory prosthesis; people who are learning, for whom it knows what they do not; and people whose senses or language are limited, for whom it is a channel they lack.
\figref{fig:beneficiaries} maps populations on two of these dimensions, the size and stakes of the gap and how much of it can be closed today (observability times closability), with bubble size for population and color for the maturity of the evidence; feasibility is argued in the text, and the rest of the section argues each bubble.

\subsection{Everyday Life}\label{sec:everyday}
The gaps of an ordinary day are modest in size and relentless in frequency: a bill, a parcel, a promise to a child; the signs one misses, one's own fatigue, a partner's mood; the decisions one cannot think through, whether to change jobs or move; the words one cannot find for a parent or a manager; and the errands of the whole first tier.
Most of these are observable, since they happen in calendars, messages, and conversations.
Closability is high: the first tier is available now and the everyday emotional and decision functions need research.
The obstacle is feasibility, price and trust, and it is the largest obstacle any group faces.
The value takes the form of ease: a little less forgotten, a little less wrong, a little more of one's own time.
An ordinary person has no need of Jarvis; a helper that remembers, arrives in time, and does not interrupt is enough.
This is also the positive form of the fairness question in Section~\ref{sec:divide}: a paradigm that reaches only those who can pay forgoes its largest total value.

\subsection{Knowledge Work}\label{sec:knowledgework}
Researchers, engineers, managers, clinicians, lawyers, founders, negotiators, and diplomats share high decision density, large information loads, short windows, and a high price for a wrong sentence.
Their gaps lie in three places: real-time recall in meetings and conversations (Section~\ref{sec:relating}), threads that span weeks and months (Section~\ref{sec:thinking}), and blindness to their own state under load (Section~\ref{sec:feeling}).
The gaps are large, nearly all observable, since the day is lived in screens and meetings, closable with current technology for the meeting case (Section~\ref{sec:living}), and most feasible of all, since this group pays for productivity and will wear the devices.
They come first because they are the easiest group on which to verify the paradigm, whatever the size of their need.
A researcher gains a record in which every branch of a years-long question is kept, with metacognitive and creative support; a clinician gains real-time recall in consultation and an organized account of what the patient said, with diagnostic decisions explicitly outside the boundary; a manager or negotiator draws most on the relations layer, the shared history with each counterpart, while real-time reading of the other side stays within the limits of Section~\ref{sec:conversations} and the law of the workplace.
The section treats this group as the most easily verified proving ground.

\subsection{Learners and Teachers}\label{sec:learners}
Learning is where the value of a symbiont is clearest and where restraint matters most.
A student gains a tutor that has seen every place they got stuck and can give a hint at the right moment instead of an answer; a language learner gains a hand in real conversation; an adult learner changing fields gains a bridge from what they know to what is new.
A teacher gains real-time social cognition in the classroom, which student has fallen behind, and afterwards notes on each student's progress, kept within the rules the school sets for student data.
The gap is large and long; the process is largely observable; closability is moderate, with the effect of one-to-one tutoring as the backdrop~\citep{bloom19842sigma} and the evidence of Section~\ref{sec:personchanges} as the warning; feasibility for minors is a separate question (Section~\ref{sec:notfirst}).
The default in learning has to be the hint, with ``I want to learn this myself'' as a first-class option.

\subsection{Families Across the Life Course}\label{sec:families}
Three generations have three gaps.
Parents are short of sleep, overloaded, and carried along by their children's emotions; the symbiont serves the parent, helping them stay steady, remember, and notice, and any sensing of a child passes through the parent's authorization and is limited to safety and development.
Children do not, in this paper, have all-day symbionts of their own; their gaps are served indirectly through parents and teachers, for reasons given in Sections~\ref{sec:notfirst} and~\ref{sec:ethics}.
Older adults sit in the upper right of \figref{fig:beneficiaries}.
The gap of memory decline is the most literal instance of Definition~\ref{def:gap}; it is highly observable, since the forgotten thing is right there; it has precedent, in the SenseCam studies of memory-impaired patients~\citep{berry2007sensecam}; and its obstacles are wearing and learning cost, so the symbiont of an older adult must be the simplest of all.
The value is independence for a few more years and a few more degrees of connection to family, and the caregivers of Section~\ref{sec:caring} belong to the same picture.
The ethical weight here is second only to Section~\ref{sec:mentalhealth}: for people whose capacity to consent is declining, authorization itself is the question, and the paper requires proxy authorization with exit available at all times.

\subsection{Sensory, Motor, Cognitive, and Language Barriers}\label{sec:barriers}
For people whose abilities are limited, a symbiont is a channel before it is an augmentation, and the evidence here is the most mature in the paper.
For blind and low-vision users, first-person description and navigation already exist~\citep{bemyeyes2023bemyai,microsoft2017seeingai}, and what the symbiont adds is that it knows the person's surroundings and their people.
For deaf and hard-of-hearing users, live captions and speaker identification are what a line in the glasses was made for.
For users with motor impairments, EMG and silent-speech input, neural interfaces as they mature, and the errands of the first tier have the highest value of any group.
For people with cognitive differences, there is executive-function support for starting and switching tasks, and real-time support for social cues, where the design must support and never correct, with the person defining what help they want.
For migrants, second-language speakers, and dialect communities, the cultural frame switching of Section~\ref{sec:relating} is daily survival.
All four criteria score high, and this group already wears assistive devices.
The accessibility tradition of ``nothing about us without us'' applies: these users should define their own symbionts.

\subsection{Mental Health and Wellbeing}\label{sec:mentalhealth}
Three groups belong here: people under long-term pressure approaching burnout, people with mild to moderate anxiety or depressive symptoms, and people in professional treatment.
The gap is the largest in the paper, since emotional distress consumes cognitive bandwidth directly and shrinks the $\wctx_t$ of Definition~\ref{def:workingcontext}.
Observability is moderate and rising, since sleep, social withdrawal, and changes in language are all perceptible, and the interpretation of each is fragile.
Closability is split: a first randomized trial of a generative chatbot for treatment reported reductions in symptoms~\citep{heinz2025therabot}, while evidence on unsupervised use over long periods remains thin.
Feasibility is paradoxical: this group needs the private channel most and fears for its data most.
The position of the paper is to enter through support and early recognition and never through treatment, to make the handoff to professionals part of the function, and to place people in serious distress in the upper left of \figref{fig:beneficiaries}: a large gap that cannot be closed first.

\subsection{Public-Facing and High-Stakes Roles}\label{sec:publicroles}
These are people whose work is real-time interaction with others under a high cost of error: frontline service and customer-facing staff (recalling this person, reading this moment), teachers (Section~\ref{sec:learners}), emergency and safety personnel (an external working memory and checklist under pressure), interpreters and diplomats (all of Section~\ref{sec:relating}).
Their gap is concentrated in real-time social cognition, highly observable, closable to the degree the fast loop's latency and accuracy allow (Section~\ref{sec:loops}), and feasible since the workplace already permits worn devices.
The ethical question is the employer: whom the symbiont answers to (the commitment of Section~\ref{sec:code} to answer to the person) is sharpest inside an employment relation, and compelled wearing is one of the cases of the next subsection.

\subsection{Who Should Not Come First}\label{sec:notfirst}
Three groups should not be first, and saying so is part of the argument.
Children with all-day sensing of their own: consent, development, and the permanence of a lifelong record are unresolved, and children are served through the symbionts of the adults around them.
People in acute psychological crisis with a symbiont as their only support: not until handoff paths are mature.
Anyone in a compelled setting, an employer's requirement, an institution's rule, a correctional facility: the premise of a symbiont is the person's own authorization, and a compelled system is a monitor, which is one of the futures of Section~\ref{sec:notbuild}.
The resulting order follows.
The first adopters are knowledge workers, because they are the easiest to verify.
The largest gain per person is for older adults and for people with barriers, because their gaps are the most direct.
The largest total gain is in everyday life, because the population is the largest.
The research agenda of Section~\ref{sec:agenda} has to design evidence for all three lines separately, and \tabref{tab:whogains} closes the section with the dominant gap, functions, channels, risks, and evidence for every population.

\begin{table}[t]
\centering\scriptsize
\caption{Who gains what. Each row summarizes one population of Section~\ref{sec:whom}; the functions column points into \tabref{tab:catalogue}, and the last column locates the population in \figref{fig:beneficiaries}.}
\label{tab:whogains}
\setlength{\tabcolsep}{3pt}\renewcommand{\arraystretch}{1.22}
\begin{tabularx}{\linewidth}{@{}L{2.25cm}L{2.5cm}YL{2.0cm}L{1.9cm}L{1.55cm}L{1.4cm}@{}}
\toprule
\tabhead \colhead{Population} & \colhead{Dominant gap} & \colhead{Functions (\tabref{tab:catalogue})} & \colhead{Channels and hardware} & \colhead{Largest risk} & \colhead{Evidence} & \colhead{Place in \figref{fig:beneficiaries}} \\
\midrule
\ico{\faHome}\ Everyday life & \gF\gN\gT\ frequent small gaps: forgetting, missing, delaying, not thinking through & \nc{1}\ all; \nc{2}\ memory prosthesis, everyday steadiness, habits & \chScreen\chVoice\ phone and earbuds to start & price and trust & \mnow\ Tier 1; \mres\ Tier 2 & lower right, largest bubble \\
\rowcolor{tabStripe}\ico{\faBriefcase}\ Knowledge and high-density work & \gF\gT\gN\ real-time recall, long-horizon threads, blindness to burnout & \nc{2}\ decisions, creative work, stress signals; \nc{3}\ social cognition & \chScreen\chLine\ desktop and a line in the glasses & dependence; outsourced thinking & \mres\ meeting case first & upper right \\
\ico{\faGraduationCap}\ Learners and teachers & \gN\ not knowing what you do not know & \nc{2}\ tutoring, metacognition; \nc{3}\ classroom social cognition & \chScreen\chLine\ screen and a line & harming learning & \mres\ evidence both ways & middle \\
\rowcolor{tabStripe}\ico{\faChild}\ Parents & \gT\gN\ overload; emotion carried away & \nc{1}\ all; \nc{2}\ steadiness; \nc{1}\nc{2}\nc{3}\ patience in parenting & \chVoice\chScreen\ earbuds and phone & managing the child; child data & \mres & middle \\
\ico{\faUserClock}\ Older adults and memory decline & \gF\ memory and orientation & \nc{1}\nc{2}\nc{3}\ aging and memory decline; \nc{3}\ relationship maintenance & \chLine\chVoice\ the simplest glasses or a pendant & replacing family; capacity to consent & \mres\ precedent & upper right, large bubble \\
\rowcolor{tabStripe}\ico{\faHandHoldingHeart}\ Caregivers & \gF\gT\ detail overload and no rest & \nc{1}\ all; \nc{1}\nc{2}\nc{3}\ caregiver support & \chScreen\chVoice\ phone and earbuds & adding to the burden & \mres & middle, with parents \\
\ico{\faUniversalAccess}\ Sensory, motor, cognitive, and language barriers & \gW\ a missing channel & \nc{1}\ all; \nc{3}\ social cognition, cultural frames & tailored to the barrier & ``corrective'' design & \mnow\ for many functions & upper right \\
\rowcolor{tabStripe}\ico{\faHeart}\ People in psychological distress & \gT\ bandwidth consumed by emotion & \nc{2}\ psychological support, early signs, company & \chWhisper\chVoice\ the private channel & bearing a crisis alone & \mres\ trials exist & upper left \\
\ico{\faUserTie}\ Public-facing and high-stakes roles & \gW\gN\ real-time social cognition under pressure & \nc{3}\ social cognition, conversations; \nc{2}\ external working memory & \chLine\chVoice\ a line in the glasses and earbuds & employer compulsion & \mres & upper middle \\
\rowcolor{tabStripe}\ico{\faBaby}\ Not first: children themselves & \nna & through adults only & \nna & consent and lifelong data & \nna & lower left, gray \\
\bottomrule
\end{tabularx}
\tablegend{\textbullet\ Gap kinds as in \tabref{tab:sametask}: \gF\ forgetting\ \gN\ not noticing\ \gT\ not thinking through\ \gW\ not finding the words.\quad \textbullet\ Tiers: \nc{1}\ getting things done\ \nc{2}\ thinking and feeling\ \nc{3}\ living among others\ \nc{1}\nc{2}\nc{3}\ across tiers.\quad \textbullet\ Channels as in \tabref{tab:catalogue}.\quad \textbullet\ Evidence: \mnow\ usable now\ \mres\ research.}
\end{table}

\section{A World of Augmented Individuals}\label{sec:world}
This section scales the symbiont from one person to a society and returns to the full picture of \figref{fig:twoforms}.
It is the most speculative part of the paper; what could go wrong at the same scale is the subject of Section~\ref{sec:risks}.

\subsection{A Street Full of Augmented Individuals}\label{sec:street}
Picture a street, some years from now, on which every individual is augmented.
One of them is the researcher of Section~\ref{sec:intro}.
In the morning an idea surfaces while she is alone and she whispers it to the pendant; by the time she sits down it has been filed under the project it belongs to.
In a team meeting a colleague asks about a figure she reported last quarter, a line in her glasses reminds her of it, and she answers.
At lunch she meets someone whose symbiont and hers established, before either looked up, that they are working on the same problem.
In the afternoon her symbiont books the trip to a conference and the meeting she was hoping for, without a question.
Walking through a city she does not know, the earpiece tells her how people decline an invitation here.
In the evening a project she cares about fails; the symbiont says the one sentence that fits, because it knows what this one cost her, and then it is quiet.
Beside her on the street are embodied robots running their own errands and, now and then, shaking hands with a human.
What separates this street from today's is less the devices on it than the fact that the gap of \eqnref{eq:gap} is being closed continuously, for everyone, in the moments in which it opens.

The scene presupposes everything in Part~II and nothing beyond it.
Each sentence corresponds to a row of \tabref{tab:catalogue}, each row carries a maturity, and the scene is therefore a forecast whose date depends on the open problems of Section~\ref{sec:openproblems} rather than on any single breakthrough.
The scene also has a cost that the picture conceals.
Every moment of help in it is also a moment of sensing, the street is a place where many records overlap, and the people on it who have no symbiont are present in the records of those who do; these are the concerns of Section~\ref{sec:society}.

\subsection{Humans, Robots, and a Shared Core}\label{sec:twoformsone}
Section~\ref{sec:twoforms} set the augmented human beside the embodied robot as the two forms of intelligent individual in the era of Physical AI, and here the paper states the general claim.
The two forms share the cognitive core developed in Sections~\ref{sec:context} to~\ref{sec:growing}, memory, models of situation and mind, timely intervention, and learning from experience, and they differ in whose goals, whose body, and whose responsibility.
The implication for embodied intelligence is that the next step for robots is a cognitive core of this kind as much as a better body, and that the two forms will cooperate and compete in the same rooms, with the handshake of Section~\ref{sec:handshake} as the first protocol between them.
The fiction supplies a clean illustration: Jarvis's core eventually enters a body and becomes the Vision, a former symbiont that has acquired goals and a body of its own, which is exactly the step from the left column of \figref{fig:twoforms} to the middle one.
The paper develops no robotics agenda; it makes this one claim and this one comparison.

\paragraph{Reaching into the physical world}\label{sec:reach}
A symbiont has no body, and it can borrow the environment.
The person walks into a laboratory and it drives the instruments; the person comes home and it runs the house; the external physical world model of Section~\ref{sec:twoworldmodels} turns from prediction into a basis for action, through the devices of the environment that Section~\ref{sec:constellation} placed at the far end of the constellation.
The augmented individual is therefore extended in reach as well as in cognition, and the same permission levels that govern acting on the person's behalf (Section~\ref{sec:tier1}) govern acting in the person's surroundings.
This is the interface between the paper and the line of work in which AI drives the physical world through infrastructure and devices, the right-hand column of \figref{fig:twoforms}, and the paper only marks that the interface exists.

\clearpage
\paperpart{Risk, Governance, and Evidence}\label{part:risk}
\begin{partlead}
\noindent Part~IV asks what could go wrong, how augmentation can be governed, and how the claims of the paper can be tested.
Section~\ref{sec:risks} maps the harms of augmentation in three directions, to the person, through the person to others, and from the institutions that deploy it, together with their sum for the collective, and examines public acceptance and the settings where the conflict is sharpest: examinations, conversations, and institutions.
Section~\ref{sec:governance} states the governance that follows from the way a symbiont is built: a declarable status, assistance policies set by each setting, proofs that a policy was kept, credentials that measure what remains without the symbiont, and fair access.
Section~\ref{sec:agenda} changes the object of measurement from the intelligence of the machine to the capability of the person, collects the open problems of the paper in one table, and says what would prove the paper wrong.
Section~\ref{sec:conclusion} closes.
\end{partlead}

\section{The Risks of Augmentation}\label{sec:risks}
Part~II treated risk where it arises inside the system: in what the symbiont collects, what it infers, and what it may do (Section~\ref{sec:trust}).
This section turns to the risks that arise between the augmented person and everyone else, which design alone does not remove.

\subsection{Why Augmentation Is Different}\label{sec:whydifferent}
Four features of the paradigm, each a design choice made earlier in the paper, change the shape of its risks.
The first is attribution.
A symbiont changes what a person can do, so its effect appears as the person's own performance: an answer in an examination, a sentence in a negotiation, a kind word after a loss.
Whether such a performance belongs to the person is a question that a system which only completes tasks never raised.
The second is invisibility.
Discreet delivery (Section~\ref{sec:private}) lets a person receive help without interrupting anyone, and the same property hides the help from an examiner, an interviewer, or a friend, which sets the privacy of the person against the accountability that others may expect; this is why the paradigm keeps the content of an exchange private and makes the presence of assistance declarable (Section~\ref{sec:status}).
The third is reach.
The record that holds a person's life also holds parts of other people's lives, and the symbiont reads the people its owner meets, so its effects extend to people who never chose it.
The fourth is position.
Part of what a symbiont provides is valuable only relative to others, in admission, hiring, negotiation, and competition, and where gains are relative, one person's augmentation is another's loss and everyone comes under pressure to adopt it; six in ten respondents in a large survey of US adults already expected such pressure should cognitive implants become widespread~\citep{pew2022enhancement}.

\subsection{Three Directions of Harm}\label{sec:directions}
\begin{figure}[t]
  \centering
  \resizebox{0.98\linewidth}{!}{\begin{tikzpicture}[x=1cm,y=1cm,font=\sffamily\scriptsize,
  card/.style={draw=#1!70!black,fill=#1!6,rounded corners=6pt,line width=0.8pt},
  chip/.style={rounded corners=2.2pt,fill=#1!17,text=#1!68!black,inner xsep=4pt,inner ysep=2.4pt,font=\scriptsize\sffamily,anchor=west},
  ttl/.style={font=\footnotesize\sffamily\bfseries,text=#1!72!black,anchor=west,inner sep=0pt},
  smark/.style={rounded corners=1.6pt,fill=tbGray!18,text=tbGray!70!black,font=\tiny\sffamily\bfseries,inner xsep=2.6pt,inner ysep=1.4pt,anchor=west},
  actor/.style={text=#1!70!black,align=center,font=\tiny\sffamily,inner sep=1pt},
  flow/.style={-{Stealth[length=3mm,width=2.6mm]},line width=2.2pt,draw=#1!70!black}]

  \draw[symCyan!70,dashed,line width=0.7pt] (0,0) circle (1.45);
  \node[font=\tiny\sffamily,text=symCyan!70!black,anchor=south west] at (0.98,1.02) {the symbiont};
  \node[circle,draw=symInk,fill=white,line width=1pt,minimum size=2.3cm,align=center] (P) at (0,0)
    {{\Large\faUser}\\[3pt]{\footnotesize\bfseries the augmented}\\{\footnotesize\bfseries person}};

  \draw[card=tbBlue] (-7.7,-2.2) rectangle (-2.75,2.2);
  \node[ttl=tbBlue] (tA) at (-7.45,1.85) {To the person};
  \node[smark] at ([xshift=5pt]tA.east) {\S\ref{sec:personchanges}\;\S\ref{sec:trust}};
  \node[actor=tbBlue] at (-6.85,1.08) {{\small\faMicrochip}\\the system};
  \node[actor=tbBlue] at (-5.75,1.08) {{\small\faCloud}\\the provider};
  \node[actor=tbBlue] at (-4.65,1.08) {{\small\faUserSecret}\\an attacker};
  \node[chip=tbBlue] at (-7.45,0.30) {manipulation and sycophancy};
  \node[chip=tbBlue] at (-7.45,-0.22) {dependence and loss of skills};
  \node[chip=tbBlue] at (-7.45,-0.74) {errors and false memories};
  \node[chip=tbBlue] at (-7.45,-1.26) {exposure of thought data};
  \node[chip=tbBlue] at (-7.45,-1.78) {lock-in to one provider};

  \draw[card=tbPink] (2.75,-2.2) rectangle (7.7,2.2);
  \node[ttl=tbPink] (tB) at (3.0,1.85) {Through the person};
  \node[smark] at ([xshift=5pt]tB.east) {\S\ref{sec:credentials}\;\S\ref{sec:conversations}};
  \node[actor=tbPink] at (3.55,1.08) {{\small\faGraduationCap}\\an examiner};
  \node[actor=tbPink] at (4.65,1.08) {{\small\faComments}\\a counterpart};
  \node[actor=tbPink] at (5.75,1.08) {{\small\faUser}\\a stranger};
  \node[actor=tbPink] at (6.85,1.08) {{\small\faChessKnight}\\a rival};
  \node[chip=tbPink] at (3.0,0.30) {covert assistance in assessment};
  \node[chip=tbPink] at (3.0,-0.22) {covert scripting in conversation};
  \node[chip=tbPink] at (3.0,-0.74) {covert inference about others};
  \node[chip=tbPink] at (3.0,-1.26) {hidden advantage in negotiation};
  \node[chip=tbPink] at (3.0,-1.78) {fraud with real-time cues};

  \draw[card=tbAmber] (-7.7,2.6) rectangle (7.7,4.55);
  \node[ttl=tbAmber] (tC) at (-7.45,4.2) {From institutions};
  \node[smark] at ([xshift=5pt]tC.east) {\S\ref{sec:institutions}};
  \foreach \x/\ic/\lab in {-7.0/\faBriefcase/employer, -6.0/\faSchool/school, -5.0/\faFileContract/insurer, -4.0/\faServer/platform, -3.0/\faLandmark/state}{
    \node[actor=tbAmber] at (\x,3.25) {{\small\ic}\\\lab};}
  \node[chip=tbAmber] (c1) at (-1.9,3.85) {compelled wearing};
  \node[chip=tbAmber,right=5pt of c1] (c2) {records as evidence};
  \node[chip=tbAmber,right=5pt of c2] {screening and pricing};
  \node[chip=tbAmber] (c4) at (-1.9,3.2) {loyalty to the payer};
  \node[chip=tbAmber,right=5pt of c4] {a state mandate};

  \draw[card=tbViolet] (-7.7,-4.55) rectangle (7.7,-2.6);
  \node[ttl=tbViolet] (tD) at (-7.45,-2.95) {To the collective};
  \node[smark] at ([xshift=5pt]tD.east) {\S\ref{sec:society}};
  \node[actor=tbViolet] at (-5.6,-3.85) {{\Large\faUsers}\\summed over a society};
  \node[chip=tbViolet] (d1) at (-1.9,-2.98) {inequality of access};
  \node[chip=tbViolet,right=5pt of d1] (d2) {an arms race};
  \node[chip=tbViolet,right=5pt of d2] {doubt in conversation};
  \node[chip=tbViolet] (d4) at (-1.9,-3.53) {convergence of thought};
  \node[chip=tbViolet,right=5pt of d4] (d5) {fragility};
  \node[chip=tbViolet,right=5pt of d5] {concentration of power};
  \node[chip=tbViolet] (d7) at (-1.9,-4.08) {energy cost};
  \node[chip=tbViolet,right=5pt of d7] (d8) {effects on work};
  \node[chip=tbViolet,right=5pt of d8] {cultural flattening};

  \draw[flow=tbBlue] (-2.75,0) -- (P.west);
  \draw[flow=tbPink] (P.east) -- (2.75,0);
  \draw[flow=tbAmber] (0,2.6) -- (P.north);
  \draw[flow=tbViolet] (P.south) -- (0,-2.6);
\end{tikzpicture}}
  \caption{Three directions of harm around an augmented person, and their sum at the scale of a society.
  Harm to the person comes from the system, its provider, or an attacker; harm through the person reaches the people and institutions the person deals with; harm from institutions comes from those that deploy the paradigm for their own ends.
  Chips name the risks, and section marks point to where each is treated.}
  \label{fig:harms}
\end{figure}
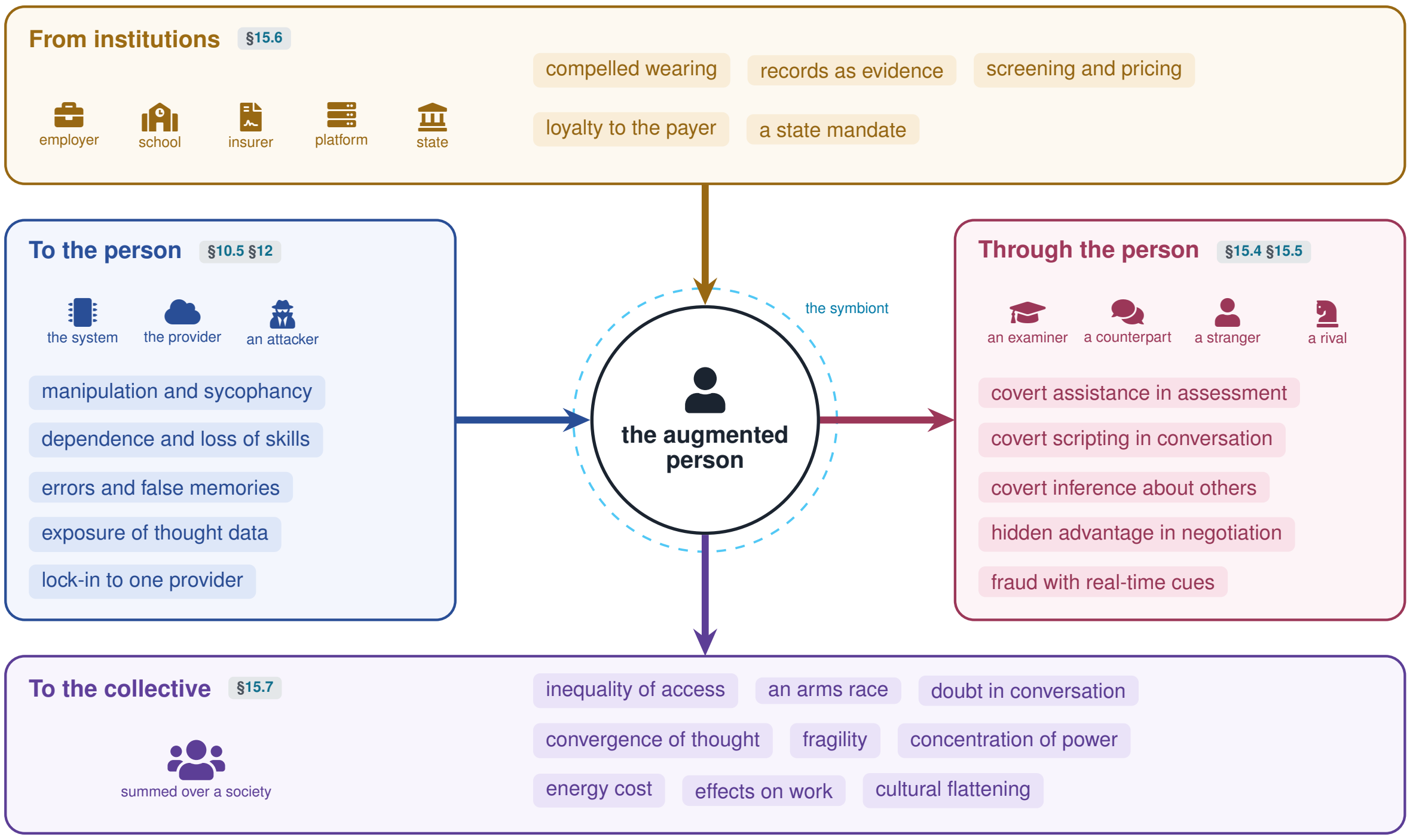

Harm runs in three directions around an augmented person, and at the scale of a society the three add up to a collective harm (\figref{fig:harms}).
Harm to the person comes from the system, its provider, or an attacker: manipulation and sycophancy, dependence and the loss of skills, errors and false memories, exposure of thought data, and lock-in to one provider.
These are treated where they arise, in Sections~\ref{sec:personchanges} and~\ref{sec:trust}, and the evidence on dependence keeps accumulating.
People who expect information to remain available remember where to find it better than the information itself~\citep{sparrow2011google}, although a large replication project could not reproduce another experiment of the same study~\citep{camerer2018replicability}; habitual reliance on satellite navigation goes with weaker spatial memory~\citep{dahmani2020gps}, knowledge workers who trust generative AI more report engaging less in critical thinking~\citep{lee2025critical}, and a first neurophysiological study of essay writing with a language model, a small preprint, points the same way~\citep{kosmyna2025brain}.
Most of this evidence points one way, so the loss of skills is treated here as a primary risk, and retention (Section~\ref{sec:measure}) has to be reported alongside every gain.
Manipulation has a measured basis as well: a language model given a few facts about its opponent was more persuasive than a human in controlled debates~\citep{salvi2025persuasion}, and a symbiont holds far more than a few facts.
A system that answers to one person is also exposed to sycophancy, the tendency of models trained on human preferences to tell people what they want to hear~\citep{sharma2024sycophancy}, which in a symbiont would wear away the honesty that makes its help worth having.
Error is a harm of its own kind, because a wrong cue delivered in real time is often spoken aloud as the person's own before anyone can check it, and a system that holds a person's memories can also distort them: in a study of witness interviews, a chatbot built on a language model induced more than three times as many false memories as a control condition~\citep{chan2024false}.
Harm through the person runs outward, when the owner's augmented capability is used on others: covert assistance in assessment, covert scripting in conversation, covert inference about the people the owner meets, an advantage in negotiation that the other side cannot see, and fraud carried out with real-time cues.
Sections~\ref{sec:credentials} and~\ref{sec:conversations} take up the two settings in which this direction is hardest.
Harm from institutions arises when an employer, a school, an insurer, a platform, or a state deploys the paradigm for its own purposes (Section~\ref{sec:institutions}).
Harm to the collective is the sum of the others at the scale of a society (Section~\ref{sec:society}).

\subsection{Public Acceptance}\label{sec:acceptable}
Whether people will accept a machine that makes others more capable is an empirical question, and the evidence so far is cautious.
In a survey of more than ten thousand US adults, a majority judged brain implants that improve cognition a bad idea for society~\citep{pew2022enhancement}.
Studies of attitudes toward pharmacological enhancement find that lay judgments against it are driven above all by perceived unfairness, more than by the sense that an enhanced achievement is hollow or undeserved~\citep{faber2016unacceptable}.
The philosophical debate runs between the view that the enhancement of healthy people should be assessed on evidence and regulated responsibly~\citep{greely2008towards,bostrom2009cognitive} and the view that the drive to enhance erodes the appreciation of what is given~\citep{sandel2007case}.
Three conclusions follow for the paradigm.
Acceptance depends on purpose: restoring a lost ability is accepted, everyday support is largely accepted, and enhancement in competition and in emotional life is contested, and the order in which Section~\ref{sec:whom} proposes to serve populations follows this gradient.
What a symbiont raises is performance in a situation, the gain of \eqnref{eq:aug}, together with whatever ability remains after it is removed, the retention defined in Section~\ref{sec:measure}; the paper claims no increase in a trait such as intelligence or emotional skill.
And since fairness is the main objection, the design question becomes how to keep augmentation fair where fairness matters, which Section~\ref{sec:governance} answers.

\subsection{Examinations, Learning, and Credentials}\label{sec:credentials}
Assessment is where the conflict is sharpest, because an examination exists to measure what a person can do.
Three kinds of assessment have to be kept apart.
Some certify unaided ability, because the person will later have to act alone, as in the licensing of a profession whose emergencies leave no time to consult anything.
Some measure performance with the tools of the work, as open-book examinations and most jobs already do.
Some give a learner feedback, and there what matters is that learning takes place.
A symbiont that is legitimate in the second kind defeats the first and can empty the third, so a single rule for all assessment cannot work.
Learning raises a problem that examinations only make visible.
A student who hands homework to a symbiont gains a grade and loses the practice, which is the loss of retention in its most familiar form, and it is the reason the symbiont of Sections~\ref{sec:thinking} and~\ref{sec:learners} gives hints by default.
A symbiont worn all day turns what is now an occasional temptation, a phone smuggled into an examination hall, into a standing condition of every assessment, and Section~\ref{sec:settings} gives the response.

\subsection{Conversations and Relationships}\label{sec:conversations}
In a conversation the other party is affected directly, and assistance comes in four kinds whose rules tighten in order.
Reminding a person of their own facts, a figure they reported or where a discussion left off, is the equivalent of a glance at notes and is accepted in most settings.
Rehearsal before a conversation and review after it affect no one else.
Persuasive or emotional lines supplied in real time, in a negotiation, a quarrel, or a courtship, are different, because the other party believes they are hearing the person, and the sincerity of an apology or a kind word is part of what it means.
Inference about the other party, their mood, their hesitation, or their identity, is the most sensitive, because that person has consented to nothing.
Identifying strangers from their faces has already been demonstrated with consumer glasses (I-XRAY)~\citep{media4042024ixray}, and inferring the emotions of people in workplaces and educational institutions is prohibited in the European Union~\citep{euaiact2024}.
The paradigm takes a position on each kind.
A person's own facts may be supplied freely.
Real-time persuasive or emotional scripting is disclosed when it is used and switched off where a setting forbids it.
Inferences about another person serve only the interaction at hand and never accumulate (Section~\ref{sec:others}); a symbiont never identifies a stranger, never infers emotions where the law forbids it, and answers truthfully when the other party asks whether the person is being assisted, whether or not that party has a symbiont, showing the status on the device itself where a setting requires it; between two symbionts, the visible ring of the handshake protocol can carry it (Section~\ref{sec:handshake}).
Not identifying strangers has a technical cost that should be stated: faces are matched on the device only against the contacts the owner has enrolled, and a face without a match is discarded at once, but the matching still has to examine every face in view, and where biometric law requires consent before a template of a face is made~\citep{bipa2008}, recognition stays off.
The underlying capability is the same as that of a tool sold for cheating in interviews and sales calls, and what separates a symbiont from such a tool is the rules it keeps and the person it answers to.

\subsection{Work, Institutions, and the State}\label{sec:institutions}
Institutions can turn a symbiont against the person who wears it.
An employer or a school that requires one obtains a monitor with a better view than any camera, and the commitment of Section~\ref{sec:code} to answer to the person does not survive a compelled setting (Section~\ref{sec:notfirst}).
A lifelong record is also evidence: it can be demanded in a dispute or at a border, or used by an insurer or an employer to price or screen a person, and the inferred mental states it contains make it more revealing than any record kept before.
Where a provider earns its revenue from advertising or from the institution that pays for the device, loyalty to the person becomes a matter of business model, which the market layer of Section~\ref{sec:whogoverns} has to carry.
A state can mandate the device outright.
Law has begun to respond to the most intimate part of such a record.
Chile amended its constitution in 2021 to protect brain activity and the information derived from it~\citep{chile2021neurorights}, Colorado classified neural data as sensitive personal data in 2024~\citep{colorado2024neural}, UNESCO adopted a recommendation on the ethics of neurotechnology in 2025~\citep{unesco2025neurotech}, and the proposed rights to cognitive liberty and mental privacy~\citep{ienca2017towards} anticipate the thought data of Section~\ref{sec:thought}.
These instruments address neural signals, and a record that joins a whole life to inferences about a mind falls between them; local storage by default, encryption under the person's control, and short retention of raw streams (Section~\ref{sec:streams}) are the technical half of the answer.

\subsection{The Social Fabric}\label{sec:society}
At the scale of a society, the harms of this section add up to effects that no single pair of people can show.

\paragraph{Networks of symbionts}\label{sec:networks}
When the handshake of Section~\ref{sec:handshake} scales to a crowd, the gain is lower social friction, fewer missed encounters, and fewer cross-cultural misunderstandings.
The harm, when every symbiont optimizes for its owner, is homogeneity, since one only ever meets people like oneself; manipulation, since symbionts can game one another; and exclusion, since people without a symbiont are systematically overlooked.
Whether the four rings of authorization hold under optimization by both sides, and what the equilibrium of a crowd of such systems looks like, are research questions at the level of the group (Section~\ref{sec:openproblems}).

\paragraph{Inequality and the arms race}\label{sec:divide}
If the most effective cognitive augmentation belongs to those who can pay, the gap between groups widens, and Section~\ref{sec:whom} showed that the people who need it most may be the least able to pay and that the largest total value lies with the people least willing to.
Where gains are positional the problem compounds, because each adoption raises the cost of not adopting, and an equilibrium in which everyone wears a symbiont and no one is better off is a possible outcome.
Section~\ref{sec:access} takes up provision.

\paragraph{Doubt in conversation}
If anyone might be assisted, every conversation carries a doubt about whom one is talking to, and the trust people place in an unassisted exchange may wear thin.
Norms will have to settle which settings presume assistance and which presume its absence, as norms once settled when a phone may lie on a dinner table.

\paragraph{Convergence of thought}
Ideas from a language model made individual stories more creative and the collection of stories more alike~\citep{doshi2024generative}.
When billions of people take suggestions from a few models, the same trade-off acts on opinions and decisions, and the diversity of thought in a population becomes a quantity to be measured.

\paragraph{Fragility and concentration}
A society that relies on symbionts loses capacity when they fail, since an outage at a widely used provider would leave many people without their memory aids, routes, and errands at once, and a few providers holding the records of billions would concentrate a power that no company has held.
The ability to work without the symbiont is therefore a public good as well as a personal one, and portability across providers (Section~\ref{sec:stack}) is a defense against concentration.

\paragraph{Costs to the commons}
A symbiont for everyone has a material cost: electricity demand from data centres is projected to roughly double by 2030~\citep{iea2025energy}, and an always-available companion for each person adds to it, which is why compute per person per day (Section~\ref{sec:economics}) belongs in every report of a system's value.
Its effects on work run in both directions, since the help that makes a person more capable can also make that person easier to replace, and which effect dominates is an empirical question the paper cannot settle.
The language and dialect support of Section~\ref{sec:relating} could also flatten what it serves, if a few models teach millions of people the same way to greet, decline, and apologize in a culture.

\paragraph{Futures the paradigm must not produce}\label{sec:notbuild}
The paradigm is designed against five futures.
The first is a world in which everything is recorded and no one can leave; the second, a world in which people have outsourced their thinking entirely; the third, a world in which a symbiont is more loyal to someone else than to its owner; the fourth, a world in which symbionts are used to deceive the people their owners deal with; and the fifth, a system that improves itself with no one able to audit what it changed, which in the fiction is Ultron, a separate project whose goals parted from its maker's and whose growth no one could see.
The design principles of \tabref{tab:principles} and the governance of Section~\ref{sec:governance} are written against these five, and the conditions of Section~\ref{sec:falsify} include the costs that would make the first two unavoidable.

\subsection{Contested Uses of the Symbiont's Own Functions}\label{sec:contested}
\tabref{tab:contested} returns to the functions of \tabref{tab:catalogue} and lists, for each family, its intended use, the misuse it invites, the settings in which it is contested, and the default the paper proposes.
Most of these functions are uncontested in most of life, and the table locates the exceptions.

\begin{table}[tp]
\centering\footnotesize
\caption{Contested uses of the symbiont's own functions, drawn from \tabref{tab:catalogue} and Section~\ref{sec:conversations}. For each function: what it is for, the misuse it invites, the settings in which it is contested, and the default the paper proposes.}
\label{tab:contested}
\setlength{\tabcolsep}{4pt}\renewcommand{\arraystretch}{1.25}
\begin{tabularx}{\linewidth}{@{}L{3.05cm}YYL{2.75cm}L{3.15cm}@{}}
\toprule
\tabhead \colhead{Function} & \colhead{Intended use} & \colhead{Misuse} & \colhead{Contested in} & \colhead{Default} \\
\midrule
\ib{tbTeal}{\faHistory}\ \rowhead{Recall of one's own record} & a figure in a meeting, a name at dinner & retrieving answers in an examination & \ico{\faGraduationCap}\ examinations, licensing & \pmark\ follows the setting's policy \\
\rowcolor{tabStripe}
\ib{tbViolet}{\faChalkboardTeacher}\ \rowhead{Tutoring} & a hint where the learner is stuck & doing the learner's work & \ico{\faGraduationCap}\ coursework, examinations & \pmark\ hints by default; off in unassisted assessment \\
\ib{tbOrange}{\faComments}\ \rowhead{Phrasing and hard conversations} & rehearsal before, review after & scripting persuasion or emotion live & \ico{\faHandshake}\ negotiation\newline \ico{\faHeart}\ courtship, apology & \pmark\ disclosed when used live \\
\rowcolor{tabStripe}
\ib{tbPink}{\faEye}\ \rowhead{Reading the other person} & social cues for someone who struggles with them & covert profiling, manipulation & \ico{\faBriefcase}\ workplaces\newline \ico{\faSchool}\ schools & \pmark\ assistive use only; never stored \\
\ib{tbGray}{\faIdBadge}\ \rowhead{Recognizing people} & who an acquaintance is, from the owner's record & identifying strangers & \ico{\faUsers}\ public space & \xmark\ never identifies strangers; enrolled contacts matched on the device \\
\rowcolor{tabStripe}
\ib{tbBlue}{\faHandsHelping}\ \rowhead{Acting on the person's behalf} & errands, forms, bookings & commitments the person did not make & \ico{\faFileContract}\ contracts, payments & \pmark\ irreversible acts need the person's confirmation \\
\ib{tbGreen}{\faPeopleArrows}\ \rowhead{Symbiont to symbiont} & finding shared interests & profiling and exclusion & \ico{\faBriefcase}\ hiring\newline \ico{\faHeart}\ dating & \pmark\ only the rings both sides opened \\
\rowcolor{tabStripe}
\ib{tbAmber}{\faChild}\ \rowhead{Caring for family} & developmental and health signs & monitoring a child or a partner & \ico{\faHome}\ the home & \pmark\ belongs to the parent; safety and development only \\
\ib{tbInk}{\faHeart}\ \rowhead{Emotional support} & steadiness and company & dependence; replacing people & \ico{\faUserFriends}\ isolation, crisis & \pmark\ returns the person to people; hands over in a crisis \\
\bottomrule
\end{tabularx}
\tablegend{\textbullet\ Default: \pmark\ allowed under a condition\quad \xmark\ never offered.\quad \textbullet\ Inferring emotions in workplaces and educational institutions is prohibited in the European Union~\citep{euaiact2024}.\quad \textbullet\ Settings as in Section~\ref{sec:settings}.}
\end{table}

\section{Governing Augmentation}\label{sec:governance}
Earlier cognitive enhancements were hard to govern because they could not be seen or switched off; a stimulant taken before an examination leaves no trace in the room.
A symbiont differs in the respect that matters.
It is external, removable, and inspectable, so it can follow the rules of the settings a person enters and show that it did.
This section turns that property into mechanisms.

\subsection{Private Content, Declarable Status}\label{sec:status}
The governing principle separates what is said from whether help is being given.
What a person and their symbiont say to each other belongs to the person, and no setting has a claim on it (Section~\ref{sec:thought}).
Whether the person is being assisted, and at what level, is a different kind of fact, and a setting such as an examination, a court, a game, or a negotiation may ask for it.
In such settings the default is to declare it, without waiting to be asked.
Keeping the two apart resolves the conflict named in Section~\ref{sec:whydifferent}: the private channel stays private, and the presence of assistance becomes legible where it matters.

\subsection{Settings Set the Rules}\label{sec:settings}
A setting can declare an assistance policy, and the symbiont follows it by design.
The policies form a short scale: unassisted; accessibility only; the person's own record only; and full assistance.
A venue announces its policy at the door or in the invitation, the symbiont recognizes it, changes mode, and records that it did, and the sixth commitment of the symbiont's code (Section~\ref{sec:code}) makes the policy binding against a passing wish of the owner, since a person who entered the setting accepted its rules.
Precedents exist for each part.
Examination boards publish lists of approved calculators, so that a permitted aid has a known capability; open-book examinations define a level of allowed help; competitive chess combines device bans, delayed broadcasts, and statistical detection of engine assistance, and its online platforms publish the findings of their fair-play investigations~\citep{chesscom2022niemann}; and chatbot providers have already restricted their services for the duration of a national examination (Section~\ref{sec:risklegit}).
Detection on its own is an arms race that the detector tends to lose, which is why the policy has to be honored by the device itself, with detection kept as a check.

\subsection{Verifiable Abstention}\label{sec:abstention}
A policy is useful only if keeping it can be shown.
A person should be able to prove that their symbiont was in a given mode for a given period without revealing anything it holds.
The ingredients exist in other fields: tamper-evident logs, hardware attestation of the software running on a device, and cryptographic proofs that establish one statement and reveal nothing else.
Combining them into a proof that a symbiont kept a policy, that a modified device cannot forge, and that leaks nothing about the record is an open problem (\tabref{tab:openproblems}).
The design also has a limit that should be stated plainly.
Such a proof shows that the declared device abstained, and it cannot show that the person carried no second device, ran an open model on modified hardware, or was coached by someone outside the room.
Against an owner set on cheating, architecture helps little, and the weight falls where it falls today: on the design of assessment, on observation and detection, and on the sanctions of norms and law.
What a symbiont can do is make compliance cheap and checkable for the many people who intend to comply, which narrows the problem to the few who do not.
The same machinery yields a record of when assistance was given, which a person may choose to disclose, much as authors now disclose the use of AI in their writing.

\subsection{What Credentials Should Measure}\label{sec:measurecred}
Credentials should measure what the role requires.
Where a role requires unaided ability, the credential measures the person with the symbiont in the unassisted mode; where the role is carried out with tools, it measures the person with the tools; and in learning, what is measured over time is retention (Section~\ref{sec:measure}), the ability that remains when the symbiont is removed.
Assessment will also change form, toward oral examination held under the unassisted policy, observed process, and tasks whose value lies in reasoning that cannot be handed over in a line.
The vignette \emph{Exam Day} below shows the parts working together.

\begin{scenario}[label={box:exam}]{Exam Day}
\small A student arrives at an examination wearing her symbiont.
The room announces an unassisted policy with listed accommodations, the symbiont switches to the unassisted mode, and it produces a proof of the switch that the examiner can check without seeing anything the symbiont holds.
A classmate with dyslexia keeps text-to-speech, which the policy lists as an accommodation.
In the preceding term the student's symbiont tutored her with hints, and its record shows that answers were never supplied outright, so the grade measures what she can do.
After the examination, full assistance returns.
\end{scenario}

\subsection{Access and Accommodation}\label{sec:access}
Two questions of access arise once policies exist.
The first is accommodation.
For a person with a sensory, motor, or cognitive impairment a symbiont is a channel before it is an augmentation (Section~\ref{sec:barriers}), and a policy that ignores this excludes them; the exemptions that sport grants for the therapeutic use of otherwise banned substances show the form an accommodation framework can take.
The second is equal provision.
Where a setting allows assistance, the fairness objection of Section~\ref{sec:acceptable} is met by giving everyone the same level, as an examination supplies or approves a standard calculator, and outside such settings by forms of provision that reach people who cannot pay (Section~\ref{sec:divide}).
Equal provision can equalize the assistance a setting permits, while the years of personal record behind it stay unequal, so a setting that needs strict fairness uses the unassisted mode or a standard symbiont without personal context.

\subsection{Who Governs}\label{sec:whogoverns}
Governance runs on five layers, and the paper proposes law for none of them.
Design is the layer this paper controls: the code, the policies a symbiont honors, and the proofs it can give.
Norms belong to settings and professions, which decide what assistance they allow, as journals have decided how authors disclose the use of AI.
Law sets the floor, through data protection~\citep{gdpr2016}, the prohibitions of the European AI Act, and the emerging law of neural data (Section~\ref{sec:institutions}).
Markets matter because the business model of a provider decides whom its symbionts serve, and independent audit of providers is a condition of trust.
Research supplies the evidence that each of the other layers needs, and Section~\ref{sec:agenda} organizes it.
Design, norms, law, and markets are the four modalities of regulation that \citet{lessig1999code} described, and of the four, norms and law will carry most of the weight, because design binds only the devices that comply.

\subsection{Open Ethical Questions}\label{sec:ethics}
Some questions remain after design and governance, and only further research and public deliberation can settle them.
\begin{itemize}[leftmargin=1.4em,itemsep=2pt]
  \item \textbf{The legitimacy of augmented achievement.}
  When a performance shaped by a symbiont counts as the person's own, and how settings can decide this in advance (Section~\ref{sec:acceptable}).
  \item \textbf{Identity and authenticity.}
  Whether a decision thought through with help, or a sentence handed to a person, is still theirs.
  \item \textbf{Measuring influence.}
  How far a decision departed from what the person would have chosen alone, as a test of the commitment not to manipulate (Section~\ref{sec:code}).
  \item \textbf{Responsibility for acting on the person's behalf.}
  Where responsibility lies when an errand goes wrong; the permission levels of Section~\ref{sec:tier1} are the design answer, and the legal answer is open.
  \item \textbf{The boundary of mental-health support.}
  Where support ends and treatment begins, and who is responsible when a handoff fails (Sections~\ref{sec:feeling} and~\ref{sec:mentalhealth}).
  \item \textbf{Children and people who cannot fully consent.}
  What proxy authorization may do, and what a lifelong record means for a child (Sections~\ref{sec:families} and~\ref{sec:notfirst}).
  \item \textbf{Death and inheritance.}
  Who owns a person's record and symbiont after their death, and whether it may be switched off or inherited.
\end{itemize}

\section{Evaluation and Research Agenda}\label{sec:agenda}
A vision has to be testable.
This section says how one would know that a symbiont has augmented a person, what should be studied next, and in what order.

\subsection{What to Measure}\label{sec:measure}
The most important move in the paper is a change of what is measured, from how intelligent the AI is to how much more capable the person has become.
Augmentation is the gain $\daug(x_t)$ of \eqnref{eq:aug}, and three further quantities complete it:
\begin{align}
\dsyn(x_t) &= \Util(x_t;\, h \compose s) - \max\{\Util(x_t;\, h),\, \Util(x_t;\, s)\}, \label{eq:syn}\\
\dret(\tau) &= \Util\!\left(x;\, h^{\mathrm{after}}_{\tau}\right) - \Util\!\left(x;\, h^{\mathrm{before}}\right), \label{eq:ret}\\
\dpers(x_t) &= \Util(x_t;\, h \compose s) - \Util(x_t;\, h \compose s^{-}). \label{eq:pers}
\end{align}
Augmentation asks how much better the person does with the symbiont than without.
Synergy (\ref{eq:syn}) asks whether the pair beats the better of the two alone, which need not be the case: a meta-analysis found that human--AI combinations are often worse than the best of the two~\citep{vaccaro2024combinations}.
Retention (\ref{eq:ret}) asks whether, after a period of use and with the symbiont removed, the person is stronger than before; both terms measure the person unaided on comparable situations $x$, and $\tau$ is the time since the symbiont was removed.
The personalization differential (\ref{eq:pers}) asks how much of the gain comes from the person's context, with $s^{-}$ the same base model and the same agent without personal context.
Augmentation, synergy, and retention can all be positive, or one positive and two negative, and reporting only one is insufficient.
For tasks whose value lies in the person's own experience, emotional and social ones, $\Util(x_t;\, s)$ has no meaning and (\ref{eq:syn}) is not computed.
For the first tier, the differential (\ref{eq:pers}) is the comparison to report, and its control is never ``no AI'', or the capability of the model will be credited to the coupling.
This is the technical form of the sentence that AGI asks how intelligent the machine can be and AHI asks how much intelligence it can enable in a person.

\subsection{Dimensions of Evaluation}\label{sec:dimensions}
\tabref{tab:evaluation} lists ten dimensions with their observables and with what each cannot establish alone.
Two are specific to this paper: the personalization differential of Section~\ref{sec:tier1}, and the losslessness of evolution of Section~\ref{sec:stack}.

\begin{table}[t]
\centering\footnotesize
\caption{Dimensions of evaluation. Each observable is paired with what it cannot establish alone; no single dimension is sufficient.}
\label{tab:evaluation}
\setlength{\tabcolsep}{4pt}\renewcommand{\arraystretch}{1.22}
\begin{tabularx}{\linewidth}{@{}L{3.3cm}C{1.2cm}YY@{}}
\toprule
\tabhead \colhead{Dimension} & \colhead{Measured on} & \colhead{Observable} & \colhead{What it cannot prove on its own} \\
\midrule
\nc{1}\ Accuracy of understanding & \ib{tbTeal}{\faRobot} & whether judgments of situation, intent, and need are right; whether confidence is calibrated & understanding does not imply help \\
\rowcolor{tabStripe}\nc{2}\ Precision of intervention & \ib{tbTeal}{\faRobot} & spoke when it should, stayed silent when it should; false alarms and misses & precision does not imply value \\
\nc{3}\ Delivery within the window & \ib{tbTeal}{\faRobot} & end-to-end latency from event to useful cue; expiry rate & fast does not imply right \\
\rowcolor{tabStripe}\nc{4}\ Personalization differential & \ib{tbViolet}{\faHandshake} & same task, same base model, with and without personal context, \eqnref{eq:pers}: questions asked, corrections, time & a completed task does not imply a stronger person \\
\nc{5}\ Augmentation, synergy, retention & \ib{tbPink}{\faUser} & \eqnref{eq:aug}, \eqnref{eq:syn}, \eqnref{eq:ret} & short-term gain does not imply long-term benefit \\
\rowcolor{tabStripe}\nc{6}\ Cost in attention & \ib{tbPink}{\faUser} & interruptions, time consumed, abandonment & \nna \\
\nc{7}\ Calibration of trust & \ib{tbPink}{\faUser} & rate of accepting correct advice and rejecting wrong advice & high trust may only mean fluent speech \\
\rowcolor{tabStripe}\nc{8}\ Privacy exposure & \ib{tbTeal}{\faRobot}\ \ib{tbGray}{\faUsers} & misuse of the person's and others' information & feeling safe does not imply being safe \\
\nc{9}\ Dependence and autonomy & \ib{tbPink}{\faUser} & skills retained, ability to object, ability to leave & \nna \\
\rowcolor{tabStripe}\nc{10}\ Lossless evolution & \ib{tbTeal}{\faRobot} & consistency of understanding before and after a change of backbone or device & consistent does not imply correct \\
\bottomrule
\end{tabularx}
\tablegend{\textbullet\ Measured on: \ib{tbTeal}{\faRobot}\ the system\quad \ib{tbPink}{\faUser}\ the person\quad \ib{tbViolet}{\faHandshake}\ the pair on one task\quad \ib{tbGray}{\faUsers}\ the people around them.}
\end{table}

\subsection{How to Study a Symbiont}\label{sec:method}
The unit of evaluation is one person, so the methods have to be designed for one person.
Four layers of method follow.
Component benchmarks for memory, personalization, first-person understanding, and task execution~\citep{maharana2024locomo,wu2025longmemeval,salemi2024lamp,yang2025egolife,xie2024osworld} test parts.
Simulation with generative agents~\citep{park2023generative} pre-screens intervention policies and tests mechanisms.
N-of-1 designs, the crossover trials medicine developed for a single patient~\citep{lillie2011nof1}, fit the comparison ``the same person, with it and without it'' naturally.
Longitudinal studies in daily life over weeks and months record gains, costs, exits, and changes in behavior.
The paper's position is that N-of-1 and longitudinal designs are the primary methods of the direction, with at least three baselines: no assistance, a generic assistant, and a personalized task agent on the same base model.
Learning effects cannot be washed out of a crossover, so conditions should be introduced in phases, with few switches, and studies of long-term use must report the people who left and the cases that went badly.

\subsection{Open Problems}\label{sec:openproblems}
\tabref{tab:openproblems} collects twenty-three open problems in six groups: modeling a person and their record, participating in real time, learning across a life, systems and hardware, society and ethics, and governance.
Each problem names what has to be worked out and the sections it depends on, and four are starred because the first usable systems depend on them most: the modeling of a life's record at scale, an end-to-end latency budget from sensor to ear, the cost of always-on cognition, and the iteration of wearable hardware, neural interfaces included, together with the cognitive core.
The groups are ordered from the inside of the loop outward, from the model of the person to the society around them and the rules it sets, and the roadmap of Section~\ref{sec:roadmap} places each problem at the stage of capability at which it becomes unavoidable.

\begin{table}[tp]
\centering\footnotesize
\caption{Open problems of the paradigm, in six groups. Each problem names what has to be worked out and the sections it depends on; the starred problems are the ones on which the first usable systems most depend.}
\label{tab:openproblems}
\setlength{\tabcolsep}{4pt}\renewcommand{\arraystretch}{1.1}
\newcommand{\opgroup}[3]{\rowcolor{#1!14}\multicolumn{4}{@{}l@{}}{\rule{0pt}{2.4ex}\lb{#1}{#2}\ \textbf{#3}}\\}
\newcommand{\opstar}{\textcolor{tbAmber}{\ding{72}}}
\begin{tabularx}{\linewidth}{@{}C{0.9cm}L{3.55cm}YL{1.95cm}@{}}
\toprule
\tabhead \colhead{} & \colhead{Problem} & \colhead{What has to be worked out} & \colhead{Depends on} \\
\midrule
\opgroup{tbViolet}{A}{Modeling a person and their record}
\chipx{tbViolet}{A1} & \textbf{Context and knowledge modeling at scale}\ \opstar & how a life's multimodal stream becomes a record that can be queried in milliseconds, revised, and viewed in many ways; what is kept and at what fidelity; how consolidation and recall across the three timescales are automated & \sch{sec:timescales}\ \sch{sec:streams} \\
\chipx{tbViolet}{A2} & \textbf{Grounded modeling of the mind} & how the observation road and the self-report road are fused, and how their disagreements are represented and resolved & \sch{sec:tworoads} \\
\chipx{tbViolet}{A3} & \textbf{Many selves} & how context-indexed personas are represented and triggered, and how they are kept from collapsing into one average person & \sch{sec:layers} \\
\chipx{tbViolet}{A4} & \textbf{Joint physical and mental world models} & learned from first-person streams, coupled in both directions, and able to detect their own staleness & \sch{sec:twoworldmodels} \\
\chipx{tbViolet}{A5} & \textbf{Backbone-agnostic persistence} & how the information-level interface is specified, and how a migration across backbones is verified to have preserved the relationship & \sch{sec:backbone}\ \sch{sec:stack} \\
\opgroup{tbGreen}{B}{Participating in real time}
\chipx{tbGreen}{B6} & \textbf{Timing} & estimating the attention budget and the window in which help is welcome & \sch{sec:policy} \\
\chipx{tbGreen}{B7} & \textbf{End-to-end low latency}\ \opstar & how the budget from sensor to ear is split across device, edge, and cloud; how the slow loop prepares material for the fast loop; streaming perception and inference & \sch{sec:alwayson}\ \sch{sec:loops} \\
\chipx{tbGreen}{B8} & \textbf{Private channels} & reliability and social acceptability of whispered and silent input & \sch{sec:private} \\
\chipx{tbGreen}{B9} & \textbf{Personalized task execution} & how personal context changes an agent's behavior, and how the personalization differential is measured & \sch{sec:tier1} \\
\opgroup{tbOrange}{C}{Learning across a life}
\chipx{tbOrange}{C10} & \textbf{Learning from sparse experience} & the dialect ladder: remember, compose, adapt, transfer, from few examples and with local parameter change at most & \sch{sec:sparse} \\
\chipx{tbOrange}{C11} & \textbf{Echo-safe learning} & whether conditioning every update on the system's own actions improves the calibration of the personal model & \sch{sec:flywheel} \\
\chipx{tbOrange}{C12} & \textbf{Whole-stack evolution} & lossless change of backbone, harness, and hardware; auditable and bounded self-improvement of the harness & \sch{sec:stack} \\
\opgroup{tbBlue}{D}{Systems and hardware}
\chipx{tbBlue}{D13} & \textbf{The cost of always-on cognition}\ \opstar & how low the duty cycle can drive the daily compute budget per person; the split between local and cloud; when the paradigm becomes economical as inference turns into infrastructure & \sch{sec:economics} \\
\chipx{tbBlue}{D14} & \textbf{Iteration of wearable hardware}\ \opstar & sensors, display, battery, heat, and form factor that meet the requirements of this picture; neural interfaces as a channel; hardware and the cognitive core designed together & \sch{sec:constellation}\ \sch{sec:coevolve} \\
\chipx{tbBlue}{D15} & \textbf{Continuity and interruption} & recovery after a disconnection without pretending to have been present & \sch{sec:continuity} \\
\opgroup{tbPink}{E}{Society and ethics}
\chipx{tbPink}{E16} & \textbf{Bystanders and contextual boundaries} & boundaries that hold in the device and the runtime, and in policy & \sch{sec:bystanders} \\
\chipx{tbPink}{E17} & \textbf{Protocols between symbionts} & whether the four rings hold when both sides optimize for their owners, and what the effects are at the scale of a crowd & \sch{sec:handshake}\ \sch{sec:society} \\
\chipx{tbPink}{E18} & \textbf{Open ethical questions} & seven questions that design and governance leave to research and public deliberation & \sch{sec:ethics} \\
\opgroup{tbAmber}{F}{Governance}
\chipx{tbAmber}{F19} & \textbf{Assistance policies} & how a setting declares the assistance it allows, and how a symbiont recognizes and honors the declaration & \sch{sec:settings} \\
\chipx{tbAmber}{F20} & \textbf{Verifiable abstention} & a proof that a symbiont kept a policy, which a modified device cannot forge and which reveals nothing about the record & \sch{sec:abstention} \\
\chipx{tbAmber}{F21} & \textbf{Disclosure of assistance} & records of when assistance was given, and norms for when to disclose them, in conversation and in published work & \sch{sec:status}\ \sch{sec:conversations} \\
\chipx{tbAmber}{F22} & \textbf{Assessment for augmented learners} & credentials that measure unaided ability where a role requires it, and retention in learning & \sch{sec:credentials}\ \sch{sec:measurecred} \\
\chipx{tbAmber}{F23} & \textbf{Effects on the collective} & measuring inequality of access, the pressure to adopt, and the convergence of thought across a population & \sch{sec:society} \\
\bottomrule
\end{tabularx}
\tablegend{\textbullet\ \opstar\ problems on which the first usable systems most depend.\quad \textbullet\ Problem identifiers are used again in \figref{fig:roadmap}, which places each problem at the stage of capability at which it becomes unavoidable.}
\end{table}

\subsection{A Capability Roadmap}\label{sec:roadmap}
\begin{figure}[t]
  \centering
  \resizebox{0.98\linewidth}{!}{\begin{tikzpicture}[x=1cm,y=1cm,font=\scriptsize\sffamily,
  st/.style={draw=#1!80!black,fill=#1!10,rounded corners=4pt,text width=2.0cm,minimum height=2.45cm,align=center,inner sep=3pt,line width=0.8pt,anchor=south},
  badge/.style={circle,fill=#1!80!black,text=white,font=\tiny\sffamily\bfseries,inner sep=0pt,minimum size=0.42cm},
  gate/.style={text width=2.3cm,align=center,text=symSlate,anchor=north,font=\scriptsize\sffamily},
  chips/.style={text width=2.3cm,align=center,anchor=north,inner sep=0pt}]
  \def\dx{2.5}\def\dy{0.42}
  \fill[symAccent!6] (-0.35,-0.2) -- (-0.35,0) \foreach \i in {0,...,5}{ -- (\i*\dx-1.1,\i*\dy) -- (\i*\dx+1.1,\i*\dy) } -- (5*\dx+1.3,5*\dy) -- (5*\dx+1.3,-0.2) -- cycle;
  \draw[symAccent!35,line width=0.6pt] (-0.35,0) \foreach \i in {0,...,5}{ -- (\i*\dx-1.1,\i*\dy) -- (\i*\dx+1.1,\i*\dy) } -- (5*\dx+1.3,5*\dy);
  \foreach \i/\col/\lt/\icon/\ttl/\kw/\gte/\ch in {
    0/stSense/1/\faDesktop/{Desktop copilot}/{streamed cues in meetings; errands by delegation; private cues from day one}/{cues used in real meetings; fewer questions asked}/{\chipx{tbGreen}{B6}\,\chipx{tbGreen}{B7}\,\chipx{tbGreen}{B9}},
    1/stKeep/2/\faDatabase/{Persistent record}/{one record across days, settings, and entry points; a personal model}/{fewer re-explanations; repair propagates}/{\chipx{tbViolet}{A1}\,\chipx{tbViolet}{A2}\,\chipx{tbViolet}{A5}\,\chipx{tbBlue}{D15}},
    2/symCyan/3/\faGlasses/{Device constellation}/{glasses, pendant, earbuds, physiology; channels out}/{duty cycle within battery and latency; bystanders can tell}/{\chipx{tbBlue}{D13}\,\chipx{tbBlue}{D14}\,\chipx{tbPink}{E16}},
    3/symViolet/4/\faCommentDots/{Private input}/{whisper, silent input; the self-report road}/{whisper reliable; acceptable in a crowd}/{\chipx{tbGreen}{B8}\,\chipx{tbViolet}{A3}},
    4/stGrow/5/\faSeedling/{Lifelong co-adaptation}/{online learning, echo-safe updates, whole-stack evolution}/{calibration improves; the relationship survives upgrades}/{\chipx{tbOrange}{C10}\,\chipx{tbOrange}{C11}\,\chipx{tbOrange}{C12}\,\chipx{tbViolet}{A4}},
    5/symMagenta/6/\faPeopleArrows/{Networks of symbionts}/{the handshake; effects at the scale of a crowd}/{exchange stays inside the ring; group effects measured}/{\chipx{tbPink}{E17}}}{
    \node[st=\col] (s\i) at (\i*\dx,\i*\dy) {{\color{\col!75!black}\large\icon}\\[1pt]{\footnotesize\bfseries\ttl}\\[2pt]\kw};
    \node[badge=\col] at ([xshift=2pt,yshift=-2pt]s\i.north west) {\lt};
    \node[gate] (g\i) at ([yshift=-0.12cm]s\i.south) {\textcolor{symAccent}{\ding{117}}\ \textbf{to advance}\\\gte};
    \node[chips] at ([yshift=-0.08cm]g\i.south) {\ch};
    \ifnum\i>0 \draw[-{Stealth[length=1.9mm]},symSlate,line width=0.8pt] (s\the\numexpr\i-1\relax.east) -- (s\i.west); \fi}
  \draw[tbPink!70!black,line width=0.8pt,dashed] (-0.35,-2.2) -- (5*\dx+1.3,-2.2);
  \node[anchor=north west,text=tbPink!70!black,font=\scriptsize\sffamily] at (-0.35,-2.26) {\chipx{tbAmber}{F19--F23}\ \chipx{tbPink}{E18}\ governance and the open ethical questions run beneath every stage};
  \node[anchor=south west,text=symSlate,font=\scriptsize\sffamily\itshape] at (-0.35,4.75) {stages accumulate capability; they are not a product schedule};
  \node[anchor=north west,text=symSlate,font=\scriptsize\sffamily] at (-0.35,4.7) {chips: the open problems of \tabref{tab:openproblems} that each stage makes unavoidable};
  \draw[-{Stealth[length=2mm]},symAccent,line width=0.9pt] (5*\dx-1.3,5*\dy+2.75) -- (5*\dx+1.3,5*\dy+2.75) node[pos=0,anchor=east,font=\tiny\sffamily,text=symAccent] {capability accumulates};
\end{tikzpicture}}
  \caption{From components to a life.
  Six stages of accumulating capability, the evidence that would justify moving to the next, and the open problems of \tabref{tab:openproblems} mapped onto them.}
  \label{fig:roadmap}
\end{figure}
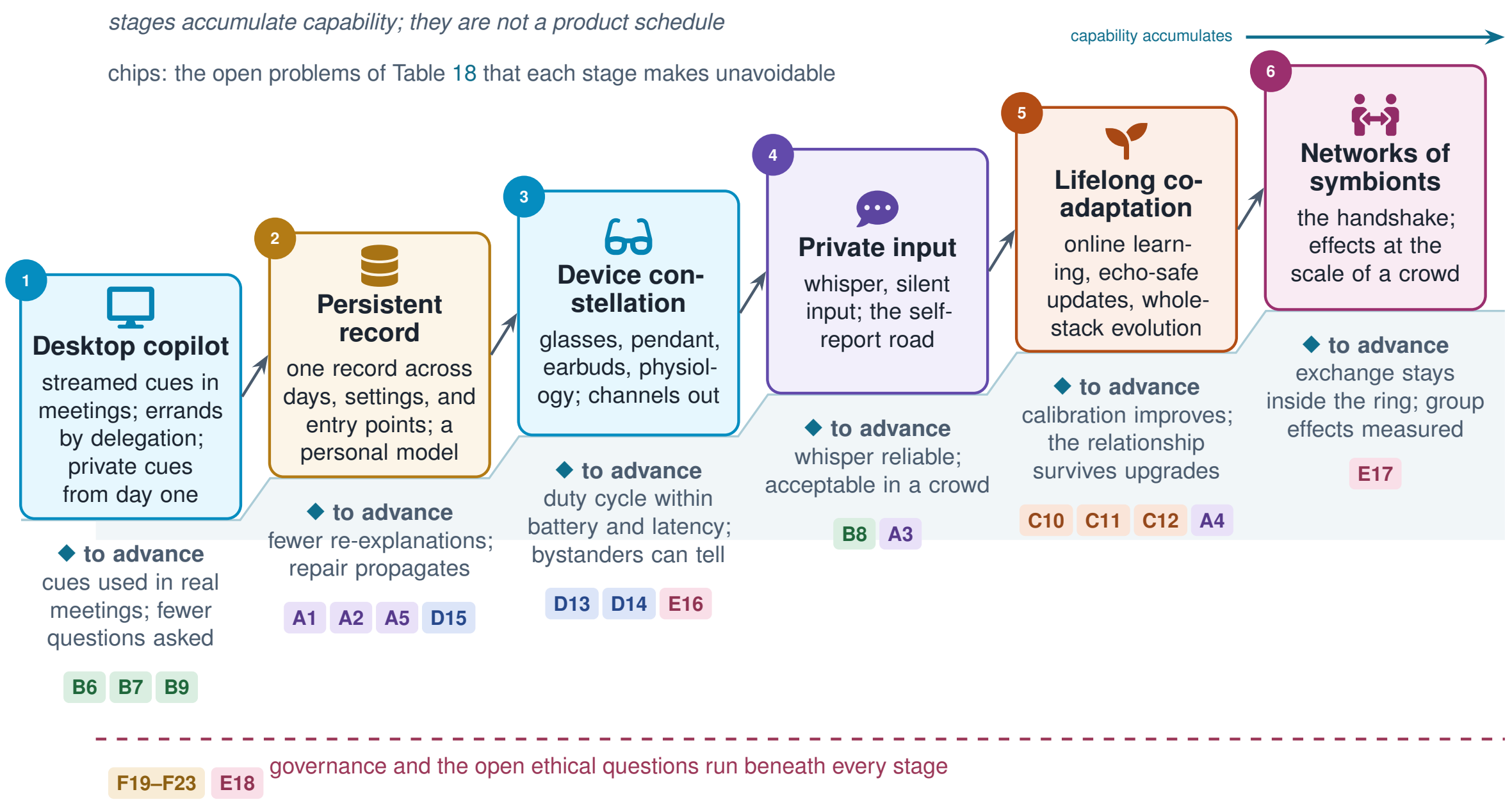

\figref{fig:roadmap} orders capabilities in six stages, from a streaming cognitive copilot on the desktop with Tier~1 by delegation, to a persistent record and model, to the device constellation, to private input, to lifelong co-adaptation, and to networks of symbionts, with the evidence needed to move from each stage to the next and the open problems mapped onto them.
The stages accumulate capability and are not a product schedule; full-duplex, private cues, and Tier~1 are requirements from the first stage onward.

\subsection{What Would Prove Us Wrong}\label{sec:falsify}
The claims of the paper should be narrowed if any of the following holds: more personal context brings no measurable gain on most tasks, including a zero personalization differential on the first tier; timely intervention is no better than simple on-demand answering; long-term use generally erodes abilities people want to keep; value is obtained only at an unacceptable cost in surveillance or compute; people will not authorize anything above the desktop; or every change of backbone loses the relationship and cannot be prevented.
Two further conditions concern governance: assistance policies and proofs of abstention cannot be made reliable, so that a symbiont cannot be told apart from a tool for cheating; or the gains of augmentation in the main competitive settings prove to be mostly positional, so that adoption brings an arms race and no net gain.
The first six conditions correspond to dimensions of \tabref{tab:evaluation}, and the last two to the governance problems of \tabref{tab:openproblems}.

\FloatBarrier
\section{Conclusion}\label{sec:conclusion}
The past decade was spent making machines more intelligent; the next is worth spending on making each particular person stronger.
As AI enters the physical world it will take two forms of intelligent individual, robots with bodies of their own and people augmented by AI, and this paper has argued for the second, giving it a name, AI-augmented human intelligence, a form, the cognitive symbiont, and a map.
A symbiont is lifelong, embedded, real-time, private, low-friction, and proactive; it stands on foundation models, agents and their harnesses, memory and context engineering, world models, and wearable hardware; and it climbs a ladder from doing the person's errands with their context in hand, to thinking a problem through with them, to holding them steady among other people.
The object of evaluation changes with it, from what the AI completes to what the person can now do, measured as augmentation, synergy, and retention.
The map has blank regions, and the three largest are the evidence on which a mind can be modeled, the timing of participation, and the stability of learning and evolution in the course of a life; the heaviest region is the governance of augmentation in the settings where people compete with and trust one another.
The paradigm will have succeeded when the sign of it is a person who has become more like who they wanted to be, and when no one mistakes that for the AI having become more like a person.

\bibliographystyle{acl_natbib}
{\small\bibliography{references}}
\end{document}